\documentclass[12pt]{article}
\usepackage[english]{babel}
\usepackage{url}
\usepackage[utf8x]{inputenc}
\usepackage[font=footnotesize,labelfont=bf]{caption}
\usepackage{amsmath}
\usepackage{amssymb}
\usepackage{graphicx}
\usepackage{wrapfig}
\graphicspath{{images/}}
\usepackage{parskip}
\usepackage{fancyhdr}
\usepackage{vmargin}
\usepackage{adjustbox}
\usepackage{makecell}
\usepackage{multirow}
\usepackage{color}
\usepackage[dvipsnames]{xcolor}
\usepackage{subcaption}
\usepackage{graphicx,rotating,booktabs}

\DeclareMathOperator*{\argmax}{arg\,max}
\usepackage[verbose]{placeins}
\newcommand{\code}[1]{\texttt{#1}}

\setmarginsrb{1.5 cm}{0 cm}{1.5 cm}{0.5 cm}{1 cm}{1.5 cm}{1 cm}{1.5 cm}

\title{FireCommander: An Interactive, Probabilistic Multi-agent Environment for Heterogeneous Robot Teams}                             % Title
\author{Esmaeil Seraj}                               % Author
\date{\today}                                           % Date

\makeatletter
\let\thetitle\@title
\let\theauthor\@author
\let\thedate\@date
\def\blfootnote{\xdef\@thefnmark{}\@footnotetext}
\makeatother

\begin{document}

%%%%%%%%%%%%%%%%%%%%%%%%%%%%%%%%%%%%%%%%%%%%%%%%%%%%%%%%%%%%%%%%%%%%%%%%%%%%%%%%%%%%%%%%%

\begin{titlepage}
    \centering 
	\includegraphics[scale = 0.5]{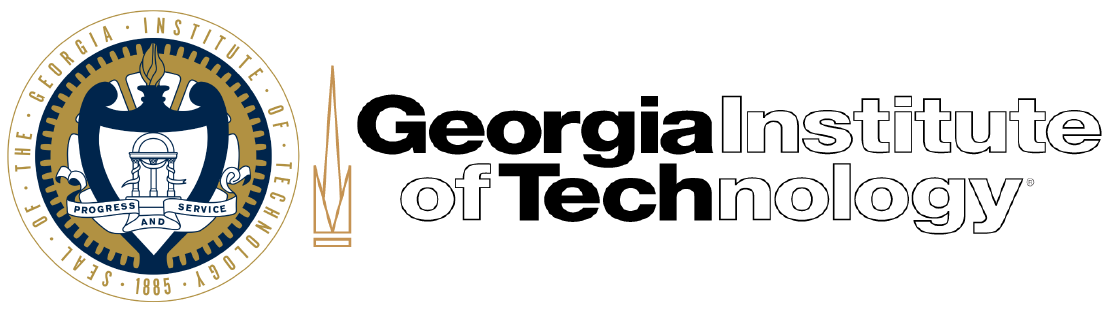}\\[1.0 cm]    
    \includegraphics[scale = 0.5]{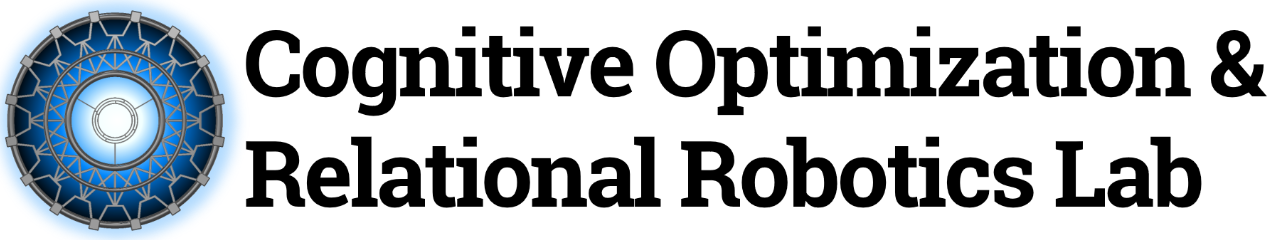}\\[2.5 cm] 
                   
    \rule{\linewidth}{0.5 mm} \\[0.4 cm]
    { \Large \bfseries \thetitle}\\
    \rule{\linewidth}{0.5 mm} \\[1 cm]
    
    {\large Esmaeil Seraj$ ^{\dagger, *} $, Xiyang Wu$ ^{\dagger} $ and Matthew C. Gombolay$ ^{\dagger} $}\\[2.0 cm]
    
    \textsf{\large Georgia Institute of Technology, Atlanta (GA), United States}\\[0.5 cm]               
    \textsf{Institute for Robotics \& Intelligent Machines}\\[0.5 cm]
    
    \textsf{CORE Robotics Lab}\\[1 cm]
    
    \vspace*{2cm}
    \rule{\linewidth}{0.1 mm}
    \begin{flushleft}
    {\scriptsize $ ^* $ \textbf{Corresponding author:} Correspondences shall be forwarded to: \textit{electronic mail:} eseraj3@gatech.edu}\\[0.1 cm]
    \end{flushleft}
    \begin{flushleft}
    {\scriptsize $ ^\dagger $E. Seraj and M. Gombolay are with the Institute for Robotics \& Intelligent Machines (IRIM), School of Interactive Computing. All authors are affiliates of the Georgia Institute of Technology, Atlanta (GA), United States (30332).}\\
    {\scriptsize $^1$\underline{Version:} User Guide Version 2.1 }\\ [0.5 cm]
    \end{flushleft}
    {\large \thedate}\\
 
    \vfill
    
\end{titlepage}

%%%%%%%%%%%%%%%%%%%%%%%%%%%%%%%%%%%%%%%%%%%%%%%%%%%%%%%%%%%%%%%%%%%%%%%%%%%%%%%%%%%%%%%%%

\vspace*{3.5 cm} 
\begin{abstract}
\noindent The purpose of this tutorial is to help individuals use the \underline{FireCommander} game environment for research applications. The FireCommander is an interactive, probabilistic joint perception-action reconnaissance environment in which a \textit{composite} team of agents (e.g., robots) cooperate to fight dynamic, propagating firespots (e.g., targets). In FireCommander game, a team of agents must be tasked to optimally deal with a wildfire situation in an environment with propagating fire areas and some facilities such as houses, hospitals, power stations, etc. The team of agents can accomplish their mission by first sensing (e.g., estimating fire states), communicating the sensed fire-information among each other and then taking action to put the firespots out based on the sensed information (e.g., dropping water on estimated fire locations). The FireCommander environment can be useful for research topics spanning a wide range of applications from Reinforcement Learning (RL) and Learning from Demonstration (LfD), to Coordination, Psychology, Human-Robot Interaction (HRI) and Teaming. There are four important facets of the FireCommander environment that overall, create a non-trivial game:
\begin{enumerate}
\item \textbf{Complex Objectives:} Multi-objective game, heterogeneous agents.

\item \textbf{Stochastic Environment:} Agents' actions result in probabilistic performance.

\item \textbf{Partially Observable:} Hidden targets that need to be explored and discovered.

\item \textbf{Uni-task Robots:} Perception-only and Action-only agents.
\end{enumerate}
\vspace*{0.5cm}
\noindent The FireCommander environment is first-of-its-kind in terms of including Perception-only and Action-only agents for coordination. It is a general multi-purpose game that can be useful in a variety of combinatorial optimization problems and stochastic games, such as applications of Reinforcement Learning (RL), Learning from Demonstration (LfD) and Inverse RL (iRL) to the following list of problems (see Introduction for details): 
\begin{enumerate}
\item {Multi-agent Coordination/Cooperation and Communication Learning}

\item {Multi-agent Learning from Heterogeneous Demonstrations (MA-LfHD)}

\item {Multi-robot Planning/Scheduling and Task Assignment (MRTA)}

\item {Human-Robot Interaction (HRI), Human-Robot Teaming and Psychology}
\end{enumerate}
\vspace*{0.5cm}
\noindent FireCommander  is open-source at\footnote{\scriptsize Distributed under the terms of the GNU GENERAL PUBLIC LICENSE as a set of Python functions.}~\cite{seraj2020firecommander}:

\begin{center} 
\underline{\texttt{https://github.com/EsiSeraj/FireCommander2020}}
\end{center}

\noindent A PowerPoint tutorial for FireCommander can be found at:

\begin{center} 
\underline{\texttt{https://hal.archives-ouvertes.fr/hal-02995093v1}}
\end{center}

\vspace*{1cm}
\end{abstract}

%%%%%%%%%%%%%%%%%%%%%%%%%%%%%%%%%%%%%%%%%%%%%%%%%%%%%%%%%%%%%%%%%%%%%%%%%%%%%%%%%%%%%%%%%

\newpage
\tableofcontents
\pagebreak

%%%%%%%%%%%%%%%%%%%%%%%%%%%%%%%%%%%%%%%%%%%%%%%%%%%%%%%%%%%%%%%%%%%%%%%%%%%%%%%%%%%%%%%%%

\section{Introduction}
\label{sec:gettingstarted}
\begin{wrapfigure}{r}{0.5\textwidth}
	\centering
	\includegraphics[width=0.48\columnwidth]{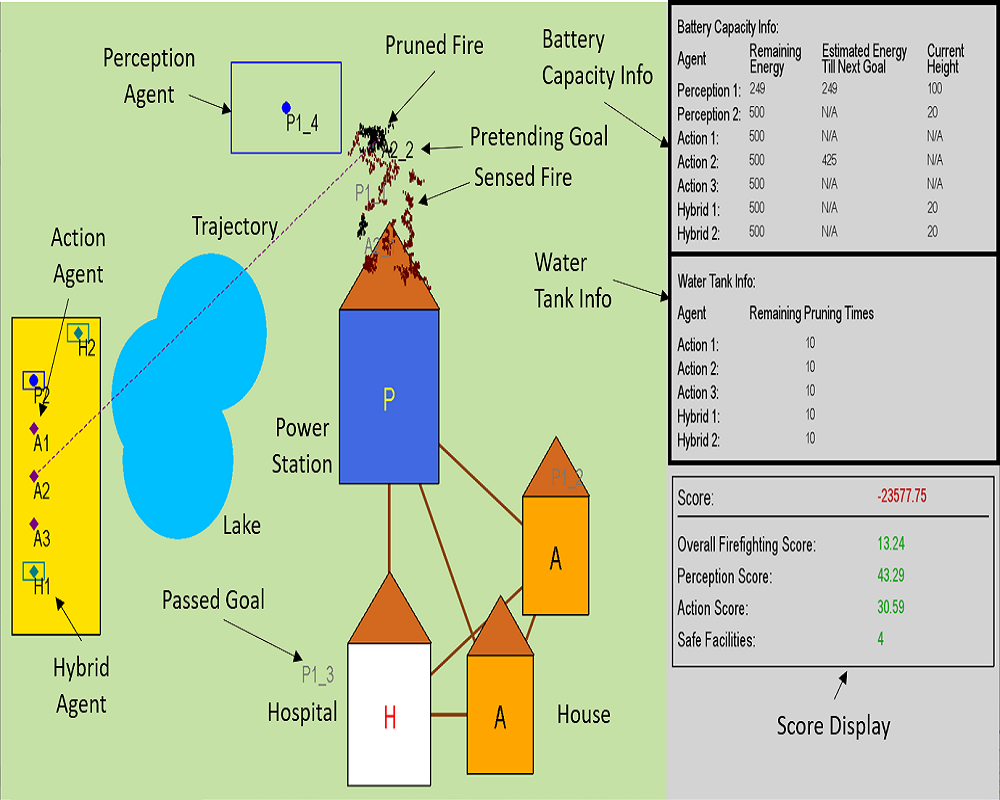}
	\caption{An example screenshot from FireCommander game Environment.}
	\label{fig:exampleScreen}
\end{wrapfigure}

\noindent The purpose of this tutorial is to help individuals use the \underline{FireCommander} game environment for research applications. The FireCommander (Figure~\ref{fig:exampleScreen}) is an interactive, probabilistic joint perception-action reconnaissance environment in which a \textit{composite} team of heterogeneous agents (e.g., specific types of simulated robots) cooperate to fight dynamic, propagating firespots (e.g., targets). 
In FireCommander game, a team of agents must be tasked to optimally deal with a wildfire situation in an environment with propagating fire areas and some facilities such as houses, hospitals, power stations, etc. The team of agents can accomplish their mission by first sensing (e.g., estimating fire states), communicating the sensed fire-information among each other and then taking action to put the firespots out based on the sensed information (e.g., dropping water on estimated fire locations). The FireCommander environment can be useful for research topics spanning a wide range of applications from Reinforcement Learning (RL) and Learning from Demonstration (LfD), to Coordination, Psychology, Human-Robot Interaction (HRI) and Teaming. There are four important facets of the FireCommander environment that overall, create a non-trivial game:

\hspace*{0.4cm} 1. \textbf{Complex, Multi-Objective Game:} The environment is generally designed as a multi-purpose game in which robots (e.g., Perception-agents which are only capable of sensing and Action-agents which can manipulate but cannot sense) are tasked by a user to collaboratively fight multiple propagating wildfires and protect facilities (e.g, power stations, hospital, houses, etc.) from the fire. Agents are constrained by limited resources, such as battery-life and tanker capacity as well as motion restrictions, communication constraints and time limitations.

\hspace*{0.4cm} 2. \textbf{Stochastic and Probabilistic Environment:} There are three sources of stochasticity in the FireCommander environment. All agents' actions change the states of the environment stochastically such that: (1) Action-agents can put out the fires in their field-of-view (FOV) according to a random probability distribution, which is designed through a confidence level coefficient. (2) Perception uncertainty varies with respective Perception-agent’s altitude. Altitude has direct and reversed relation with observable area (e.g., FOV) and sensing quality (uncertainty), respectively. We model a perception agent's altitude-dependent sensing quality (e.g., the lower the altitude, the higher the quality of estimation) such that, perception-agents can sense (e.g., detect/locate) the fires within their FOV according to a random probability distribution which depends on their altitude. A perception agent at its lowest safe-altitude can sense 100\% of the firespots in its FOV while a perception agent at its highest allowable altitude can only sense 40\% of the firespots within its FOV. (3) The third stochasticity is associated with the fire behavior. Fire can appear at anytime during the game, anywhere on the map. Moreover, fire propagates according to a stochastic mathematical model (e.g., see FARSITE in Section~\ref{subsubsec:FARSITE}).

\hspace*{0.4cm} 3. \textbf{Partially Observable Environment:} Initially, no firespot is visible on the screen. Users will only see a raw map, including the UAV base (e.g., robot depot) and the various facilities on the map. Firespots can only be seen through Perception-agents. Fires can also appear at anytime anywhere around the map. Moreover, since the firespots are dynamic, once the Perception agent leaves the fire location, the new location of the firespot becomes uncertain, until a Perception agent is summoned to the area again for an updated observation.

\hspace*{0.4cm} 4. \textbf{Composite Robot Team:} We define a composite robot team as a group of agents that perform different tasks according to their respective capabilities while their tasks are co-dependent on accomplishing an overarching mission. In FireCommander environment, the robot team is composed of (1) perception-only and (2) action-only agents. As such, we introduce \textit{Perception-agents} and \textit{Action-agents} in this environment, which together, form a composite robot team.  As such, the designed coordination policies must take into account the Perception-Action hierarchy and communication problems (e.g., firespots must first be observed by a perception agent, the sensed information must be communicated to an action agent and then, the firespots are put out by the action agent.)

\noindent The FireCommander environment is first-of-its-kind in terms of including Perception-only and Action-only agents for coordination. Considering such composite agents helps to represent a clearer picture of coordination and efficient communication problems in heterogeneous teams. FireCommander is a handcrafted environment than can be modified to cover a wide range of game complexities from simple, single-objective games with only one (or two in heterogeneous agent cases) agents, to complex, multi-objective stochastic games with numerous agents and constraints. Many of the existing multi-agent environments, such as the OpenAI multi-agent particle environments\footnote{Available Online: \underline{\texttt{https://github.com/openai/multiagent-particle-envs}}} are not inherently designed to include heterogeneous agents with different capabilities and are often heavily modified to pose a meaningful heterogeneous communication or coordination problem. Moreover, other environments such as StarCraft II are often too complex and hard, if not impossible, to be modified to match a specific scenario, while FireCommander is open-source and the game logistics can be easily modified to adapt a desirable setting. Additionally, FireCommander is a general multi-purpose game that is designed to be readily leveraged in a variety of combinatorial optimization problems~\cite{papadimitriou1998combinatorial,wolsey1999integer,schrijver2003combinatorial,sghir2018multi}, and stochastic games~\cite{shapley1953stochastic,littman1994markov}, such as applications of Reinforcement Learning (RL)~\cite{sutton2018reinforcement}, Learning from Demonstration (LfD)~\cite{argall2009survey} and Inverse RL (iRL)~\cite{gao2012survey} to the following list of problems:
\begin{enumerate}
\item \textbf{Multi-agent Coordination/Cooperation and Communication Learning:} Due to inherently heterogeneous agents (e.g., agents with different capabilities), the FireCommander environment poses an interesting multi-agent coordination problem for Perception and Action-agents to learn how to cooperate and how to efficiently communicate. Figure.~\ref{fig:FireCommanderGame} represents the FireCommander environment logic and various multi-agent coordination and communication problems it covers. We have developed various ready-to-use versions of the FireCommander which can be directly leveraged to test multi-agent reinforcement learning (MARL) algorithms. We also developed a graphical user interface (GUI) which can be used to record expert data for learning from demonstration (LfD) and human-robot interaction (HRI) studies dealing with multi-agent, heterogeneous coordination/cooperation and communication Learning, similar to the works in~\cite{foerster2016learning,sukhbaatar2016learning,zhang2013coordinating,ghavamzadeh2004learning,jiang2018learning,kim2019learning,mao2017accnet,zhang2019efficient, seraj2021heterogeneous}. Please refer to Section~\ref{subsubsec:CoordinationCooperationCommunicationLearning} for more details.
\begin{figure}[t!]
\centering
\includegraphics[width=\textwidth]{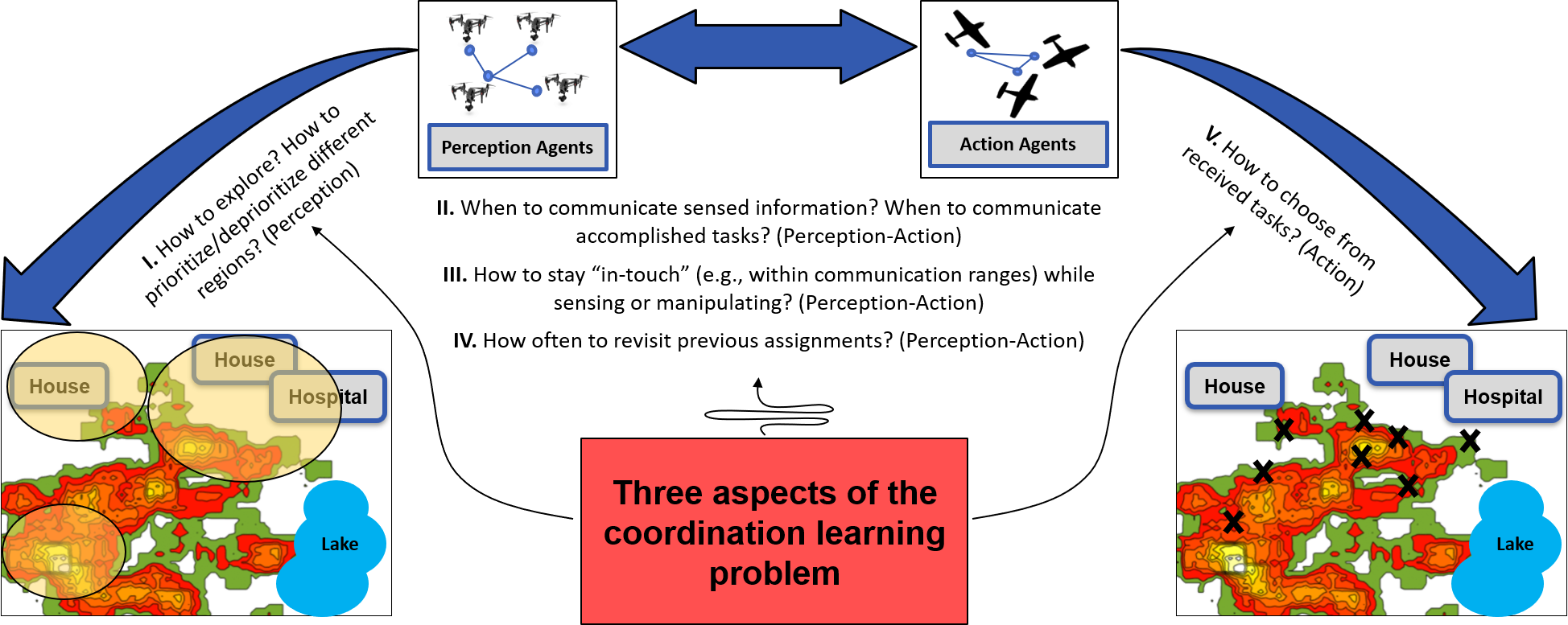}
\caption{The FireCommander environment logic and various multi-agent coordination problems it covers. The environment generally includes three aspects of the multi-agent coordination problem: (1) coordination among Perception-agents and how they optimize their performance in the environment, (2) coordination among Action-agents and how they optimize their performance in the environment, and (3) coordination among Perception and Action (e.g., Perception-Action) agents and how they optimize their communication for improved collective behavior in the environment~\cite{seraj2020firecommanderslides}.}
\label{fig:FireCommanderGame}
\end{figure}

\item \textbf{Multi-agent Learning from Heterogeneous Demonstrations (MA-LfHD):} FireCommander  inherently includes two categories of heterogeneous robots: (1) Perception (e.g., Sensing) agents and (2) Action (e.g., Manipulator) agents, and is a multi-objective game. Missions and games in FireCommander do not have unique solutions and each category of agents must learn both a local team-objective (e.g., either perform sensing or acting) as well as a global composite-objective (e.g., fight the fire and protect the facilities). Accordingly, the game objectives can be interpreted and interacted with in various different ways by a user and thus, FireCommander can be a perfect environment for developing learning from \textit{heterogeneous demonstrations} algorithms~\cite{chen2020joint,song2018multi,yu2019multi,vsovsic2016inverse,paleja2019inferring}, particularly for multi-agent coordination.

\item \textbf{Multi-robot Planning/Scheduling and Task Assignment (MRTA):} As described in Section~\ref{subsubsec:CombinatorialOptimization}, in FireCommander, Perception and Action agents must efficiently coordinate to cooperatively fight propagating fires. Doing such however, requires solving various combinatorial optimization problems (COPs) such as ``\textit{How should agents choose where to go?}", or ``\textit{How should agents divide up the tasks?}", or ``\textit{How should agents be distributed among tasks, areas of the map, etc., given we have limited resources?}", and more. Solving for these questions falls within the realms of combinatorial optimization (see Section~\ref{subsubsec:CombinatorialOptimization}). Our FireCommander environment can be a complex environment for many challenging COPs such as dynamically learning scheduling policies~\cite{wang2019learning,wang2020learning}, and multi-agent task assignment~\cite{ravichandar2019strata}.

\item \textbf{Human-Robot Interaction (HRI), Human-Robot Teaming and Psychology:} FireCommander includes an interactive game-based graphical user interface (GUI) which can be of great use in HRI researches. As an instance, the FireCommander game can be leveraged to design environments with heavy/light workload and then test how an expert’s policy design efficiency and quality is affected under situational stress. Various other HRI topics can be similarly modeled to leverage FireCommander as their test-bed, be such as trust and accountability~\cite{natarajan2020effects,heyer2010human}, anthropomorphism~\cite{zlotowski2015anthropomorphism,fink2012anthropomorphism,natarajan2020effects}, human-robot co-adaptation~\cite{gao2017personalised}, human-guided optimization~\cite{schaff2020residual,jevtic2018robot,huttenrauch2017guided}, cognitive BCI~\cite{seraj2017robust,karimzadeh2017distributed,seraj2017improved,sameni2017robust,seraj2016investigation} and many more~\cite{goodrich2008human,kolling2013human,seraj2019instantaneous,karimzadeh2015presenting}.

\item \textbf{Wireless Sensor and Actor Networks (WSAN):} Wireless sensor and actor networks refer to a group of sensors and actors linked by wireless medium to perform distributed sensing and acting tasks~\cite{akyildiz2004wireless}. FireCommander environment is by definition a WSAN, in which Perception-agents play the role of wireless sensors and Action-agents reflect the wireless actors. Asu such, FireCommander is a perfectly fit environment for developing coordination algorithms for WSANs~\cite{melodia2005distributed,melodia2007communication,akyildiz2004wireless}. Moreover, WSANs might include static sensors and actors, static sensors and dynamic actors or vice versa. FireCommander environment can easily be modified to match these settings while other existing games and environments lack such flexibility.
\end{enumerate}

\subsection{License - No Warranty}
\label{subsec:license}
FireCommander: An Interactive, Probabilistic Multi-agent Environment for Heterogeneous Robot Teams.

Copyright (C) 2020 Esmaeil Seraj, Xiyang Wu and Matthew C. Gombolay

This program is free software; you can redistribute it and/or modify it under the terms of the GNU GENERAL PUBLIC LICENSE as published by the Free Software Foundation; either version 3.0 of the License, or (at your option) any later version.

This program is distributed in the hope that it will be useful, but WITHOUT ANY WARRANTY; without even the implied warranty of MERCHANTABILITY or FITNESS FOR A PARTICULAR PURPOSE. See the GNU GENERAL PUBLIC LICENSE for more details. 

You should have received a copy of the GNU GENERAL PUBLIC LICENSE along with this program; if not, see $ \langle $ \underline{\texttt{http://www.gnu.org/licenses/}} $ \rangle $ or write to the Free Software Foundation, Inc., 51 Franklin Street, Fifth Floor, Boston, MA  02110-1301, USA.

\subsection{Citations: Code-Base, Tutorials and Documentations}
\label{subsec:citation}
Within the limits of the GNU GENERAL PUBLIC LICENSE, you can use the toolbox as you please; however, if you use the toolbox in a work of your own that you wish to publish, please make sure to cite the code-base on GitHub~\cite{seraj2020firecommander} this user manual~\cite{seraj2020firecommandermanual} properly, as shown below. This way you will contribute to helping other scholars find these items.

\begin{itemize}

\item Esmaeil Seraj, Xiyang Wu and Matthew C. Gombolay, "FireCommander", (2020), GitHub Repository, Release 1.1, [Online] https://github.com/EsiSeraj/FireCommander2020\footnote{
\code{@misc\{seraj2020FireCommander,\\
  \hspace*{0.75cm}author = {Seraj, Esmaeil and Wu, Xiyang, and Gombolay, Matthew},\\
  \hspace*{0.75cm}title = {FireCommander},\\
  \hspace*{0.75cm}volume = {Release 1.1},\\
  \hspace*{0.75cm}year = {2020},\\
  \hspace*{0.75cm}publisher = {GitHub},\\
  \hspace*{0.75cm}journal = {GitHub Repository},\\
  \hspace*{0.75cm}howpublished = {\url{https://github.com/EsiSeraj/FireCommander2020}}\\
  \hspace*{0.35cm}\}}}.

\item Esmaeil Seraj, Xiyang Wu and Matthew C. Gombolay, "FireCommander: An Interactive, Probabilistic Multi-agent Environment for Joint Perception-Action Tasks" arXiv Preprint, 2020.

\end{itemize}

\subsection{Download and Utilization}
\label{subsec:download}
The latest version of the "FireCommander" software can be downloaded directly from from the public repository on GitHub, at~\cite{seraj2020firecommander}:

\begin{center} 
\underline{\texttt{https://github.com/EsiSeraj/FireCommander2020}}
\end{center}

\subsection{Getting Help}
\label{subsec:help}
The codes are standardized and commented as frequently as possible. Moreover, the current user manual tutorial, a PPT tutorial, a video tutorial and a project publication, all are provided to help individuals utilize this code in their research studies. Links to the mentioned blog-posts, tutorials and documentations are presented below:

\begin{itemize}
    \item \textbf{Code \& GitHub Blog:}~\cite{seraj2020firecommander} ................ \underline{\texttt{https://github.com/EsiSeraj/FireCommander2020}}
    
    \item \textbf{User Manual Tutorial:}~\cite{seraj2020firecommandermanual} ................................ \underline{\texttt{https://arxiv.org/pdf/1907.02862.pdf}}
    
    \item \textbf{PowerPoint Tutorial:}~\cite{seraj2020firecommanderslides} .............. \underline{\texttt{https://hal.archives-ouvertes.fr/hal-02995093v1}}
    
    % \item \textbf{Video Tutorial:} ................................................................. \underline{\texttt{https://youtu.be/UQsWPh9c3eM}}
    
    % \item \textbf{Project Publication:} ......................................... \underline{\texttt{https://arxiv.org/pdf/1907.02862.pdf}}
\end{itemize}

You can also contact the corresponding author\footnote{Esmaeil (Esi) Seraj: \texttt{eseraj3@gatech.edu}} directly to ask any related questions or discuss possible difficulties or errors you might encounter. Please feel free to contact in either case.

\section{Environment User Guide}
\label{sec:userguide}

\subsection{Fundamentals and General Applicability}
\label{subsec:fundamentals}
\subsubsection{Multi-agent Systems}
\label{subsubsec:Multi-agentSystems}
\noindent In general, an agent is an autonomous physical (e.g., robot) or virtual (e.g., simulation) entity that can act, perceive its environment (e.g., fully or partially) and communicate, and has skills to achieve a desired goals. In a multi-agent system, multiple agents \textit{interact} in an environment and on objects to accomplish some local or global objectives. Interactions can be defined as relations between all the entities, a set of operations that can be performed by the entities and the changes in the environment over the time and due to agents' actions~\cite{ferber1999multi}. Agents in a multi-agent system can be of similar type and with similar dynamics (e.g., homogeneous agents) or otherwise (e.g., heterogeneous agents). 

Multi-agent teams are able to execute time-sensitive, complex missions by cooperatively leveraging their unique capabilities and design~\cite{busoniu2008comprehensive,seraj2020hierarchical,ravichandar2019strata,seraj2021adaptive}. Heterogeneity in robots' design characteristics and their roles are introduced into these multi-robot systems to (1) leverage the relative merits of the different agents and their capabilities~\cite{busoniu2008comprehensive,seraj2020hierarchical,ravichandar2019strata} and, (2) deal with the dynamic and unpredictable nature of the real-world for which designing homogeneous, versatile robot teams that can effectively adjust to all circumstances is difficult and costly~\cite{busoniu2008comprehensive,seraj2020hierarchical,rizk2019cooperative,seraj2021heterogeneous}.

A set of operations/policies taken by agents in a multi-agent system forms interaction. Interacting agents in a multi-agent systems create a notion of Markov Games (MG)~\cite{littman1994markov}, which are a special case of Stochastic Games (SG)~\cite{shapley1953stochastic}. Interactions among a set of $N$ agents in an SG can generally take three forms: (1) Cooperative~\cite{panait2005cooperative,kapetanakis2002reinforcement,tan1993multi}, (2) Non-cooperative~\cite{tan1993multi,pendharkar2012game,etesami2020non} and (3) Competitive~\cite{guttman1998cooperative,wang2018competitive}. Cooperative agents have the same reward structure and the goal of the system is to \textit{collectively} maximize a common discounted return~\cite{busoniu2008comprehensive,panait2005cooperative,kapetanakis2002reinforcement,tan1993multi}. Non-cooperative agents also try to achieve a common goal; however, they do not behave collectively and each agent is only trying to maximize their own local rewards (which leads to maximizing the total return as well)~\cite{tan1993multi,pendharkar2012game,etesami2020non}. Competitive agents are similar to non-cooperative agents in only caring about their own local rewards; however, in a competitive SG setup, maximizing each agent's local return does not result in global return maximization~\cite{guttman1998cooperative,wang2018competitive}. An SG can also be a combination of the aforementioned settings, for instance in the game of soccer where agents are cooperative and competitive at the same time~\cite{stone2005reinforcement,riedmiller2009reinforcement,asali2016using}.

We classify FireCommander  as a cooperative stochastic game. In the FireCommander environment, there are two categories of agents: (1) Perception (e.g., sensing or fire-observing) and (2) Action (e.g., manipulator or firefighting) agents. We note that these agents are uni-task and can only perform their specific task. The objective of the game is to maximize the firefighting reward by cooperatively putting out all of the dynamic, propagating firespots and keep the facilities (e.g., house, hospital, etc.) on the map safe from the fire. Nevertheless, firespots are initially invisible. To fight the fire, one must first find it (e.g., the spots of fire) on the map. We note that, Action-agents cannot fight the fire, unless the fire is first sensed by a Perception-agent, and Perception-agent cannot put out the fires without an Action-agent fighting the fire after it is sensed. As such, the FireCommander game environment is cooperative, since agents must cooperate to accomplish a common, complex mission. We note that, we introduce the term \textit{composite teams} to refer to our FireCommander environment in which agents are co-dependent on accomplishing an overarching mission~\cite{seraj2020hierarchical}.

\subsubsection{Combinatorial Optimization Problems}
\label{subsubsec:CombinatorialOptimization}
\noindent An optimization problems (e.g., the problem of finding the best solution to a maximization or minimization problem) can generally be divided into two categories: (1) continuous variable optimization and, (2) discrete variable optimization problem (OP). A discrete variable optimization is also known as a \textit{combinatorial} optimization problems (COP)~\cite{papadimitriou1998combinatorial,wolsey1999integer,schrijver2003combinatorial}. In the continuous OPs, solutions are a set of real numbers, while in a COP, solutions are objects, such as integers, permutations, graphs, etc., from a large finite set (or countable infinite)~\cite{sghir2018multi,papadimitriou1998combinatorial,wolsey1999integer,schrijver2003combinatorial}. Two simple but well-known examples of typical COPs are the traveling salesman problem (TSP) and the mixed linear integer programming (MILP)~\cite{papadimitriou1998combinatorial,gombolay2013fast,wolsey1999integer,schrijver2003combinatorial,Gombolay:2017a}:
\begin{itemize}
    \item \textbf{TSP:} given the (x, y) positions of N different cities, find the shortest possible path that visits each city exactly once.
    
    \item \textbf{MILP:} maximize a specified linear combination of a set of integers $ \{x_1, x_2, \cdots, x_n\} $ subject to a set of linear constraints of the form $ a_1x_1 + \cdots + a_nx_n \leq c $
\end{itemize}

Many multi-agent planning, scheduling, resource allocation and task assignment problems are categorized as COPs, since the solution to these problems typically consists of finding an optimal object from a finite set of objects~\cite{papadimitriou1998combinatorial,wolsey1999integer,schrijver2003combinatorial}. In many COPs, exhaustive search (e.g., brute force) is not tractable as the space of possible solutions is typically too large. In some cases, problems can be solved exactly using Branch and Bound techniques~\cite{marinescu2009and,papadimitriou1998combinatorial}. However, in other cases no exact algorithms are feasible, and randomized search algorithms must be employed~\cite{papadimitriou1998combinatorial,wolsey1999integer,schrijver2003combinatorial}.

In FireCommander , Perception and Action agents must be coordinated to cooperatively fight propagating fires. Doing such however, rises various questions such as:
\begin{itemize}
    \item \textit{How perception agents should select where to go?}
    
    \item \textit{How action agents should divide up the tasks received?}
    
    \item \textit{Which perception agent goes to explore somewhere in a map that has not been explored yet?}
    
    \item \textit{Which action agent should start a task, given a set of constraints such as battery-life, distance, tasks on the queue, etc?}
    
    \item \textit{How perception agents should balance between exploring new ares of the map versus exploiting the firespots they already found (e.g., spend time to communicate the sensed information to an action agent)?}
    
    \item \textit{How often should perception agents re-visit previously detected firespots to update their information?}
    
    \item \textit{What is an optimal path for an agent to take towards a goal/task?}
    
    \item \textit{How agents should be distributed between tasks, areas of the map, etc., given we have limited resources?}
    
\end{itemize}
All of the above questions, to some extent, define a constraint satisfaction problem (CSP), solving which falls within the realms of COPs. As such, our FireCommander  environment can be a complex testbed environment for many challenging combinatorial optimization problems.

\subsubsection{Multi-agent Reinforcement Learning (RL)}
\label{subsubsec:ReinforcementLearning}
\begin{figure}[t!]
\centering
\includegraphics[width=\textwidth]{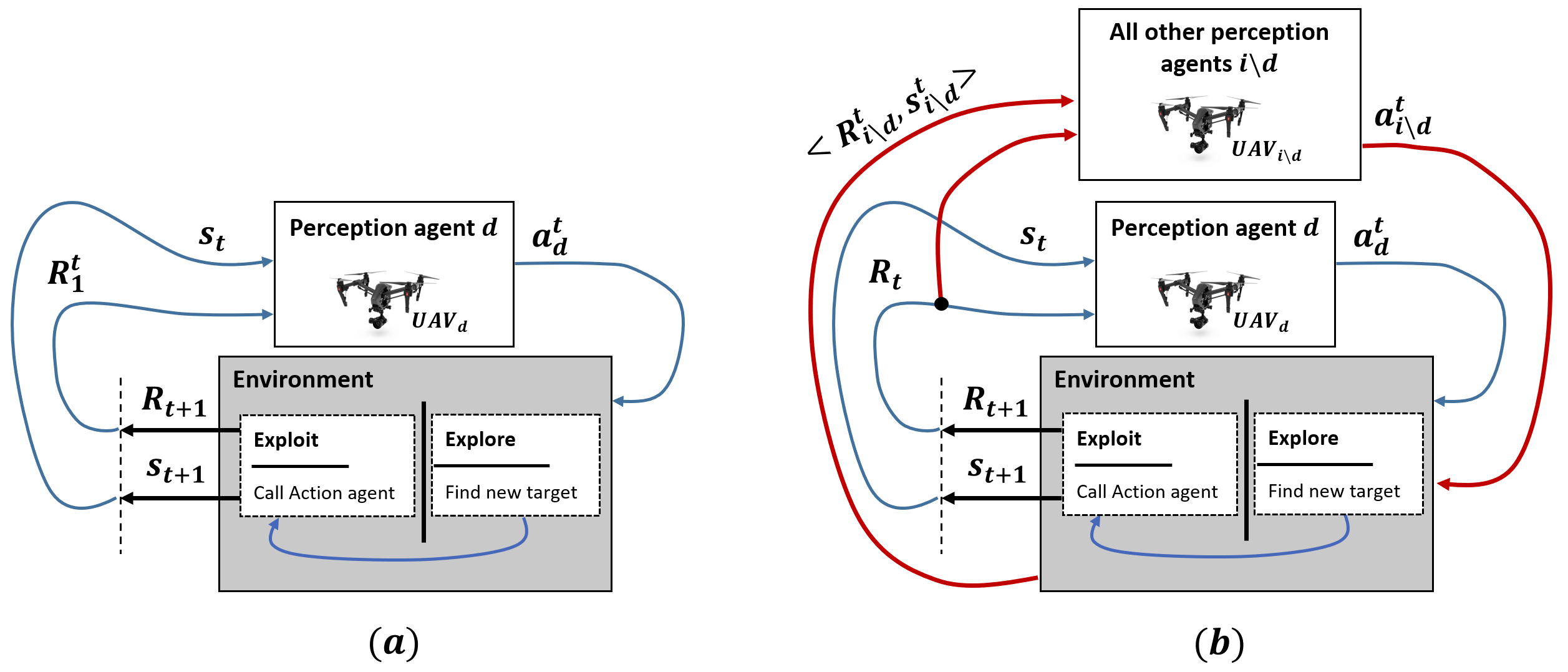}
\caption{Left-side figure represents the single-agent reinforcement learning (RL) problem  in which an agent interacts with an environment by taking an action $a$ at state $s$ and receiving a reward $R$. Right-side figure presents the multi-agent reinforcement learning (MARL) problem  in which a group of agents interact with the environment. Note that here, the reward of taking a rewarding action by one agent is also received by all other agents that did not take that action, since the agents are cooperating and the environment state-change occurs for all agents, regardless of the actor. The objective in both RL and MARL is to maximize a cumulative discounted return.}
\label{fig:RLModels}
\end{figure}

\noindent Figure~\ref{fig:RLModels} represents the single-agent reinforcement learning (RL) and the multi-agent reinforcement learning (MARL) problems. Left-side figure in Figure~\ref{fig:RLModels}, represents the single-agent RL in which an agent interacts with an environment by taking an action $a$ at state $s$ and receiving a reward $R$. Right-side figure in Figure~\ref{fig:RLModels}, presents the MARL problem in which a group of agents interact with the environment. Note that here, the reward of taking a rewarding action by one agent is also received by all other agents that did not take that action, since the agents are cooperating and the environment state-change occurs for all agents, regardless of the actor. The objective in both RL and MARL is to maximize a cumulative discounted return.

In FireCommander, we are dealing with a MARL problem. However, due to the unique characteristics of the FireCommander environment, such as the dynamicity of the environment (e.g., constantly state-changing firespots) and a set of initially invisible firespots, the problem resembles \textit{restless decision making} problem~\cite{le2008multi}. In our restless decision-making problem, regardless of the collective decisions of the perception-action team, the states of all observed or unobserved targets are constantly evolving through time. Moreover, as long as a fire spots is not being observed by a perception agent, the team does not have any information about the state of the fire and the uncertainty about its states keep increasing. 

This problem can be considered as a variant of POMDP. Accordingly, our objective is to find an optimal policy, $ \pi^* $, over all admissible policies in set, $ \pi\in \Pi $, that maximizes the expected, time-discounted reward collected by all perception agents over an infinite horizon, as in Equation~\ref{eq:restlessObjective}~\cite{seraj2020hierarchical}.
\begin{equation}
	\label{eq:restlessObjective}
	\pi^* = \argmax_{\pi\in \Pi}\mathbb{E}_\pi\left[ \sum_{t=0}^{\infty}\gamma^t\left(r_t^{P1}+\cdots+r_t^{PN_P}\right)\right]
\end{equation}

\subsubsection{Multi-agent Learning from Demonstrations: Inverse RL}
\label{subsubsec:InverseReinforcementLearning}
\noindent An optimal (or near-optimal) policy, enables a robot to select a wise action based upon the current states of the world which leads to a maximized reward. The development of such policies by hand is often very challenging and as a result machine learning techniques such as RL have been applied to policy development~\cite{argall2009survey,brys2015reinforcement}. RL works based on agent's experience through interacting with the environment and often, the more experience-data an agents achieves, the better policy it can generate. Nevertheless, acquiring enough experience data for an RL algorithm to perform well is a limiting factor due to its need for an often prohibitively large number of environment samples before the agent reaches a desirable level of performance~\cite{argall2009survey,brys2015reinforcement}. As such, LfD algorithms have been proposed to solve the agent's need for exhaustive search over the state-action space~\cite{gombolay2016apprenticeship,argall2009survey,soans2020sa}.

\begin{figure}[t!]
\centering
\includegraphics[width=\textwidth]{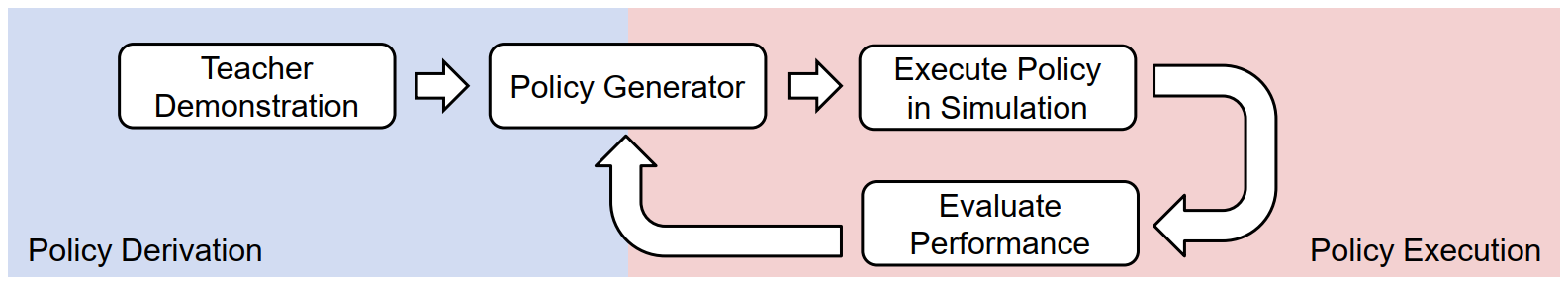}
\caption{This figure represents the Learning from Demonstration (LfD) framework.}
\label{fig:LfD}
\end{figure}

Figure~\ref{fig:LfD} represents the Learning from Demonstration (LfD) framework. In LfD, a policy is learned from demonstrations (e.g., sample trajectories), provided by a supposed expert teacher~\cite{bertsimas2000restless}. Sample trajectories are defined as sequences of state–action pairs that are recorded during the expert teacher’s demonstration of the desired behavior. LfD algorithms utilize produced the trajectory dataset to derive a policy that reproduces the demonstrated behavior (Figure~\ref{fig:LfD}). Obtaining a policy in this way is in contrast to RL techniques in which a policy is learned from exploration and experience~\cite{argall2009survey,brys2015reinforcement}. LfD is specifically of interest when the reward function structure is unknown.

\begin{itemize}
    \item \textbf{Multi-agent Learning from Heterogeneous Demonstrations (MA-LfHD)--} FireCommander  inherently includes two categories of heterogeneous robots: (1) Perception (e.g., Sensing) agents and (2) Action (e.g., Manipulator) agents, and is a multi-objective game. Missions and games in FireCommander do not have unique solutions and thus, can be interpreted and interacted with in various different ways by a user and thus, can be a perfect test-bed for developing learning from \textit{heterogeneous demonstrations} algorithms~\cite{paleja2019interpretable,chen2020joint,paleja2019heterogeneous,song2018multi,yu2019multi,vsovsic2016inverse,paleja2019inferring,chen2020learning}.
\end{itemize}

\subsubsection{Perception-Action Networks}
\label{subsubsec:Perception-ActionNetworks}
\noindent Robot intelligence requires a real-time connection between sensing and action~\cite{lee1997sensor}. This fact directly extends to multi-agent systems, such as the FireCommander environment in which in order to achieve \textit{team intelligence}, the perception and action robot teams must communicate to accomplish a complex overarching mission. Communication channels between agents form a network and thus, communicating perception and action teams, forms a Perception-Action Network.

Perception-Action Networks (PAN) can be static or dynamic~\cite{sabor2017comprehensive,ramasamy2017mobile}. One famous instance of dynamic PANs are Mobile Sensor Networks (MSN) which consist of a group of perception agents (or sometimes even hybrid sensing-acting agents) that can move around an environment to search for their targets of interest (e.g., firespots in wildfire fighting)~\cite{elhabyan2019coverage,sabor2017comprehensive,hu2004localization}. MSNs are specifically useful for covering large environments with limited number of agents and for field coverage problems in which the environment is dynamic (e.g., dynamic field coverage)~\cite{elhabyan2019coverage,poduri2004constrained,seraj2020coordinated,seraj2019safe,seraj2020hierarchical}.

\subsubsection{Wireless Sensor and Actor Networks}
\label{subsubsec:WirelessSensorActorNetworks}
\noindent Wireless sensor and actor networks (WSAN) refer to a group of communicating perception and action agents (e.g., PANs) which are used to perform distributed sensing and acting tasks~\cite{akyildiz2004wireless}. In WSANs, sensors gather information about the physical world (e.g., similar to Perception agents in FireCommander environment), while actors take decisions and then perform appropriate actions upon the environment (e.g., similar to Action agents in FireCommander). Similar to the FireCommander game in which users remotely use perception and action agents to fight the fire, WSANs allow a user to effectively sense and act from a distance~\cite{akyildiz2004wireless}. In order to provide effective sensing and acting, coordination mechanisms are required among sensors and actors~\cite{akyildiz2004wireless,melodia2005distributed,melodia2007communication}. 

An important note to be considered in WSANs is that, to perform correct and timely actions, sensing data must be valid at the time of acting~\cite{akyildiz2004wireless}. This is particularly important in dynamic environments such as the FireCommander  environment. This is because firespots are propagating (e.g., moving) and their position can only be seen when they are within the field-of-view (FOV) of a Perception agent. As such, when a perception agent observes a firespot, simply passing the spot's current position to an action agent and then leaving to find another firespot will not work, since while the action agent is on its way the firespot will move to a new location and also, action agents cannot sense and thus, cannot see the firespot's new location. Accordingly, the exploration/exploitation dilemma becomes a more challenging problem for effectively coordinating the Perception-Action agents in a composite team~\cite{seraj2020hierarchical}. Moreover, WSANs might include static sensors and actors, static sensors and dynamic actors or vice versa. FireCommander environment can easily be modified to match these settings while other existing games and environments lack such flexibility.

\subsubsection{Learning to Cooperate: Coordination and Communication Learning}
\label{subsubsec:CoordinationCooperationCommunicationLearning}
\noindent As discussed earlier in Sections~\ref{subsubsec:Multi-agentSystems} and~\ref{subsubsec:WirelessSensorActorNetworks}, in a multi-agent system, in order to accomplish the complex mission (or maximize a time-discounted final return) agents must collaborate. Effective collaboration/cooperation requires a \textit{robust}, \textit{accurate}, \textit{efficient} and \textit{scalable} coordination mechanism and, such coordination among agents, requires communications, which is constrained by many limiting factors. As a solution to all of the mentioned coordination and communication problems, RL, specifically when combined with deep learning (Deep RL\footnote{Despite neural networks (NN) being large and requiring massive computational power, Deep RL algorithms are in fact becoming more and more popular in online real-world robotics through applying neural network pruning and deep compression and quantization algorithms such as~\cite{karimzadeh2019hardware,han2015deep,karimzadeh2020hardware}}) and function approximation, has been widely studied in recent years~\cite{foerster2016learning,sukhbaatar2016learning,zhang2013coordinating,ghavamzadeh2004learning,jiang2018learning,kim2019learning,mao2017accnet,zhang2019efficient,asali2020deepmsrf}. These frameworks intend to enable the agents in a team to "\textit{learn how to cooperate}".

Multi-agent reinforcement learning (MARL) provenly performs better when there is communication among participating agents allowing them to coordinate their actions for maximising their shared utility. However, information sharing among miscellaneous agents (of different types and capabilities) can lead to heterogeneous communication protocols. When communication protocols include such miscellaneous information, agents of different types might not be able to differentiate the heterogeneity in the globally sent and received massages and extract valuable information that helps cooperative decision making. Therefore, communication may become unhelpful, and could even deteriorate the multi-agent cooperation learning. FireCommander game not only covers almost all of the mentioned coordination and communication problems, it also adds some challenging twists such as:
\begin{itemize}
    \item Complex, multi-objective environment
    
    \item Partially observable and dynamic environment
    
    \item Constrained Communication
    
    \item Uni-task robots
\end{itemize}
As such, FireCommander can be considered as a good, challenging testbed for learning multi-agent cooperation and developing end-to-end differentiable communication protocols for heterogeneous and composite teams~\cite{seraj2020hierarchical}.

\subsubsection{Multi-robot Planning/Scheduling and Task Assignment (MRTA)}
\label{subsubsec:Multi-robotPlanningSchedulingTaskAssignment}
\noindent Many multi-agent planning, scheduling, resource allocation and task assignment problems are categorized as COPs, since the solution to these problems typically consists of finding an optimal object from a finite set of objects~\cite{papadimitriou1998combinatorial,wolsey1999integer,schrijver2003combinatorial}. In many COPs, exhaustive search (e.g., brute force) is not tractable as the space of possible solutions is typically too large. In some cases, problems can be solved exactly using Branch and Bound techniques~\cite{marinescu2009and,papadimitriou1998combinatorial}. However, in other cases no exact algorithms are feasible, and randomized search algorithms must be employed~\cite{papadimitriou1998combinatorial,wolsey1999integer,schrijver2003combinatorial}.

Nevertheless, search algorithms are time-consuming, require significant computational power and resources and can easily get intractable when the problem dimensions increases. Accordingly, similar to many other optimization problems, COPs, such as dynamic scheduling, have also been tried to be solved by RL algorithms combined with deep learning (Deep RL) and Graph Neural Networks (GNN)~\cite{wang2019learning,kim2019learning,wang2020learning}. Here, learning frameworks intend to enable the agents in a team to "\textit{learn the scheduling or planning policies}"~\cite{wang2019learning,wangheterogeneous}.

As described in Section~\ref{subsubsec:CombinatorialOptimization}, in FireCommander, Perception and Action agents must efficiently coordinate to cooperatively fight propagating fires. Doing such however, requires solving various combinatorial optimization problems (COPs) such as ``\textit{How should agents choose where to go?}", or ``\textit{How should agents divide up the tasks?}", or ``\textit{How should agents be distributed among tasks, areas of the map, etc., given we have limited resources?}", and more. Solving for these questions falls within the realms of combinatorial optimization (see Section~\ref{subsubsec:CombinatorialOptimization}). Our FireCommander environment can be a complex environment for many challenging COPs such as dynamically learning scheduling policies~\cite{wang2019learning,wang2020learning}, and multi-agent task assignment~\cite{ravichandar2019strata}.

\subsubsection{Human-Robot Interaction (HRI), Human-Robot Teaming and Psychology}
\label{subsubsec:Human-RobotInteraction}
\noindent FireCommander includes an interactive game-based graphical user interface (GUI) which can be of great use in HRI researches. As an instance, the FireCommander game can be leveraged to design environments with heavy/light workload and then test how an expert’s policy design efficiency and quality is affected under situational stress. Various other HRI topics can be similarly modeled to leverage FireCommander as their test-bed, be such as trust and accountability~\cite{natarajan2020effects,heyer2010human}, anthropomorphism~\cite{zlotowski2015anthropomorphism,fink2012anthropomorphism,natarajan2020effects}, human-robot co-adaptation~\cite{gao2017personalised,gombolay2017computational}, human-guided optimization~\cite{gombolay2018human,schaff2020residual,gombolay2015decision,jevtic2018robot,huttenrauch2017guided,gombolay2015coordination} and many more~\cite{goodrich2008human,kolling2013human}.

\subsection{FireCommander: Game Structure and Dependencies}
\label{subsec:InverseReinforcementLearning}
\subsubsection{Functions, Python Packages and Dependencies}
\label{subsubsec:FunctionsPythonPackageDependencies}
\noindent Once you have cloned/downloaded the toolbox, 10 files and folder as represented in following Table shall be appeared in your chosen directory. Table (1) represents all Python files (i.e. core functions and internal computational functions) that come with this toolset and a short descriptions of each. Additionally, a copy of both GNU general public license and this user manual are also included.
\begin{table}[h!]
\centering
\label{tab:AllOthermfiles}
\caption{FireCommander Python files and assets (MARL and LfD/HRI library packages)}
\begin{tabular}{ |c|c||c| }
\hline
\multicolumn{3}{|c|}{\textbf{** Multi-agent LfD and HRI Package **}} \\
\hline
\hline
\textbf{row} & \textbf{name (function)} & \textbf{description} \\ 
 \hline
 \hline
 1 & \tt{FireCommander.py} & Main script to run the FireCommander GUI \\ 
 \hline
 2 & \tt{Utilities.py} & Inner functions and assets required to run the environment \\ 
 \hline
 3 & \tt{WildFire\_Model.py} & Script for FARSITE wildfire propagation model \\
 \hline 
 4 & \tt{Scenario\_Mode\_Params.py} & Script including the Scenario Mode Parameters \\ 
 \hline
 5 & \tt{Demo\_Visualization.py} & Script for deploying an existing teacher demo (data) \\
 \hline
 \hline
 \textbf{row} & \textbf{name (folder)} & \textbf{description} \\
 \hline
 \hline
 6 & \tt{Open\_World\_Data} & Contains recorded user-data for Open-World Mode \\
 \hline
 7 & \tt{Scenario\_Data} & Contains recorded user-data for Scenario Mode \\
 \hline
 8 & \tt{Images} & Contains images required to run the GUI \\
 \hline
 \hline
 \multicolumn{3}{|c|}{\textbf{** Multi-agent RL Package **}} \\
 \hline
 \hline
 1 & \tt{FireCommander\_Base.py} & Homogeneous version of FireCommander for MARL  \\ 
 \hline
 2 & \tt{FireCommander\_Cmplx1.py} & Heterogeneous version of FireCommander (Cooperative) \\ 
 \hline
 3 & \tt{FireCommander\_Cmplx2.py} & Full Heterogeneous version of FireCommander for MARL \\
 \hline 
 4 & \tt{Utilities.py} (per level) & Inner functions and assets required to run the environments \\ 
 \hline
 5 & \tt{WildFire\_Model.py} & Script for FARSITE wildfire propagation model \\
 \hline
\end{tabular}
\end{table}

\noindent The codes are written in Python 3.6.4 and PyGame 1.9.6 is used for designing the GUI. The current versions have been tested and as a routine, any later version of these softwares should also be compatible with our code. Using older versions of Python such as Python 2.7 (or other earlier versions) are not recommended.

\subsubsection{FARSITE: Wildfire Propagation Mathematical Model}
\label{subsubsec:FARSITE}
\noindent We leverage the simplified Fire Area Simulator (FARSITE) wildfire propagation model~\cite{finney1998farsite, seraj2020coordinated,seraj2019safe,seraj2020hierarchical,seraj2020coordinatedslides}, where $ q_t^i $ indicates how firespot $ i $ propagates according to Eq.~\ref{eq:firedynamics01} along the X-Y coordinates and thus, $ q $ is a 2-dimensional vector of size $ 1\times2 $~\cite{seraj2020hierarchical}.
\begin{equation}
    q_t^i = q_{t-1}^i + \dot{q}_{t-1}^i\delta t
    \label{eq:firedynamics01}
\end{equation}
The above equation represents the fire motion model in which, $ \dot{q}_{t}^i = dq_t^i/dt $ is the motion dynamics and identifies a fire’s growth rate (i.e., fire propagation velocity) and is a function of fuel and vegetation coefficient ($ R_t $), wind speed ($ U_t $), and wind azimuth ($ \theta_t $). The first-order firespot dynamics, $ \dot{q}_{t} $, can be estimated for each propagating spot by Equation~\ref{eq:qdot1}, where $\mathcal{D}=\sin(\theta_t)$ and $\mathcal{D}=\cos(\theta_t)$ along \textit{X} and \textit{Y} axes, respectively~\cite{finney1998farsite}.
\begin{equation}
	\dot{q}_{t} = C(R_t, U_t)\mathcal{D}(\theta_t) \label{eq:qdot1}
\end{equation}In Equation~\ref{eq:qdot1}, the spread rate $ C(R_t, U_t) $ can be calculated as in Equation~\ref{eq:C}, in which $ LB(U_t) = 0.936e^{0.256U_t} + 0.461e^{-0.154U_t} - 0.397 $ and $ GB(U_t) = LB(U_t)^2 - 1 $.
\begin{equation}
	\label{eq:C}
	C(R_t, U_t) = R_t\left(1-\frac{LB(U_t)}{LB(U_t) + \sqrt{GB(U_t)}}\right)
\end{equation}
\begin{figure}[t!]
\centering
\includegraphics[width=\textwidth]{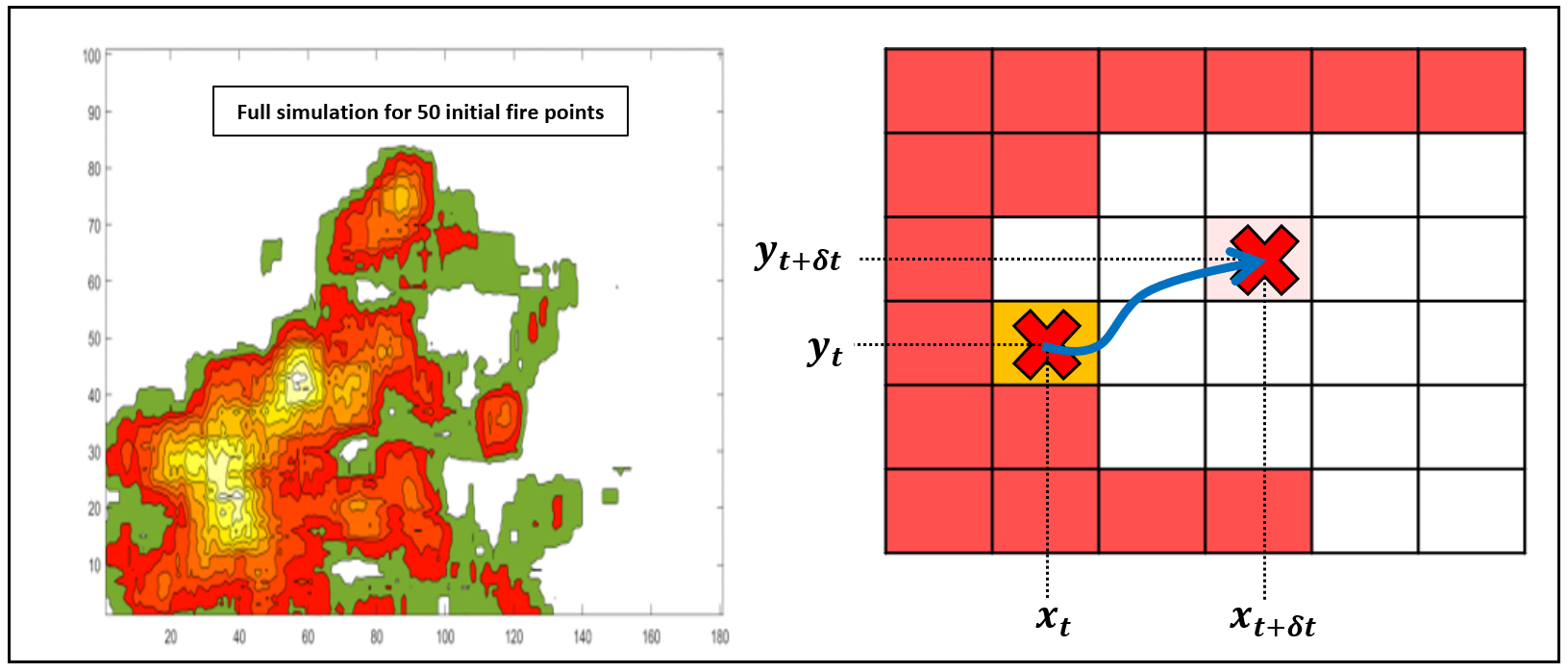}
\caption{Wildfire simulations using FARSITE~\cite{finney1998farsite} fire propagation model. Left: simulation with 50 initial fire points. Right: simple grid visualization of the fire propagation based on the Equation~\ref{eq:firedynamics01}.}
\label{fig:FireModel}
\end{figure}

Moreover, to account for the fire intensity (measured in kilo-watts per meter, $ [\frac{KW}{m}] $), we leverage the method proposed by~\cite{alexander1982calculating} which utilizes flame height $ h_t^q $ (measured in meters) and flame tilt angle $ \alpha_t^q $ with respect to vertical horizon line (measured in degrees) for each fire-spot $ q $ at time $ t $ to calculate the fire intensity $ I_t^q $, as follows in Eq.~\ref{eq:intensity}~\cite{seraj2020hierarchical}.
\begin{equation}
	\label{eq:intensity}
	I_t^q = 259.833\left(\frac{h_t^q}{\cos\left(\alpha_t^q\right)}\right)^{2.174}
\end{equation}While fire is propagating, we assume each fire-spot radiates heat according to a Gaussian distribution and the intensities of adjacent points are linearly summed if the two points are within a predefined radiation range, which is in accordance with prior studies~\cite{seraj2020coordinated,seraj2019safe,seraj2020hierarchical}. Figure~\ref{fig:FireModel} demonstrates the wildfire simulations using FARSITE~\cite{finney1998farsite} mathematical fire propagation model. Furthermore, due to the fuel exhaustion, we model the intensity decay of a fire spot during its ignition time $ \delta t_q $, as a dynamic exponential decay rate $ \lambda $ over time, presented below in Equation~\ref{eq:intensityDecay}~\cite{seraj2020hierarchical}.
\begin{equation}
	\label{eq:intensityDecay}
	I_{t+\delta t}^q = I_{t}^q\left(e^{-\lambda\frac{\delta t_q}{R_t}}\right)
\end{equation}Finally, when a firefighting UAV drops the extinguisher fluid over an area of fire, we cut the fire intensities of the respective fire spots according to a predefined extinguisher fluid coefficient. A fire point is pruned from the fire-map if its intensity falls below a threshold value, leaving a burnt spot on the terrain map which cannot catch fire anymore~\cite{seraj2020hierarchical}.

\subsubsection{Game Structure}
\label{subsubsec:GameStructure}
\noindent As mentioned earlier there are two main library packages of FireCommander provided. First, we developed various ready-to-use versions of the Fire Commander which can be directly leveraged to test multi-agent reinforcement learning (MARL) algorithms. Second, we also developed a graphical user interface (GUI) which can be used to record expert data for learning from demonstration (LfD) and human-robot interaction (HRI) studies dealing with multi-agent, heterogeneous coordination/cooperation and communication Learning.

We classify FireCommander as a cooperative stochastic game (see Section~\ref{subsubsec:Multi-agentSystems}) in which all agents share a common objective and reward function, although the full complex version can also be considered as a non-cooperative Markov game which includes different objectives and rewards for different groups od agents. The environment includes uni-task and multi-task agents. The objective of the game is, given a limited time and number of heterogeneous agents, maximize the firefighting reward by (1) cooperatively putting out all propagating firespots and (2) keeping the targets (e.g., facilities such as house, hospital, etc.) on the map safe from the fire. Firespots are initially invisible. To fight the fire, one must first find the firespots on the map. We note that Action-agents cannot fight the fire unless the fire is first sensed by a Perception-agent, and Perception-agent cannot put out the fires without an Action-agent fighting the fire after it is sensed. As such, the FireCommander game environment is cooperative, since agents must cooperate to accomplish a common, complex mission. We introduce the term \textit{composite teams} to refer to our FireCommander environment in which agents are co-dependent on accomplishing an overarching mission~\cite{seraj2020hierarchical}. Accordingly, the FireCommander  game environment and objectives can be summarized as follows:

\begin{itemize}
    \item \textbf{Game environment:} 
    \begin{itemize}
        \item \textit{Types of Agents:} Generally, In the FireCommander environment, there are three categories of agents: (1) Perception-agents (e.g., sensing or fire-observing), (2) Action-agents (e.g., manipulator or firefighting) and, (3) Hybrid-agents (e.g., agents that can both sense and manipulate). Most of the designed scenarios in the game (see Section~\ref{subsec:ScenarioMode}) are focused on coordinating between uni-task Perception and Action agents, however Hybrid agents that can do both tasks are also provided for generality of the environment. Agents can be homogeneous or heterogeneous in their parameters and motion characteristics such as velocity, safe altitudes, etc. See Section~\ref{subsec:Open-worldMode} for more details.
        
        \item \textit{Types of Targets/Facilities:} There are six different types of targets and facilities in the game: (1) agent depot, (2) power station, (3) hospital, (4) house, (5) road and (6) lakes. Fire can propagate to any of these targets which results in an increased rate of negative-reward received, respective to the type of the target. Each target has a pre-designed size on the map. See Section~\ref{subsec:Open-worldMode} for more details.
        
        \item \textit{Firespots Details:} Firespots are initially invisible to the user and must be discovered by a Perception agent (e.g., or a Hybrid agents). Fire is initialized in multiple spots and propagates through time and space according to the introduced FARSITE model (see Section~\ref{subsubsec:FARSITE}). Moreover, fire can start at anytime during the game and anywhere on the map (but not on any of the existing targets). See Section~\ref{subsec:Open-worldMode} for more details.
        
        \item \textit{Stochasticity in Environment:} All agents actions change the states of the environment stochastically such that: (1) Action-agents can put out the fires in their FOV according to a random probability distribution, which is designed through a confidence level coefficient. (2) We model a perception agent's altitude-dependent sensing quality (e.g., the lowest the altitude, the highest the quality of estimation) such that, perception-agents can sense (e.g., detect/locate) the fires within their FOV according to a random probability distribution which depends on their altitude. A perception agent at its lowest safe-altitude can sense 100\% of the firespots in its FOV while a perception agent at its highest allowable altitude can only sense 40\% of the firespots within its FOV.
        
        \item \textit{Battery-Life, Tanker-capacity and Motion Restrictions:} All agents have limited battery-life, upon exhaustion of which agents retreat to base depot for recharge. Moreover, Action and Hybrid agents have a limited capacity for dumping water on fire (e.g., tanker capacity). If the tanker is empty, agents will retreat to base depot for refuel. Moreover, in order to distinguish between the motion characteristics of Perception and Action agents, we assume that action agents are large, expensive devices that cannot be held inactive within the environment. As such, after driving an action agent to a location, the agent can only remain there for three seconds; afterwards, it will either retreat to base depot or will move to a new location specified by the user within the 3-second deadline. Perception agents can remain still in a location until their battery-life reaches zero.
    \end{itemize}

    \item \textbf{Game Objectives and Priorities:} 
    \begin{itemize}
        \item \textit{Game Objectives:} The objective of the game is, given a limited time and number of heterogeneous agents, maximize the firefighting reward by (1) cooperatively putting out all propagating firespots and (2) keep the targets (e.g., facilities such as house, hospital, etc.) on the map safe from the fire.
        
        \item \textit{Target Priorities:} If fire propagates on any of the targets, it results in an increased rate of negative-reward received, respective to the type of the target. As such, targets have different pre-defined priorities which can be sorted as follows: (1) agent depot, (2) power station, (3) hospital, (4) house, (5) road and (6) lakes, in which agent depot and power stations are the most important targets with -5 reward per firespot per time-step that an active fire is on the target. Moreover, priorities of lakes and roads are set to zero.
    \end{itemize}
\end{itemize}

\subsubsection{Interacting with the Game: Inputs \& Outputs}
\label{subsubsec:InteractingwiththeGameDataOutput}
\noindent FireCommander  is an interactive environment, designed such that it can be readily used for user-data acquisition in LfD research. The game environment can be interacted with and the resulting behaviours are saved on the computer in data-files, described as follows:
\begin{itemize}
    \item \textbf{Environment Inputs:} A user must select between multiple existing agents for sensing and firefighting tasks and then pan a trajectory to drive them to desired locations or areas. As such, the environment inputs can be summarized as follows:
    \begin{itemize}
        \item \textit{Agent Switch:} For this purpose digit keys on the keyboard are used. Agents are indexed $1-9$ each of which can be called by pressing the corresponding key number.
        
        \item \textit{Planar Motion Trajectory Planning:} A computer mouse is used for this purpose by which a User can click on any point on the map to drive the selected agent to that location. Planar trajectories in FireCommander  have two types: (1) Normal trajectories (Figure~\ref{fig:normalTraj}) and, (2) Patrolling trajectories (Figure~\ref{fig:ptrollingTraj}). Normal trajectories simply drive an agent to a desired location while in a patrolling trajectory, an agent keeps following that trajectory to ``\textit{patrol}" the specified area until it is told otherwise or its battery-life reaches zero.
        \begin{figure}[t!]
        \centering
            \begin{subfigure}{.45\textwidth}
              \centering
              % include first image
              \includegraphics[width=1\linewidth]{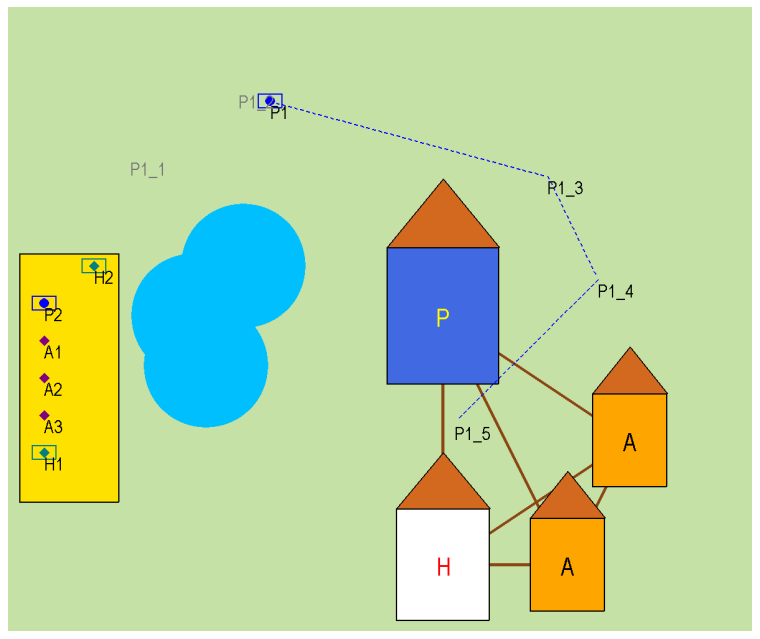}  
              \caption{An example normal trajectory}
              \label{fig:normalTraj}
            \end{subfigure}
            \begin{subfigure}{.45\textwidth}
              \centering
              % include second image
              \includegraphics[width=1\linewidth]{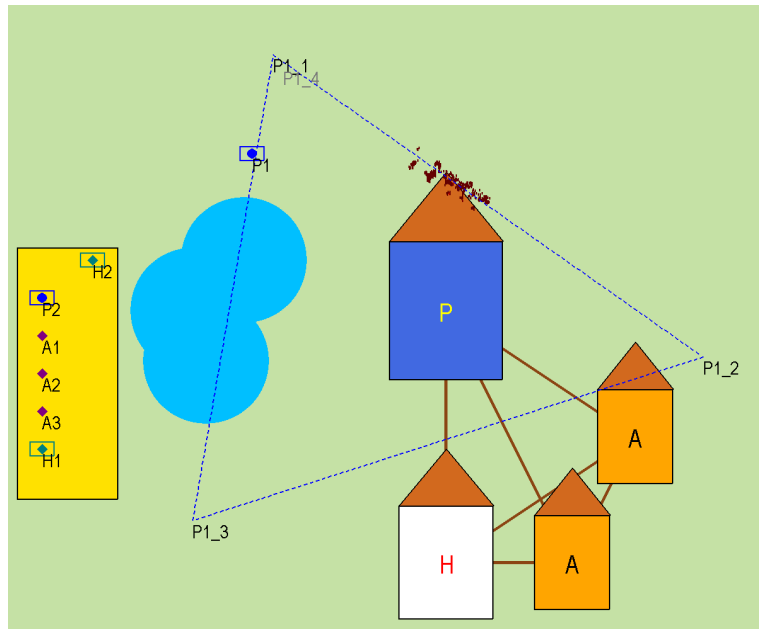}  
              \caption{An example patrolling trajectory}
              \label{fig:ptrollingTraj}
            \end{subfigure}
            \caption{Types of Planar trajectories in FireCommander.}
            \label{fig:typesoftraj}
        \end{figure}
        
    \item \textit{Vertical Motions:} Perception and Hybrid agents can increase their altitudes to gain more information (e.g., to see a larger area), or decrease their altitude to improve their information quality (e.g., to reduce estimation errors). As such, vertical motions are also required. To this end,  ``\textit{Up}" and ``\textit{Down}" arrow keys on the keyboard are used.
    \end{itemize}
    
    \item \textbf{Environment Outputs:} When one round of game is finished, 13 files with \textit{*.pkl}\footnote{A PKL file (e.g., \textit{*.pkl}) is a file created by pickle, a well-known Python module.} format are saved on the computer. The data-files include a comprehensive set of information regarding the game that was played, ranging from agents trajectories, fire information, target information and etc. The recorded information are ``\textit{complete}", in the sense that a \underline{replay video} of the exact game played by the user can be reconstructed from these information. See Section~\ref{subsec:DataFormats} and~\ref{subsec:AnimationReconstruction} for details of data formats and structures as well as the replay video reconstruction process.
\end{itemize}

\subsubsection{Game Feedback: Score Policy \& Performance Evaluation}
\label{subsubsec:GameFeedbackandScorePolicy}
\noindent Users receive an online score while playing the game and a general performance evaluation at the end of the game\footnote{Please note that most of the score policies and performance evaluation metrics are subject to change based on the underlaying application. For instance for some applications the online score might not be required, or a metric might be interesting in a setup that is not provided here, or the verbal performance evaluation might be needed to a fully quantitative system. As such, while we provided a somewhat general score and evaluation policy, users are encouraged to modify the online score and final performance evaluation policies according to their application of interest.}. Below, descriptions of both the online score policy and the performance evaluation metrics are provided (for more details and equations please see Section~\ref{subsec:OtherPages}):
\begin{itemize}
    \item \textbf{Online Score System:} The online score is shown on the information box during the game. The online score consists of various Negative and positive rewards, listed as follows:
    \begin{itemize}
        \item \textit{Negative Rewards:} User receives two types of negative rewards: (1) a constant, cumulative reward of -0.1 (negative reward) for each fire spot that is generated, per time-step, which is designed to encourage users to act fast. (2) When a fire propagates on any of the targets, it results in an increased rate of negative-reward received, respective to the target's priority. As such, targets have different pre-defined penalty coefficients (PC) sorted as follows: (1) agent depot  ($\text{PC}=-5$), (2) power station ($\text{PC}=-5$), (3) hospital ($\text{PC}=-2$), (4) house ($\text{PC}=-1$), (5) road ($\text{PC}=0$) and (6) lakes ($\text{PC}=0$). The penalty coefficients are applied per firespot per time-step that an active fire is on the target\footnote{The penalty coefficients can be changed if required.}.
        
        \item \textit{Positive Rewards:} User also receives three different positive rewards: (1) finding firespots (e.g., Perception Score), (2) putting out firespots (e.g., Action Score) and (3) keeping targets/facilities safe (e.g., Facility Protection Score).
    \end{itemize}

    \item \textbf{Final Performance Evaluation:} When a round of game ends, a final performance evaluation page is displayed (see Figure~\ref{fig:score_display} as an example) to the user. In this page, an ``\textit{Overall Firefighting Performance Score}" is reported as the ratio of number of firespots that have been put out over the total number of firespots in the map (observed and unobserved). Moreover, the perception and action scores as well as the facility protection score and the total negative reward from the online score panel are also reported on this page. Eventually, a "\textit{General Evaluation}" is reported as a verbal feedback, which is designed
    \begin{figure}[t!]
        \centering
        \includegraphics[width=\textwidth]{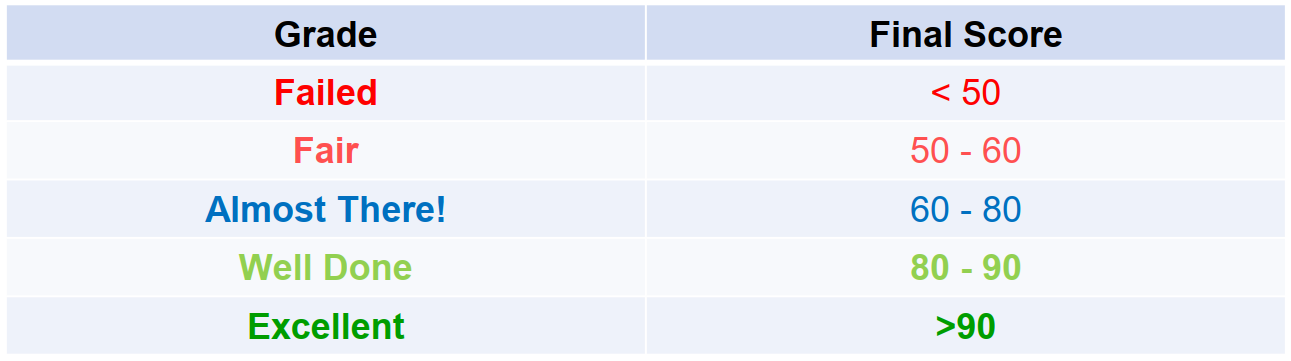}
        \caption{The ``\textit{General Evaluation}" verbal feedback system.}
        \label{fig:FireModel}
    \end{figure}
\end{itemize}

\subsection{How-Tos and Tips}
\label{subsec:How-Tos}
\subsubsection{How to Generate a Coordination Policy?}
\label{subsubsec:HowtoGenerateaCoordinationPolicy}
\noindent A coordination policy can take various forms based on the underlying application. However, generally speaking, in a multi-agent system a policy should include actions of all agents during the time of the game, given the states of the environment and the agents. In FireCommander environment we record all of the environment and agents' states as PKL files. As such, in order to generate a coordination policy the comprehensive set of recorded data can be used. See Section~\ref{subsec:DataFormats} for details.

\subsubsection{How to Deploy an Existing Coordination Policy?}
\label{subsubsec:HowtoDeployanExistingCoordinationPolicy}
\noindent When the coordination policy, as defined in the previous section, exists, it means that no interaction is required with the environment and we only wish to deploy the existing policy (as a set of data) to visualize the environments and the agents' behaviors. In FireCommander  an existing teacher policy (as a set of data) can be deployed on the environment (e.g., visualized) through the function ``\texttt{Demo\_Visualization.py}". See Section~\ref{subsec:AnimationReconstruction} for details of this process.

\subsubsection{How to Design a New Scenario?}
\label{subsubsec:HowtoDesignaNewScenario}
\noindent We have created 24 pre-designed scenarios in three categories of easy, moderate and hard. These scenarios are designed based on the requirements in our own multi-agent learning from heterogeneous demonstrations study [RF\footnote{Reference to be added.}] which can also be used in specific HRI and human-robot teaming experiments. As such, one might need to modify these scenarios based on the specific requirements in their own study. For this purpose, the design-parameters for these scenarios are gathered in a separate function, named "\texttt{Scenario\_Mode\_Params.py}", and can be accessed easily by users for further modifications. See Section~\ref{subsec:ScenarioMode} for more details on the pre-designed scenarios.

\subsubsection{How to Generate Expert Data for LfD Problems?}
\label{subsubsec:HowtoGenerateExpertDataforLfDProblems}
\noindent A user's use of clicks and keyboard buttons represents their ``\textit{expert}" actions. These actions can be put-together with the environment and the agents' states through time to generate a list of expert-actions and environment states. These state-action pairs in the created list form an example trajectory (e.g., demonstration), which is the input data to the LfD algorithm.

% % % % % % % % % % % % % % % % % % % % % % % % % % % % % % % % % % % % % % % % % % % % % % % % % % % % % % % % % % % % % % % % % % % % % % % % % % % % % % % % % % % % % % % % % % % % % % % % % % % % % % % % % % % % % % % % % %
\begin{figure}[t!]
    \centering
    \includegraphics[width=\textwidth]{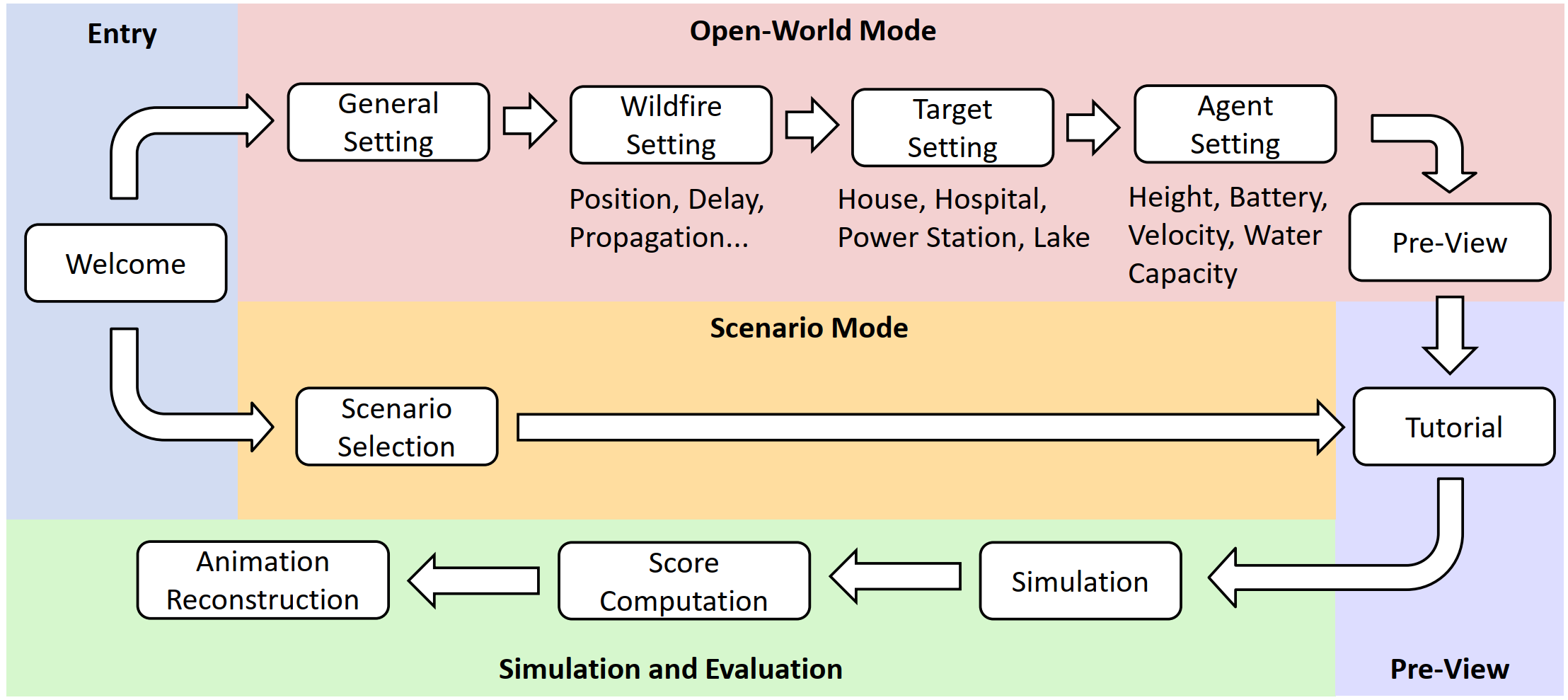}
    \caption{The Graphical User Interface (GUI) structure.}
    \label{fig:MA-SMDP}
\end{figure}

\section{GUI Reference Manual}
\label{sec:referencemanual}
\noindent The basic motivation for designing a graphical user interface (GUI) is to offer users a convenient way to enter existing scenarios or design their own scenarios. Moreover, in related HRI and Psychology studies, subjects need to interact with the game environment and apply changes online. To make the game design concise and interactive, our GUI helps the users design scenarios step by step, from the general number of each targets, to the their exact locations and orientations, as well as the operating parameters for some specific elements of the environment like agents and wildfire areas. 

Moreover, another purpose for the GUI is to easily collect subject (e.g., player) data. The pre-determined scenarios in the Scenario Mode are divided into three genres: \textbf{Easy}, \textbf{Moderate} and \textbf{Hard}, based on the overall complexity of the scenario. We also provide a simple practice scenario to help the novice player get accustomed to the GUI and controlling agents.

\subsection{GUI General Structure}
\label{subsubsec:GUIFrame}
\noindent Figure~\ref{fig:MA-SMDP} shows the general structure of the GUI. As the first page in the GUI, the welcome page offers a choice between the open-world mode, which allows the users to design a new scenario, and the scenario mode in which users play an existing scenario (e.g., similar to a mission in other common strategic games). For the open-world mode, the design process covers the following parts: 
\begin{itemize}
    \item \textbf{General Setting:} Number of each environment element (e.g., agents, firespots, facilities, etc.)
    
    \item \textbf{Wildfire Setting:} Locations and propagation parameters of each fire region. 
    
    \item \textbf{Target/Facilities Setting:} Locations and orientation (only for agent base) for each of facilities and targets, including the agent base, house, hospital, power station and lake. 
    
    \item \textbf{Agent setting:} Agents specific parameters and characteristics such as the flight height, battery, velocity and water capacity.
    
    \item \textbf{Preview:} Visualizing a static version of the designed scenario with the current parameters to user for final confirmation before starting the game.
\end{itemize}

%%%%%%%%%%%%%%%%%%%%%%%%%%%%%%%%%%%%%%%%%%%%%%%%%%%%%%%%%%%%%%%%%%%%%%%%%%%%%%%%%%%%%%%%%%%%%%
\begin{figure}[t!]
\centering
\includegraphics[width=0.7\textwidth]{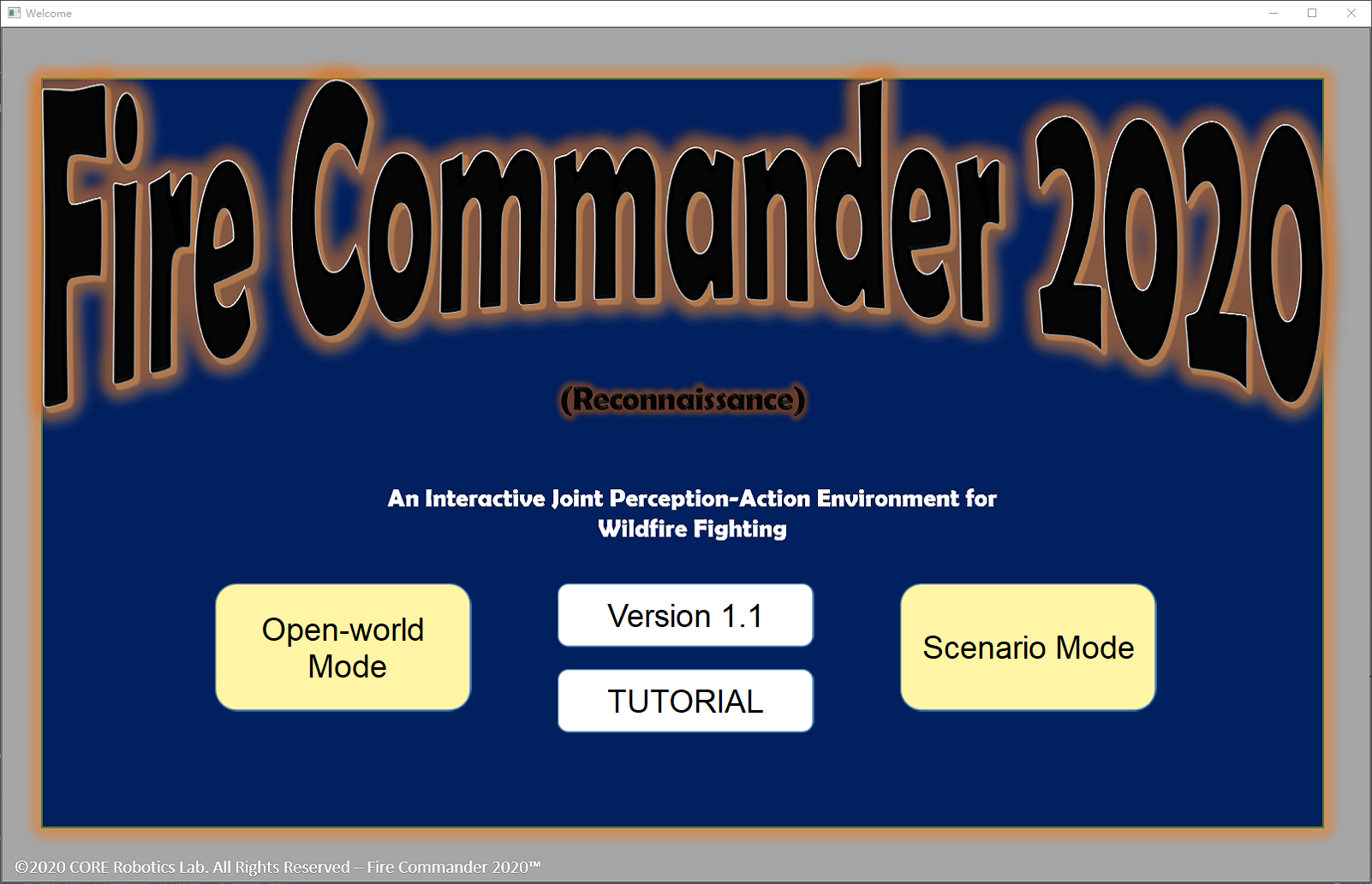}
\caption{The welcome screen.}
\label{fig:Welcome}
\end{figure}
%%%%%%%%%%%%%%%%%%%%%%%%%%%%%%%%%%%%%%%%%%%%%%%%%%%%%%%%%%%%%%%%%%%%%%%%%%%%%%%%%%%%%%%%%%%%%%

Users don't need to define parameters manually in the scenario mode. Players and subjects of the scenario mode can enter the game by choosing a scenario among a list of possible 24 pre-designed scenarios. After a game ends (either in open-world mode or scenario mode), a score-board of each trial is represented as a report of the user's performance in successfully accomplishing game's objectives. Users could choose to exit the game, to reconstruct a replay animation of their most recent game played as a video data, or to return to the welcome or scenario mode page.

\subsection{Welcome Screen}
\label{subsec:WelcomeScreen}
\noindent Figure~\ref{fig:Welcome} represents the welcome screen in FireCommander GUI. the welcome page offers a choice between the \underline{Open-World Mode}, which allows the users to design a new scenario, and the \underline{Scenario Mode} in which users play an existing scenario (e.g., similar to a mission in other common strategic games). Here are the details of the available options on the welcome screen:
\begin{itemize}
    \item \textbf{Scenario Mode:} Users will start the game directly with the existing parameters. The available scenarios include the 24 pre-designed scenarios, categorised from easy to hard, and a practice scenario. Player's data generated during playing any of the scenarios will be recorded. 
    
    \item \textbf{Open-world Mode:} Users can design their own scenarios by specifying necessary parameters for elements of the environment in a step-by-step procedure. Player's data generated during playing a user-designed scenario will also be recorded. 
    
    \item \textbf{Tutorial:} The tutorial page serves as a list of instructions on how to work with the GUI and play the game.
\end{itemize}

\subsection{Scenario Mode}
\label{subsec:ScenarioMode}
\begin{figure}[t!]
\centering
\includegraphics[width=0.7\textwidth]{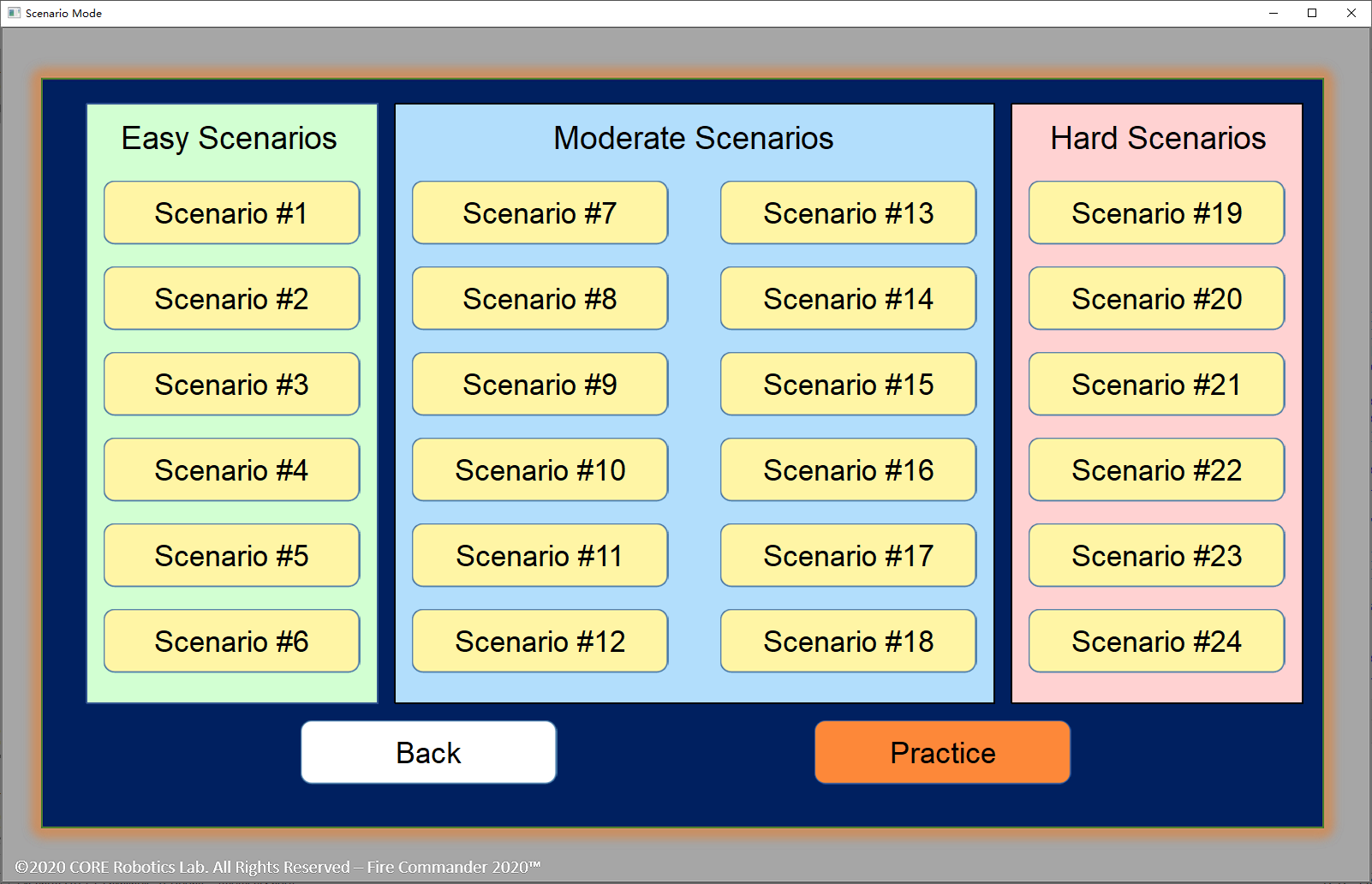}
\caption{The scenario mode screen.}
\label{fig:scenario_mode}
\end{figure}

\noindent Figure~\ref{fig:scenario_mode} shows the scenario mode screen. In the scenario mode, all the formal scenarios are divided into three difficulty groups based on the complexity of the scenario. The pre-designed scenarios are designed be diverse while also sharing some aspects in order to achieve both generalization and discrimination. Moreover, to help users practice and get used to the environments and game controls, we also provided a simple practice scenario. User's data while playing the practice scenario will not be recorded. Details of each of these 24 scenarios, as well as the practice scenario is presented in the following sub-sections.
% \begin{wrapfigure}[17]{r}{0.4\textwidth}
% 	\centering
% 	\includegraphics[width=0.4\columnwidth]{Practice.png}
% 	\caption{The Practice Scenario.}
% 	\label{fig:practice}
% \end{wrapfigure}

\paragraph{Practice Scenario:}
The purpose of the practice scenario is to help the novice users familiarize with the environment and game controls before attempting to play the pre-designed scenarios. The practice scenario incorporates the online score and final performance display, while the data storage function is disabled for this tutorial scenario. Once the game ends, users will return to the scenario mode page and their data and score for this trial will not be stored. Table~\ref{tab:practice_target} and~\ref{tab:practice_fire} present the target, agent and wildfire setting in the practice scenario.
\begin{table}[h!]
    \centering
    \begin{tabular}{|c||c|c|c|c||c|c|}
        \hline
        \multirow{2}{*}{\textbf{Scenario Index}} & \multicolumn{4}{|c||}{\textbf{\# of Targets}} & \multicolumn{2}{|c|}{\textbf{\# of Agents}} \\
        \cline{2-7} & House & Hospital & Power Station & Lake & Perception & Action \\
        \hline
        \hline
        \textbf{Practice} & 1 & 1 & 1 & 1 & 2 & 2\\
        \hline
    \end{tabular}
    \caption{The Targets and Agent Setting in the Practice Scenario}
    \label{tab:practice_target}
\end{table}
\begin{table}[h!]
    \centering
    \begin{tabular}{|c||c|c|c|c|c|c|}
        \hline
        \makecell[c]{\textbf{Scenario Index}} & \makecell[c]{\textbf{Region-wise} \protect\\ \textbf{\# of Fire Front}} & \makecell[c]{\textbf{Fire Delay} \\ \textbf{(Sec)}} & \makecell[c]{\textbf{Fuel} \\ \textbf{Coefficient}} & \makecell[c]{\textbf{Wind Speed}} & \makecell[c]{\textbf{Wind} \\ \textbf{Direction}} \\
        \hline
        \hline
        \textbf{Practice} & 15 & 0 & 10 & 5 & 45\\ 
        \hline
    \end{tabular}
    \caption{The Fire Setting in the Practice Scenario}
    \label{tab:practice_fire}
\end{table}

\paragraph{Formal Pre-designed Scenario Categories:} The formal scenarios are divided into easy, moderate and hard scenarios, based on the complexity of the scenario setting. Scenario categories are divided based on number of targets/facilities, number of agents and wildfire model parameters such as the firefront spots, fire delay, fuel coefficient, wind speed and wind direction\footnote{Wind direction is the counter-clockwise angle between the fire's moving direction and the vertical axis.}. Below, we elaborate the details for each scenario:
\begin{itemize}
    \item \textbf{Easy Scenarios (\#1 -- \#6):} Easy cenarios include the Scenario \#1 -- \#6, and their respective settings are presented in Figure~\ref{fig:easy_scenarios}. Tables~\ref{tab:easy_target} and~\ref{tab:easy_fire} present the target/agent and the fire settings for the easy scenarios. For the cases that multiple fire regions exist in the same scenario, parameters defining different fire regions are separated by a comma.
    \begin{figure}[t!]
        \centering
        \includegraphics[width=0.7\textwidth]{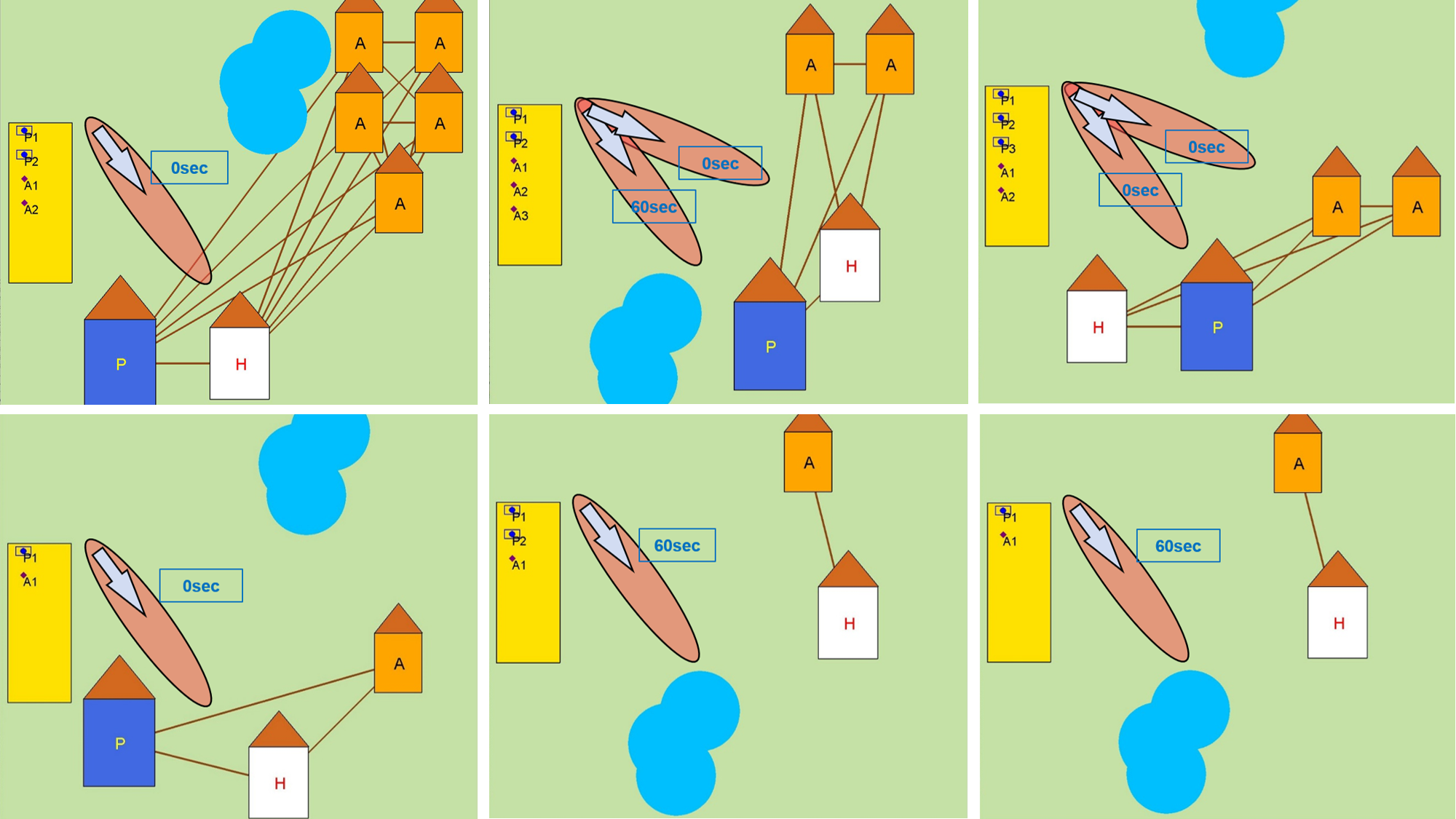}
        \caption{The Easy Scenarios (Scenario \#1 -- \#6)}
        \label{fig:easy_scenarios}
    \end{figure}
    \begin{table}[h!]
        \centering
        \begin{tabular}{|c||c|c|c|c||c|c|}
            \hline
            \multirow{2}{*}{\textbf{Scenario Index}} & \multicolumn{4}{|c||}{\textbf{\# of Targets}} & \multicolumn{2}{|c|}{\textbf{\# of Agents}} \\
            \cline{2-7} & House & Hospital & Power Station & Lake & Perception & Action \\
            \hline
            \hline
            \textbf{\#1} & 5 & 1 & 1 & 1 & 2 & 2\\
            \hline
            \textbf{\#2} & 1 & 1 & 0 & 1 & 2 & 1\\
            \hline
            \textbf{\#3} & 1 & 1 & 1 & 1 & 1 & 1\\
            \hline
            \textbf{\#4} & 1 & 1 & 0 & 1 & 1 & 1\\
            \hline
            \textbf{\#5} & 2 & 1 & 1 & 1 & 3 & 2\\
            \hline
            \textbf{\#6} & 2 & 1 & 1 & 1 & 2 & 3\\
            \hline
        \end{tabular}
        \caption{The Targets and Agent Setting in the Easy Scenario (Scenario \#1 -- \#6)}
        \label{tab:easy_target}
    \end{table}
    \begin{table}[h!]
        \centering
        \begin{tabular}{|c||c|c|c|c|c|c|}
            \hline
            \makecell[c]{\textbf{Scenario Index}} & \makecell[c]{\textbf{Region-wise} \protect\\ \textbf{\# of Fire Front}} & \makecell[c]{\textbf{Fire Delay} \\ \textbf{(Sec)}} & \makecell[c]{\textbf{Fuel} \\ \textbf{Coefficient}} & \makecell[c]{\textbf{Wind Speed}} & \makecell[c]{\textbf{Wind} \\ \textbf{Direction}} \\
            \hline
            \hline
            \textbf{\#1} & 10 & 0 & 10 & 5 & 45\\ 
            \hline
            \textbf{\#2} & 15 & 60 & 15 & 3 & 45\\ 
            \hline
            \textbf{\#3} & 5 & 0 & 15 & 5 & 45\\ 
            \hline
            \textbf{\#4} & 12 & 60 & 5 & 3 & 45\\ 
            \hline
            \textbf{\#5} & 3, 8 & 0, 0 & 5, 10 & 5, 3 & 45, 15\\ 
            \hline
            \textbf{\#6} & 5, 3 & 60, 0 & 10, 10 & 5, 5 & 45, 15\\ 
            \hline
        \end{tabular}
        \caption{The Fire Setting in the Easy Scenario (Scenario \#1 -- \#6)}
        \label{tab:easy_fire}
    \end{table}
    
    \item \textbf{Moderate Scenarios (\#7 -- \#18):} Moderate scenarios include the Scenario \#7 -- \#18, and their respective settings are presented in Figure~\ref{fig:moderate_scenarios1} and Figure~\ref{fig:moderate_scenarios2}. Table~\ref{tab:moderate_target} presents the facility/target and agent setting and Table~\ref{tab:moderate_fire} presents the fire setting for the moderate scenarios. For the cases that multiple fire regions exist in the same scenario, parameters defining different fire regions are separated by a comma.
    \begin{figure}[t!]
        \centering
        \includegraphics[width=0.7\textwidth]{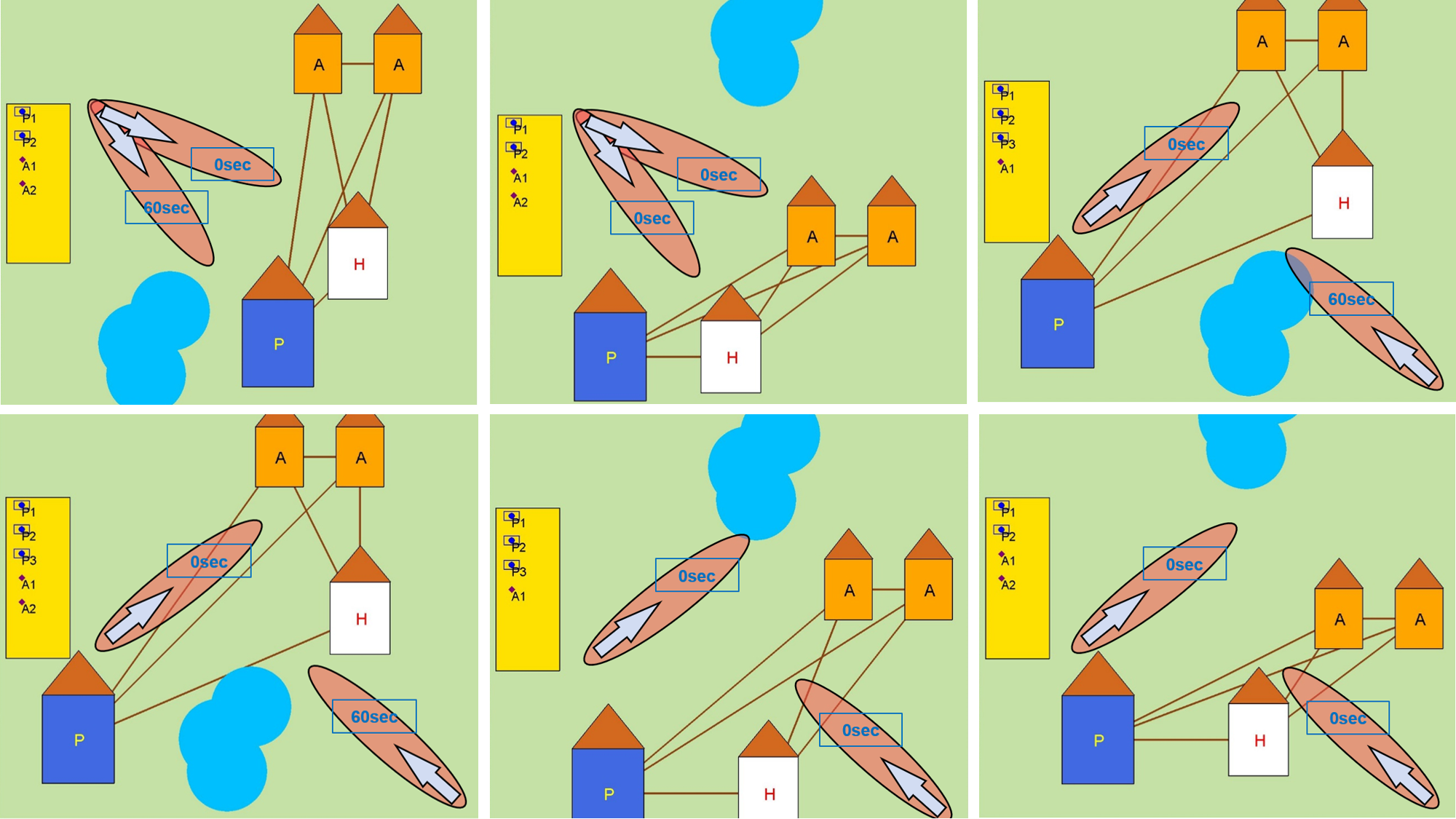}
        \caption{The first batch of Moderate Scenarios (Scenario \#7 -- \#12)}
        \label{fig:moderate_scenarios1}
    \end{figure}
    \begin{figure}[t!]
        \centering
        \includegraphics[width=0.7\textwidth]{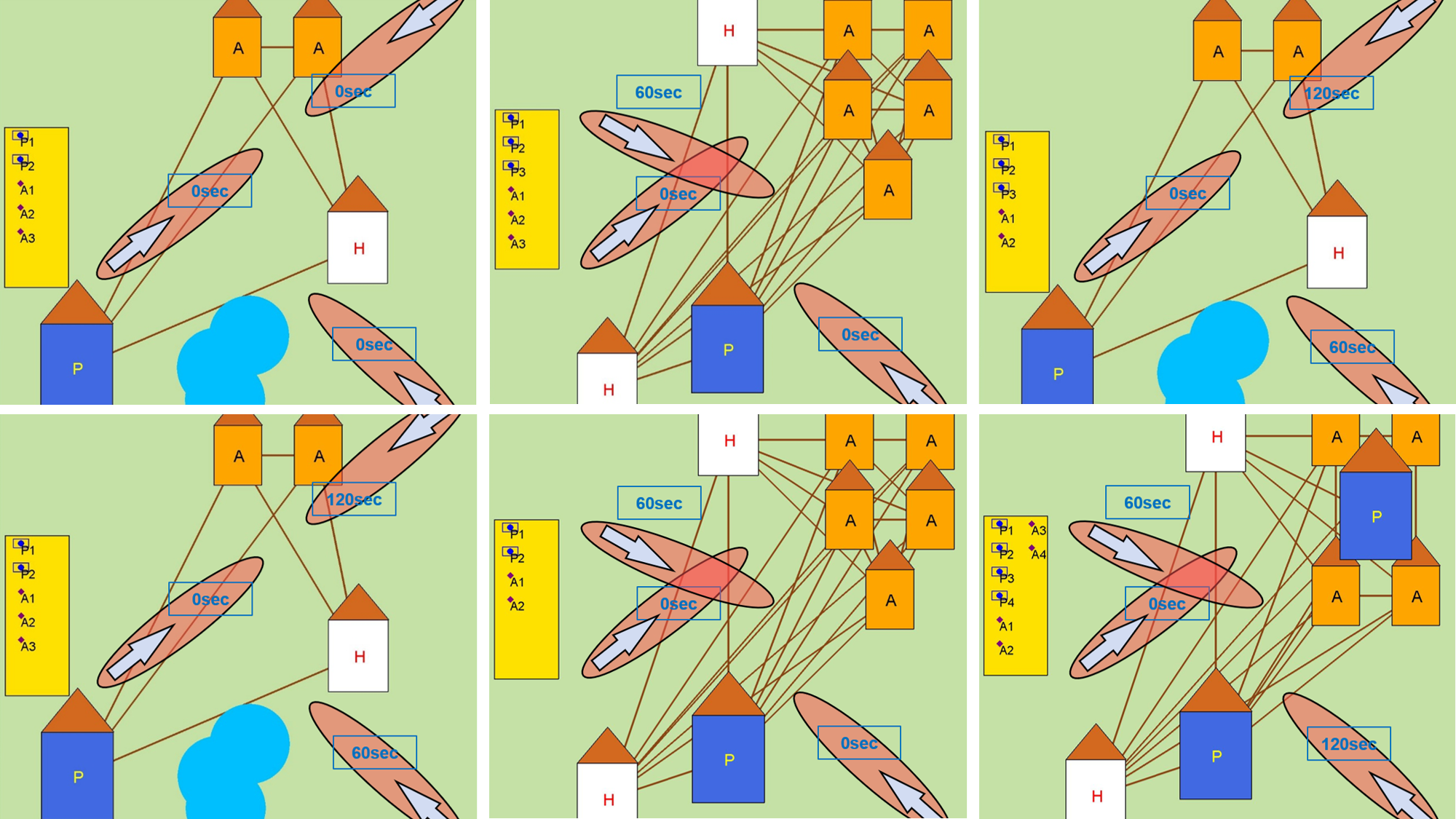}
        \caption{The second batch of Moderate Scenarios (Scenario \#13 -- \#18)}
        \label{fig:moderate_scenarios2}
    \end{figure}
    \begin{table}[h!]
        \centering
        \begin{tabular}{|c||c|c|c|c||c|c|}
            \hline
            \multirow{2}{*}{\textbf{Scenario Index}} & \multicolumn{4}{|c||}{\textbf{\# of Targets}} & \multicolumn{2}{|c|}{\textbf{\# of Agents}} \\
            \cline{2-7} & House & Hospital & Power Station & Lake & Perception & Action \\
            \hline
            \hline
            \textbf{\#7} & 2 & 1 & 1 & 1 & 3 & 1\\
            \hline
            \textbf{\#8} & 2 & 1 & 1 & 1 & 3 & 1\\
            \hline
            \textbf{\#9} & 2 & 1 & 1 & 1 & 2 & 2\\
            \hline
            \textbf{\#10} & 2 & 1 & 1 & 1 & 2 & 2\\
            \hline
            \textbf{\#11} & 2 & 1 & 1 & 1 & 2 & 2\\
            \hline
            \textbf{\#12} & 2 & 1 & 1 & 1 & 3 & 2\\
            \hline
            \textbf{\#13} & 5 & 2 & 1 & 0 & 2 & 2\\
            \hline
            \textbf{\#14} & 2 & 1 & 1 & 1 & 3 & 2\\
            \hline
            \textbf{\#15} & 5 & 2 & 1 & 0 & 3 & 3\\
            \hline
            \textbf{\#16} & 2 & 1 & 1 & 1 & 2 & 3\\
            \hline
            \textbf{\#17} & 4 & 2 & 2 & 0 & 4 & 4\\
            \hline
            \textbf{\#18} & 2 & 1 & 1 & 1 & 2 & 3\\
            \hline
        \end{tabular}
        \caption{The Targets and Agent Setting in the Moderate Scenario (Scenario \#7 -- \#18)}
        \label{tab:moderate_target}
    \end{table}
    \begin{table}[h!]
        \centering
        \begin{tabular}{|c||c|c|c|c|c|c|}
            \hline
            \makecell[c]{\textbf{Scenario} \\ \textbf{Index}} & \makecell[c]{\textbf{Region-wise} \protect\\ \textbf{\# of Fire Front}} & \makecell[c]{\textbf{Fire Delay} \\ \textbf{(Sec)}} & \makecell[c]{\textbf{Fuel} \\ \textbf{Coefficient}} & \makecell[c]{\textbf{Wind Speed}} & \makecell[c]{\textbf{Wind} \\ \textbf{Direction}} \\
            \hline
            \hline
            \textbf{\#7} & 3, 3 & 0, 0 & 5, 5 & 5, 5 & 135, 225\\ 
            \hline
            \textbf{\#8} & 5, 7 & 0, 60 & 3, 3 & 10, 10 & 135. 225\\ 
            \hline
            \textbf{\#9} & 5, 5 & 0, 0 & 10, 10 & 5, 5 & 45, 15\\ 
            \hline
            \textbf{\#10} & 5, 5 & 60, 0 & 10, 10 & 5, 5 & 45, 15\\ 
            \hline
            \textbf{\#11} & 5, 5 & 0, 0 & 5, 5 & 5, 5 & 135, 225\\ 
            \hline
            \textbf{\#12} & 3, 10 & 0, 60 & 5, 10 & 5, 10 & 135, 225\\ 
            \hline
            \textbf{\#13} & 3, 3, 3 & 60, 0, 0 & 10, 10, 10 & 5, 5, 5 & 75, 135, 225\\ 
            \hline
            \textbf{\#14} & 5, 5, 5 & 0, 120, 60 & 10, 10, 10 & 5, 5, 5 & 135, 315, 225\\ 
            \hline
            \textbf{\#15} & 3, 5, 7 & 60, 0, 0 & 10, 10, 10 & 3, 5, 10 & 75, 135, 225\\ 
            \hline
            \textbf{\#16} & 3, 5, 7 & 0, 120, 60 & 3, 5, 10 & 5, 5, 5 & 135, 315, 225\\ 
            \hline
            \textbf{\#17} & 5, 5, 5 & 60, 0, 120 & 10, 10, 10 & 5, 5, 5 & 75, 135, 225\\ 
            \hline
            \textbf{\#18} & 3, 3, 5 & 0, 0, 0 & 5, 5, 10 & 3, 3, 5 & 135, 315, 225\\ 
            \hline
        \end{tabular}
        \caption{The Fire Setting in the Moderate Scenario (Scenario \#7 -- \#18)}
        \label{tab:moderate_fire}
    \end{table}
    
    \item \textbf{Hard Scenarios (\#19 -- \#24):} Hard scenarios include the Scenario \#19 -- \#24, and their respective settings are presented in Figure~\ref{fig:hard_scenarios}. Table~\ref{tab:hard_target} presents the facility/target and agent setting and Table~\ref{tab:hard_fire} presents the fire setting for the hard scenarios. For the cases that multiple fire regions exist in the same scenario, parameters defining different fire regions are separated by a comma.
    \begin{figure}[t!]
        \centering
        \includegraphics[width=0.7\textwidth]{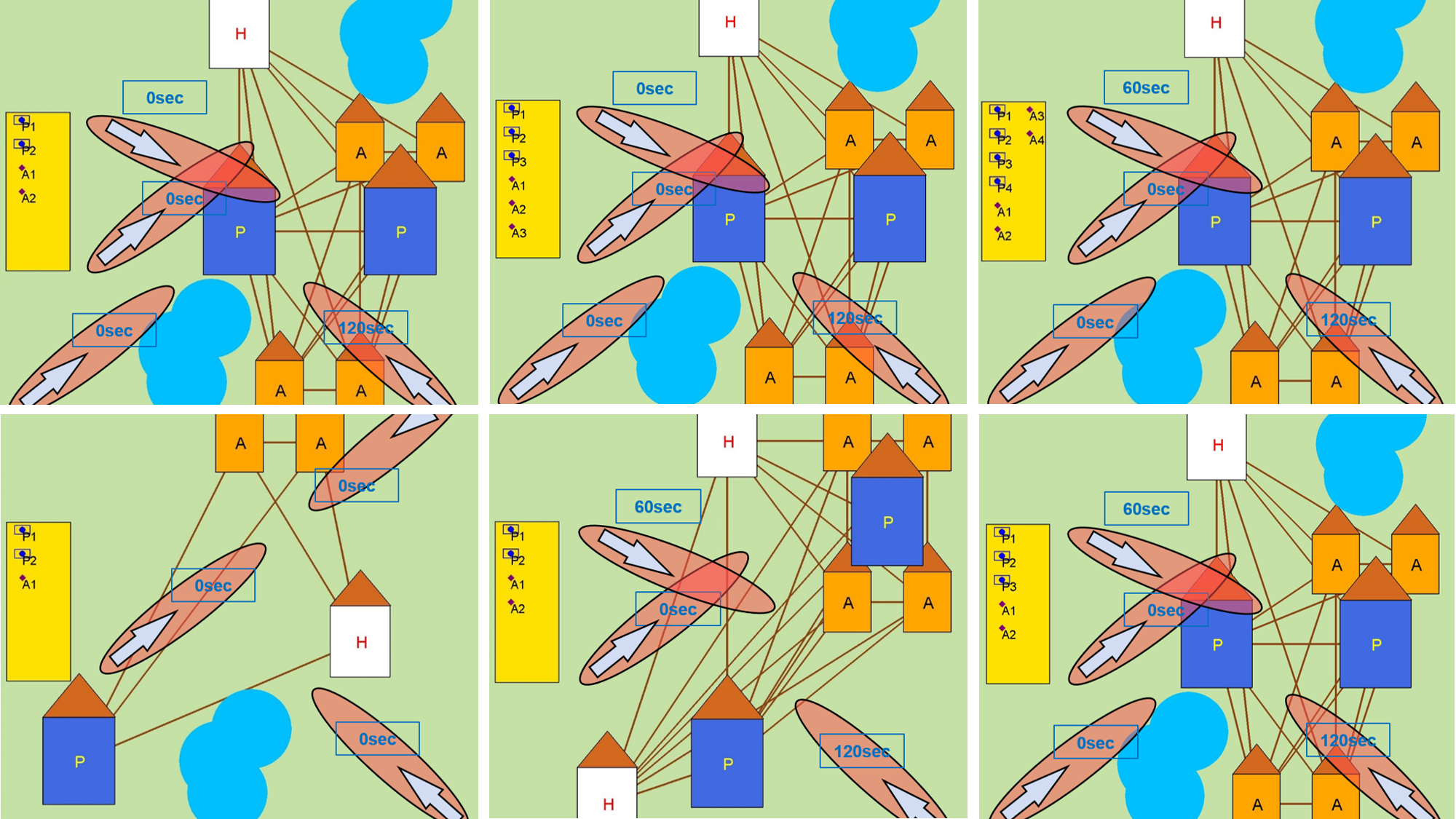}
        \caption{The Hard Scenarios (Scenario \#19 -- \#24)}
        \label{fig:hard_scenarios}
    \end{figure}
    \begin{table}[h!]
        \centering
        \begin{tabular}{|c||c|c|c|c||c|c|}
            \hline
            \multirow{2}{*}{\textbf{Scenario Index}} & \multicolumn{4}{|c||}{\textbf{\# of Targets}} & \multicolumn{2}{|c|}{\textbf{\# of Agents}} \\
            \cline{2-7} & House & Hospital & Power Station & Lake & Perception & Action \\
            \hline
            \hline
            \textbf{\#19} & 4 & 2 & 2 & 0 & 2 & 2\\
            \hline
            \textbf{\#20} & 2 & 1 & 1 & 1 & 2 & 1\\
            \hline
            \textbf{\#21} & 4 & 1 & 2 & 2 & 4 & 4\\
            \hline
            \textbf{\#22} & 4 & 1 & 2 & 2 & 3 & 3\\
            \hline
            \textbf{\#23} & 4 & 1 & 2 & 2 & 3 & 2\\
            \hline
            \textbf{\#24} & 4 & 1 & 2 & 2 & 2 & 2\\
            \hline
        \end{tabular}
        \caption{The Targets and Agent Setting in the Hard Scenario (Scenario \#19 -- \#24)}
        \label{tab:hard_target}
    \end{table}
    \begin{table}[h!]
        \centering
        \begin{tabular}{|c||c|c|c|c|c|c|}
            \hline
            \makecell[c]{\textbf{Scenario} \\ \textbf{Index}} & \makecell[c]{\textbf{Region-wise} \protect\\ \# \textbf{of Fire Front}} & \makecell[c]{\textbf{Fire Delay} \\ \textbf{(Sec)}} & \makecell[c]{\textbf{Fuel} \\ \textbf{Coefficient}} & \makecell[c]{\textbf{Wind Speed}} & \makecell[c]{\textbf{Wind} \\ \textbf{Direction}} \\
            \hline
            \hline
            \textbf{\#19} & 5, 5, 5 & 60, 0, 120 & 10, 10, 10 & 5, 5, 5 & 75, 135, 225\\ 
            \hline
            \textbf{\#20} & 3, 3, 3 & 0, 0, 0 & 5, 5, 5 & 5, 5, 5 & 135, 315, 225\\ 
            \hline
            \textbf{\#21} & 5, 5, 5, 5 & 60, 0, 0, 120 & 10, 10, 10, 10 & 5, 5, 5, 5 & 75, 135, 135, 225\\ 
            \hline
            \textbf{\#22} & 5, 5, 5, 5 & 0, 0, 0, 120 & 5, 5, 10, 10 & 5, 5, 5, 5 & 75, 135, 135, 225\\ 
            \hline
            \textbf{\#23} & 3, 3, 5, 7 & 60, 0, 0, 120 & 3, 5, 3, 10 & 5, 5, 5, 5 & 75, 135, 135, 225\\ 
            \hline
            \textbf{\#24} & 3, 5, 7, 8 & 0, 0, 0, 120 & 8, 8, 8, 8 & 3, 3, 3, 3 & 75, 135, 135, 225\\
            \hline
        \end{tabular}
        \caption{The Fire Setting in the Hard Scenario (Scenario \#19 -- \#24)}
        \label{tab:hard_fire}
    \end{table}
\end{itemize}

%%%%%%%%%%%%%%%%%%%%%%%%%%%%%%%%%%%%%%%%%%%%%%%%%%%%%%%%%%%%%%%%%%%%%%%%%%%%%%%%%%%%%%%%%%%%%%%%%%%%%%%%%%%%%%%%%%%%%%%%%%%%%%%%%%%%%%%%%%%%%%%%%%%%%%%%%%%%%%%%%%%%%%%%%%%%%%%%%%%%%%%%%%%%%%%%%%%%%%%%%%%%%%%%%%%%%%%%%%%%%%%%%%%%%%%%%%%%%%%%%%%%%%%%%%%%%%%%%%%%%%%%%%%%%%%%%%%%%%%%%%%%%

\subsection{Open-world Mode}
\label{subsec:Open-worldMode}
\noindent The open-world mode is provided as a tool for users to design their own scenarios, in a step-by-step procedure. In each step, the parameters required for specific objects in the environment (such as agents, targets/facilities and even wildfire) are asked to be inserted by the user, through a delicate, user-friendly GUI. Each section of the Open-World Mode is elaborated in the following Sections.

\subsubsection{General Setting Page}
\label{subsubsec:GeneralSettingPage}
\begin{figure}[t!]
\centering
\includegraphics[width=0.7\textwidth]{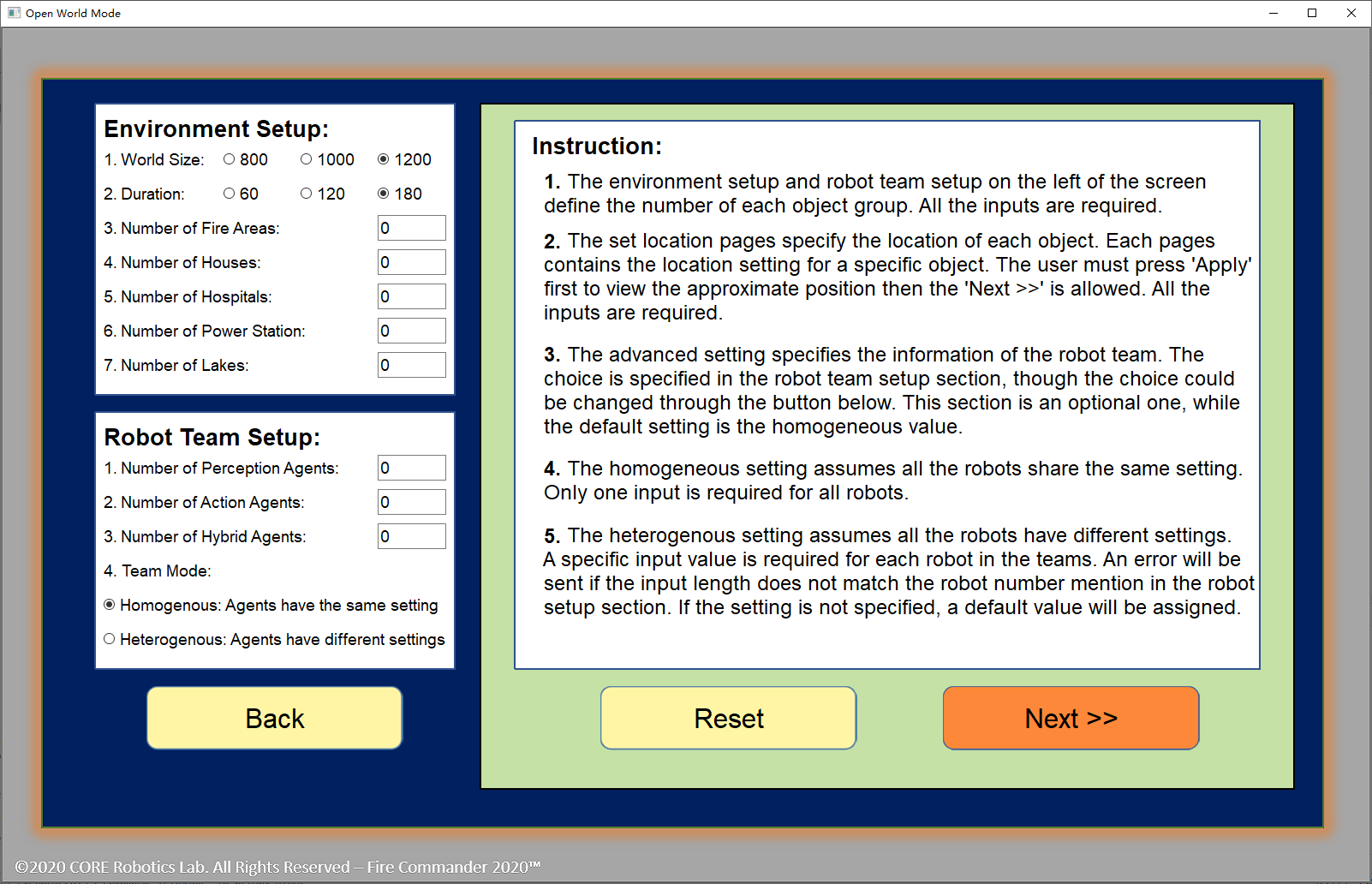}
\caption{The General Setting Page.}
\label{fig:General_Setting}
\end{figure}

\noindent Represented in Figure~\ref{fig:General_Setting}, the general setting page serves as the entry page of the open-world mode and defines the basic parameters for the user-designed scenario, such as the number of each element in the environment. These elements can be specified on the left-side of the general setting page. The environment setup section include parameters such as the world size, the scenario duration and the number of each kind of targets, including the fire's initial areas, houses, hospitals, power stations and lake. Table.~\ref{tab:target_constraint} presents the constraints and default values on these parameters. The world size offers three choices: small (800$\times$800 grid world), medium (1000$\times$1000 grid world) and large (1200$\times$1200 grid world). The duration of the scenario offers three choices as well: short game (60 Seconds), medium-duration game (120 Seconds) and long game (180 Seconds). To avoid extremely dense environments, the number of each element cannot be more than five, while their minimum numbers can be set to zero (expect for the number of fire areas which has a minimum of one region allowed). On the right-side of the general setting page a brief instruction on how to use the open-world mode is provided and on the bottom of the page, there is the \underline{Back} button to return to the welcome page, the \underline{Reset} button to clear all the inputs and the \underline{Next} button to proceed to next page.

The robot team setup section include parameters such as the number of each type of agents, e.g., Perception, Action and Hybrid agent, as well as a choice of homogeneous or heterogeneous modes for agents of the same type (e.g., agents of the same type have similar or different characteristics such as the velocity, battery capacity, etc.). The constraints on the robot team composition are as follow: 
\begin{itemize}
    \item Since switching between the agents is made by pressing 1-9 digit keys on the keyboard, the sum of all agents cannot exceed nine. This restriction is only for the GUI version and the MARL package does not enforce any restriction on the number of agents.
    
    \item The composition of the robot team should include at least one agent that can perform perception and at least one agent for taking action. As such, when the number of Hybrid agents is set to zero, the minimum number of Perception and Action agents will be one.
\end{itemize}
\begin{table}[h!]
\centering
\begin{tabular}{ |l||c|c| }
    \hline
    \textbf{Target Description} & \textbf{Limitations (Range or Acceptable Value)} & \textbf{Default Value} \\ 
    \hline
    \hline
    \textbf{World Size} & 800/1000/1200 (grid) & 1200 \\ 
    \hline
    \textbf{Duration} & 60/120/180 (Sec) & 180 \\ 
    \hline
    \textbf{Fire Region} & 1 -- 5 (grid) & 0 \\
    \hline
    \textbf{Agent Base} & 1 (\#) & 1 \\ 
    \hline
    \textbf{House} & 1 -- 5 (\#) & 0 \\
    \hline
    \textbf{Hospital} & 1 -- 5 (\#) & 0 \\
    \hline
    \textbf{Power Station} & 1 -- 5 (\#) & 0 \\
    \hline
    \textbf{Lake} & 1 -- 5 (\#) & 0 \\
    \hline
\end{tabular}
\caption{The constraints and limitations on map elements and objects.}
\label{tab:target_constraint}
\end{table}

\subsubsection{Wildfire Setting Page}
\label{subsubsec:WildfireSettingPage}
\begin{figure}[t!]
\centering
    \begin{subfigure}{.45\textwidth}
      \centering
      % include first image
      \includegraphics[width=1\linewidth]{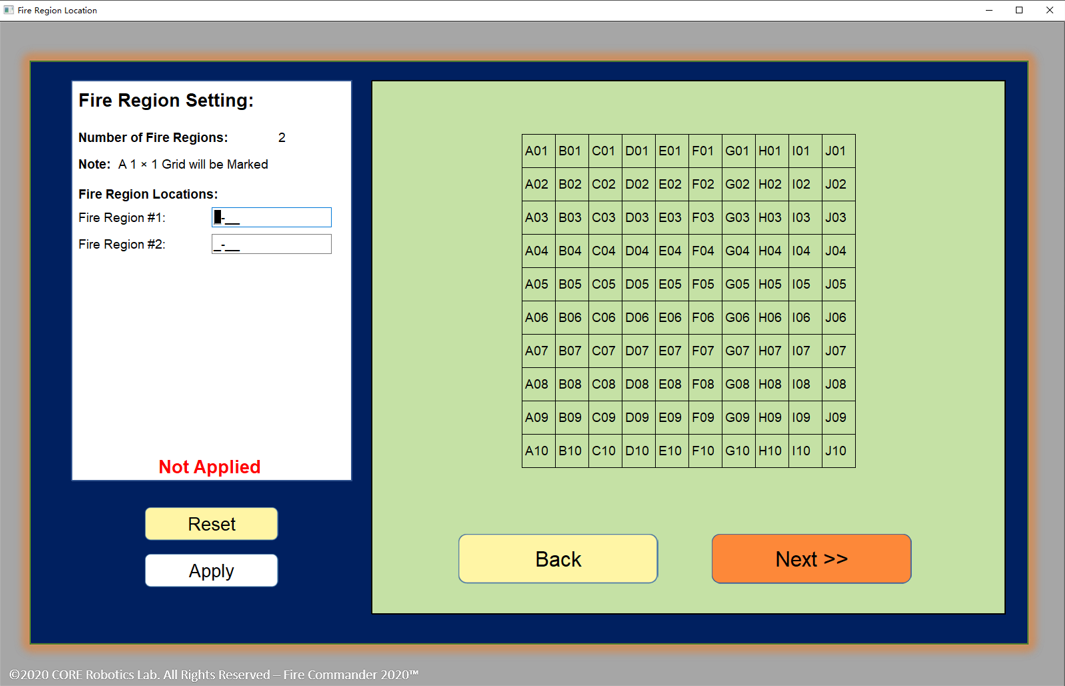}  
      \caption{Before Apply}
      \label{fig:fire_region-first}
    \end{subfigure}
    \begin{subfigure}{0.45\textwidth}
      \centering
      % include second image
      \includegraphics[width=1\linewidth]{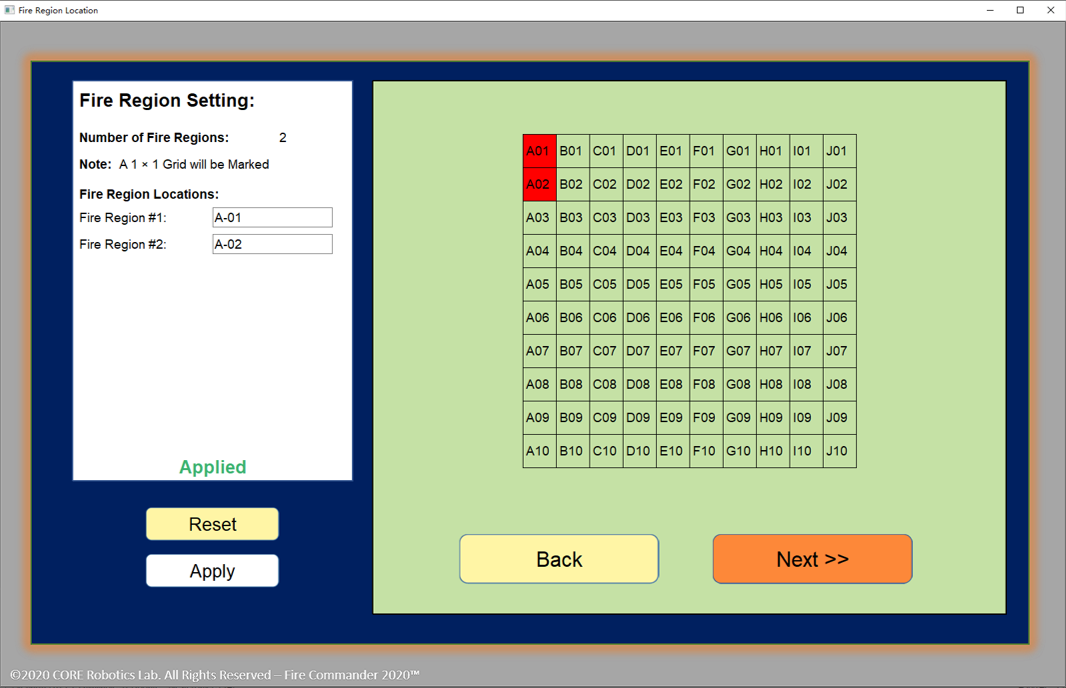}  
      \caption{After Apply}
      \label{fig:fire_region-second}
    \end{subfigure}
    \caption{The Fire Region Setting Page}
    \label{fig:fire_region}
\end{figure}

\noindent Wildfire settings are divided into two separate pages. First, users need to define the initial location of the fire regions on the map, and then, specify the details of the fire propagation model parameters and penalty coefficients on a separate page. Figures~\ref{fig:fire_region-first} and~\ref{fig:fire_region-second} present the structure of the fire region location setting page. On the left-side of the page user can specify the location of each fire area, by choosing one of the grids depicted on the right-side panel, which follows a chess-like naming convention. Each initial fire region occupies a 1$\times$1 grid on the right-side grid-map. As an instance, the fire region on 'D-05' represents a fire area in the actual environment that centers at (350, 450). Note that, while specifying the initial fire areas, two or more fire regions could overlap (e.g., start at the same location).

After all the fire region locations have been set, users need to press the \underline{Apply} button to visualize the intended fire region on the grid map. The initiated fire regions will be marked with red squares on the selected grid. Once the apply button is pressed and all the regulations on the fire regions have been satisfied, the ``\textcolor{red}{\textbf{Not Applied}}" flag will turn into ``\textcolor{green}{\textbf{Applied}}". If users are not satisfied with the current fire region, they could press the \underline{Reset} button to restart the design.

The grid-map on the right-side of the screen will include an approximate visualization of all facilities/targets as well so that users can track their scenario design. Different types of facilities/targets are marked with different colors. The colors used to represent different targets and their respective sizes are listed below.
\begin{itemize}
\item \textcolor{green}{\textbf{Green}}: Grassland, the default element on the map

\item \textcolor{red}{\textbf{Red}}: Fire region, 1$\times$1 Grid

\item \colorbox{black}{\textcolor{yellow}{\textbf{Yellow}}}: Agent base, 4$\times$2 Grid (Vertical) or 2$\times$4 Grid (Horizontal)

\item \textcolor{BurntOrange}{\textbf{Orange}}: House, 2$\times$2 Grid

\item \colorbox{black}{\textcolor{white}{\textbf{White}}}: Hospital, 2$\times$2 Grid

\item \textcolor{blue}{\textbf{Dark Blue}}: Power Station, 2$\times$2 Grid

\item \textcolor{cyan}{\textbf{Light Blue}}: Lake, 4$\times$3 Grid
\end{itemize}
\begin{figure}[t!]
\centering
    \begin{subfigure}{.45\textwidth}
      \centering
      % include first image
      \includegraphics[width=1\linewidth]{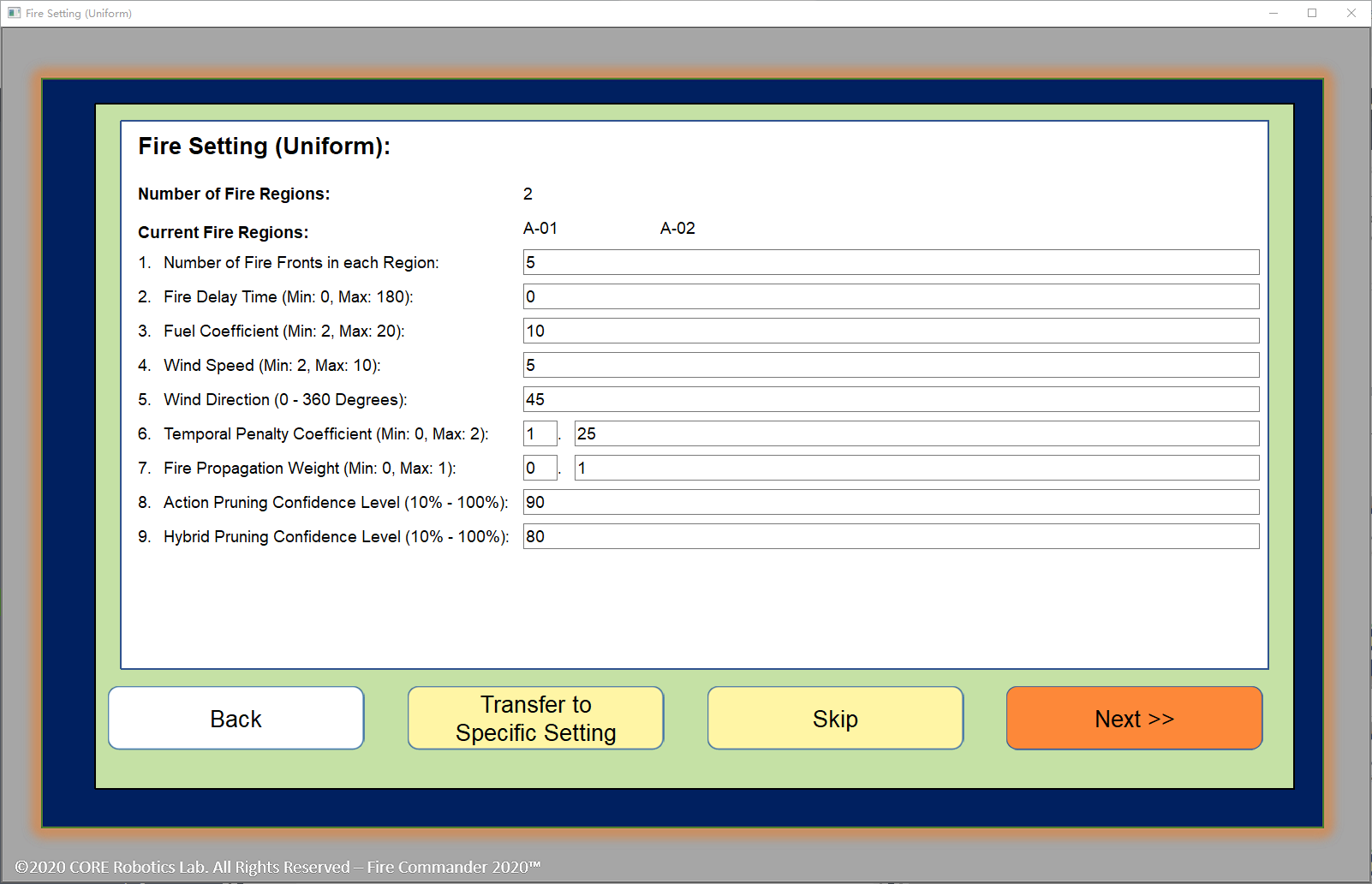}  
      \caption{Uniform Fire Setting}
      \label{fig:fire_setting-first}
    \end{subfigure}
    \begin{subfigure}{.45\textwidth}
      \centering
      % include second image
      \includegraphics[width=1\linewidth]{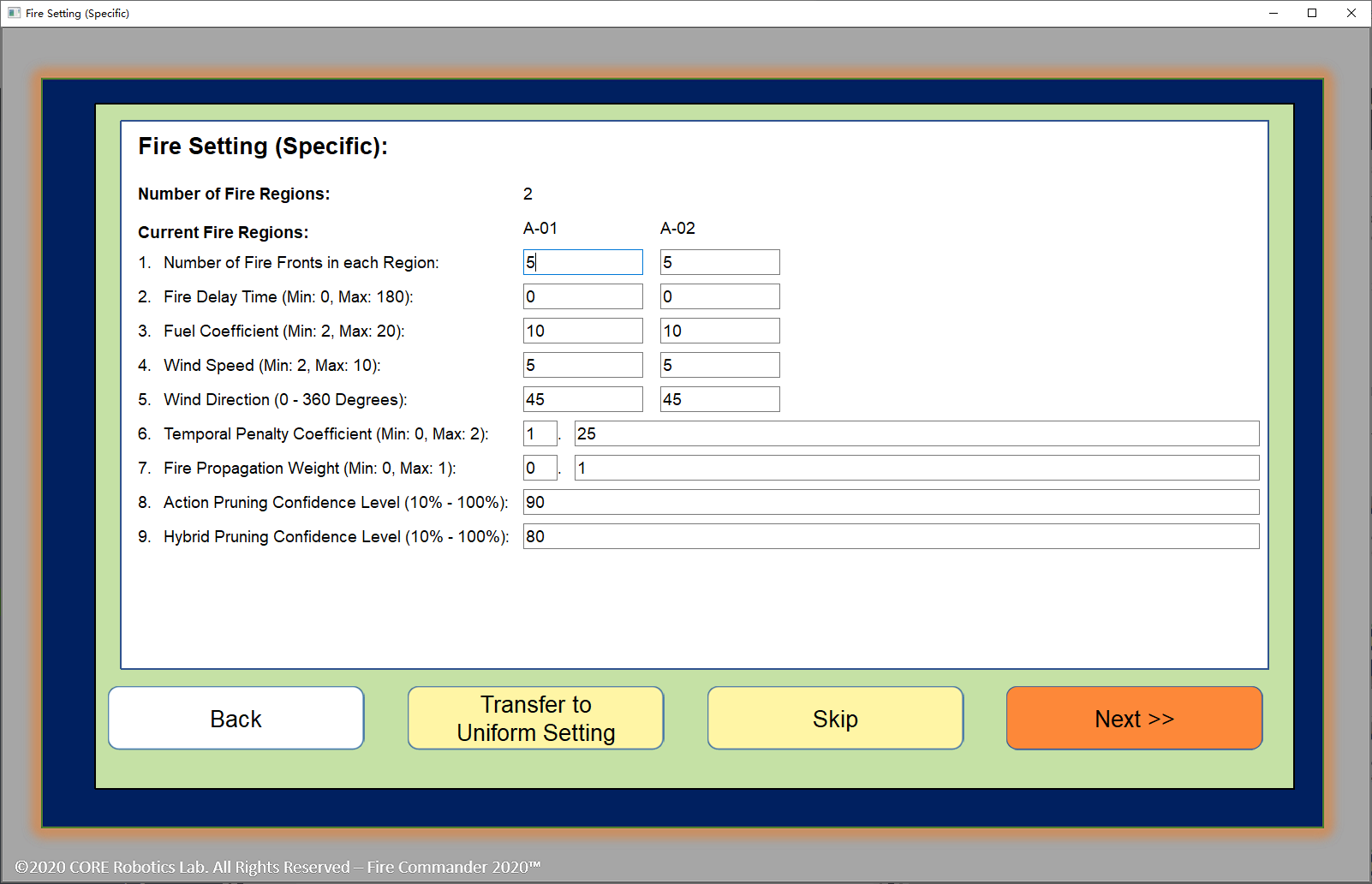}  
      \caption{Specific Fire Setting}
      \label{fig:fire_setting-second}
    \end{subfigure}
    \caption{The Fire Setting Page}
    \label{fig:fire_setting}
\end{figure}

The next step in the wildfire setting is to specify the fire propagation model parameters as well as some penalty coefficients. In our GUI, we offer two options: (1) the \underline{uniform setting} in which all fire regions use the same model parameters and, (2) the \underline{specific setting} in which fire regions use separate model parameters, each defined individually. The penalty coefficients, which related to agents' performance and how their rewards are going to be calculated, are always defined globally. In fire setting page, there are five inputs for the fire propagation model parameters and two for the penalty coefficients. In the \underline{specific setting}, users only need to fill in such fields for each separate fire region. Tables~\ref{tab:wildfire_constraint} and~\ref{tab:wildfire_score_constraint} present the parameters in this page (fire model parameters and the two score-related penalty coefficients) and their constraints.
\begin{table}[h!]
\centering
\begin{tabular}{ |l||c|c| }
\hline
\textbf{Parameter Description} & \textbf{Limitations (Range or Acceptable Value)} & \textbf{Default} \\ 
\hline
\hline
\textbf{Region-wise \# of Firefronts} & 1 -- 30 (\#) & 10 \\ 
\hline
\textbf{Fire Delay Time} & 0 -- 180 (Sec) & 0 \\ 
\hline
\textbf{Fuel Coefficient} & 2 (unit) & 10 \\
\hline
\textbf{Wind Speed} & 2 -- 10 ($m/s$) & 5 \\ 
\hline
\textbf{Wind direction} & 0 -- 360 ($\theta$) & 45 \\
\hline
\end{tabular}
\caption{The constraint and limitations on wildfire model parameters.}
\label{tab:wildfire_constraint}
\end{table}
\begin{table}[h!]
\centering
\begin{tabular}{ |l||c|c| }
\hline
\textbf{Description} & \textbf{Limitations (Range or Acceptable Value)} & \textbf{Default} \\ 
\hline
\hline
\textbf{Temporal Penalty Coefficient} & 0 - 2 & 1.25 \\
\hline
\textbf{Fire Propagation Weight} & 0 - 1 & 0.1 \\
\hline
\end{tabular}
\caption{The Constraint on Wildfire Propagation Score}
\label{tab:wildfire_score_constraint}
\end{table}

We also incorporated in this page, the pruning confidence level for the Action and Hybrid agents, which relates to the stochasticity in the environment, as introduced in Section 2.2.3, and is fixed during the game. Table~\ref{tab:pruning_confidence_level_constraint} presents the limitations and default value for the pruning confidence level. The pruning confidence level in all the pre-designed scenarios (in Scenario Mode) is set to 90\% for Action agents and 80\% for Hybrid agents, while users could choose any value between 10\% and 100\% for the Action and Hybrid agents in the Open-World Mode. Users can switch between the uniform and specific fire setting pages by choosing either \underline{Transfer to Specific Setting} or \underline{Transfer to Uniform Setting}, or choose \underline{Skip} and use the pre-determined values for parameters.
\begin{table}[h!]
\centering
\begin{tabular}{ |l||c|c| }
\hline
\textbf{Description} & \textbf{Limitations (Acceptable Range)} & \textbf{Default} \\ 
\hline
\hline
\textbf{Action Pruning Confidence Level} & 10\% - 100\% & 90\% \\
\hline
\textbf{Hybrid Pruning Confidence Level} & 10\% - 100\% & 80\% \\
\hline
\end{tabular}
\caption{The Constraint on Pruning Confidence Level for Action and Hybrid Agents}
\label{tab:pruning_confidence_level_constraint}
\end{table}

\subsubsection{Facilities/Targets Setting Page}
\label{subsubsec:TargetsFacilitiesSettingPage}
\begin{figure}[t!]
\centering
\begin{subfigure}{.45\textwidth}
  \centering
  % include first image
  \includegraphics[width=1\linewidth]{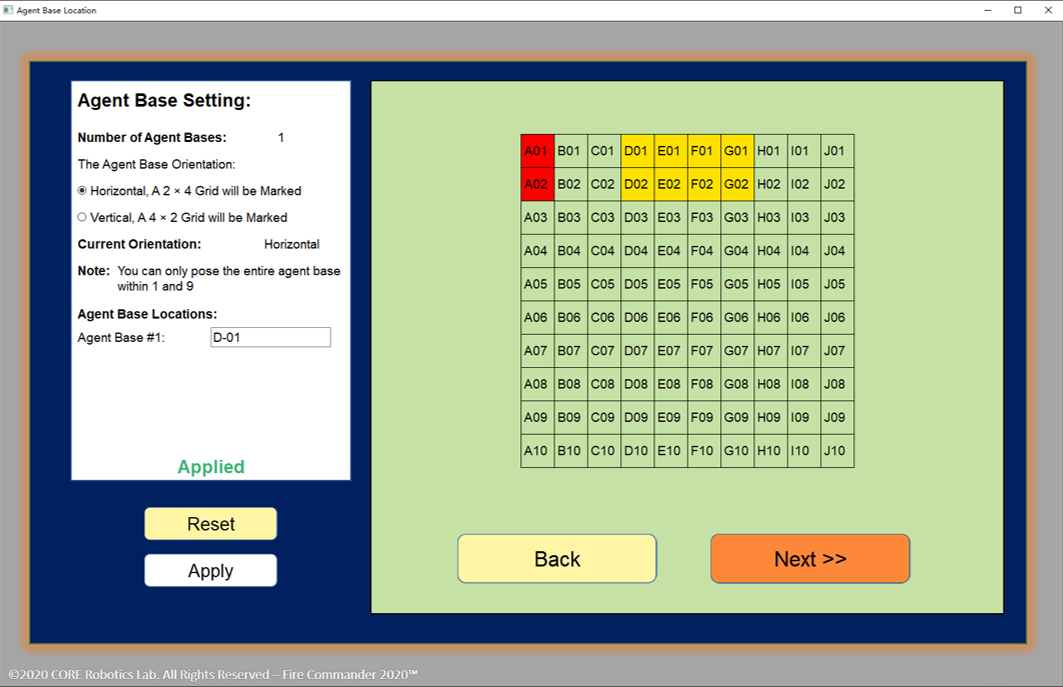}  
  \caption{Horizontal}
  \label{fig:base_horizontal}
\end{subfigure}
\begin{subfigure}{.45\textwidth}
  \centering
  % include second image
  \includegraphics[width=1\linewidth]{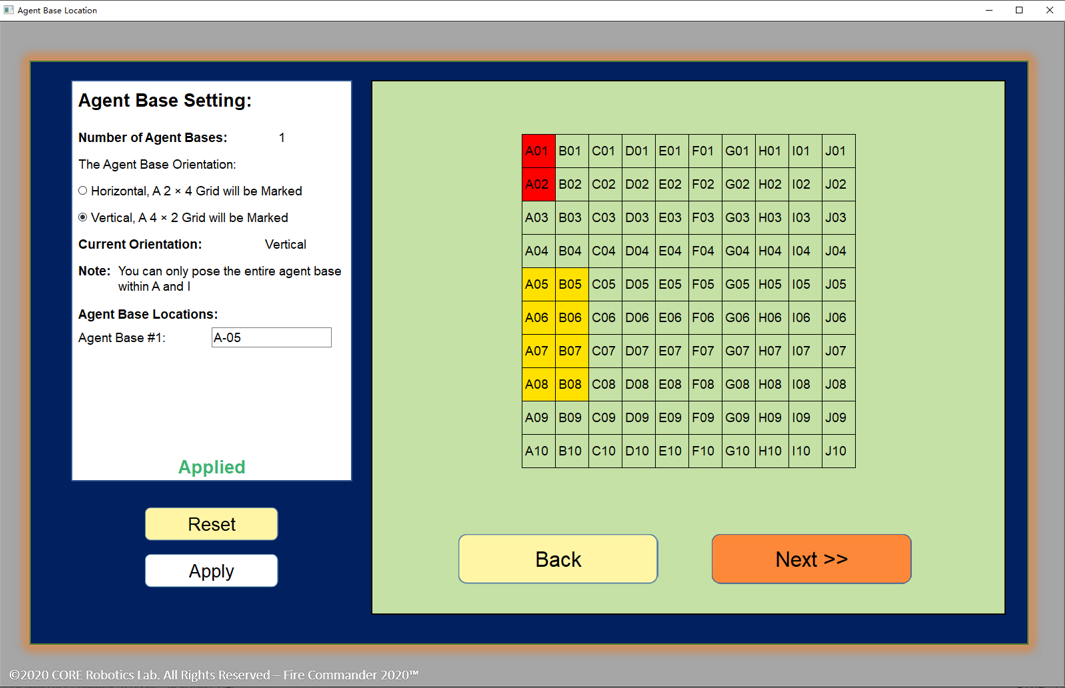}  
  \caption{Vertical}
  \label{fig:base_vertical}
\end{subfigure}
\caption{Agent Base (e.g., Depot) Setting Page}
\label{fig:base_setting}
\end{figure}
\begin{figure}[t!]
\centering
\begin{subfigure}{.45\textwidth}
  \centering
  % include first image
  \includegraphics[width=1\linewidth]{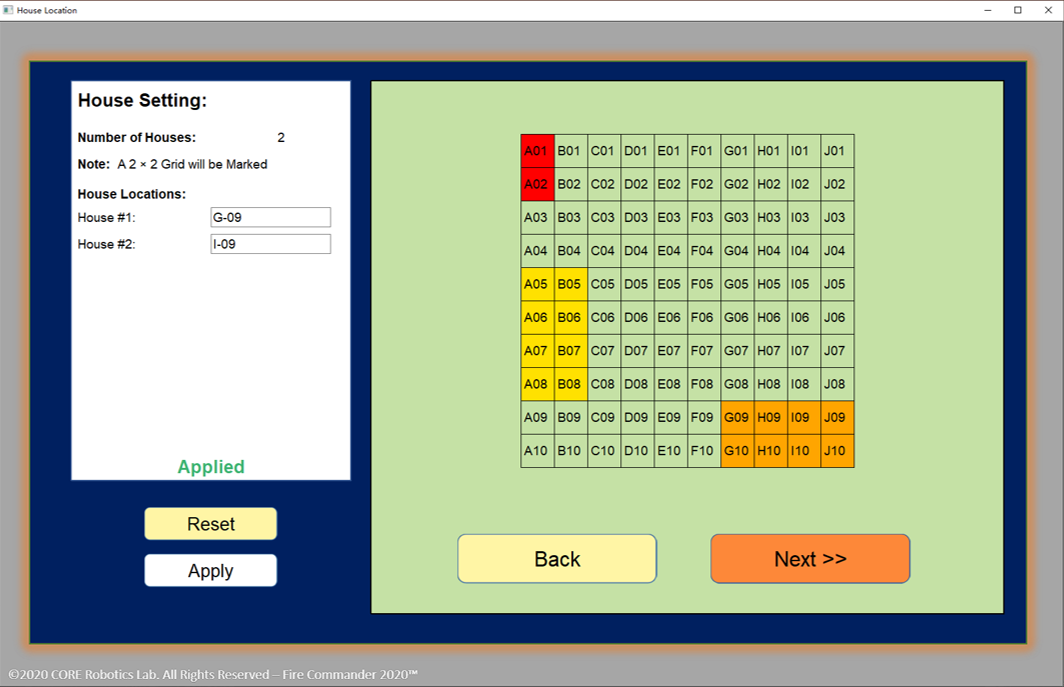}  
  \caption{House}
  \label{fig:house}
\end{subfigure}
\begin{subfigure}{.45\textwidth}
  \centering
  % include second image
  \includegraphics[width=1\linewidth]{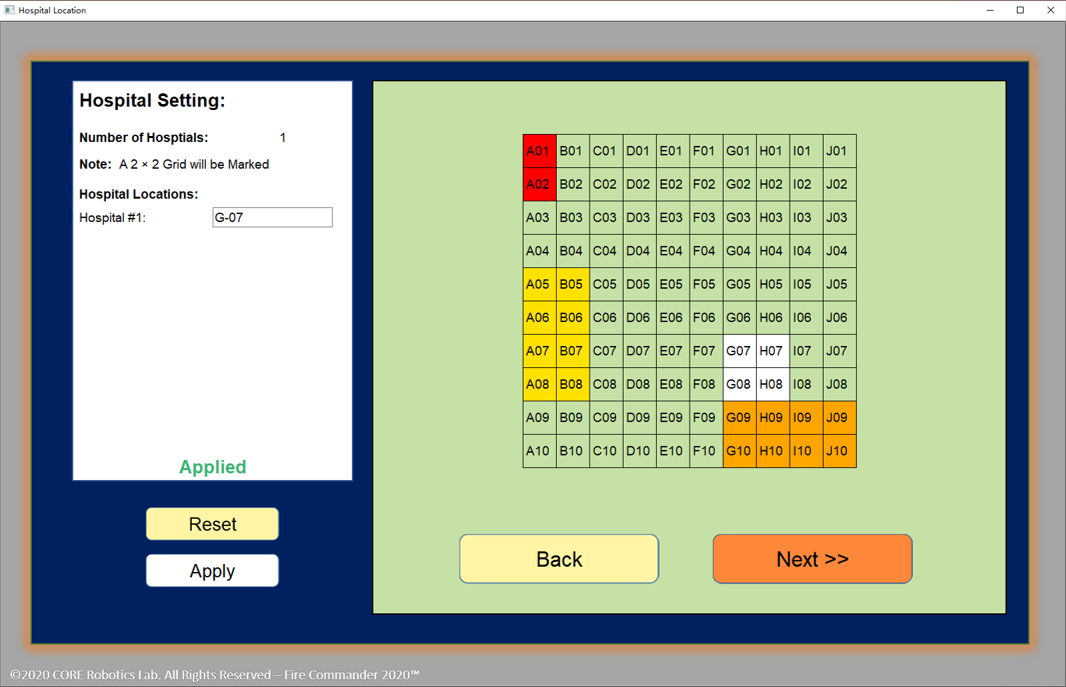}  
  \caption{Hospital}
  \label{fig:hosptial}
\end{subfigure}
\begin{subfigure}{.45\textwidth}
  \centering
  % include second image
  \includegraphics[width=1\linewidth]{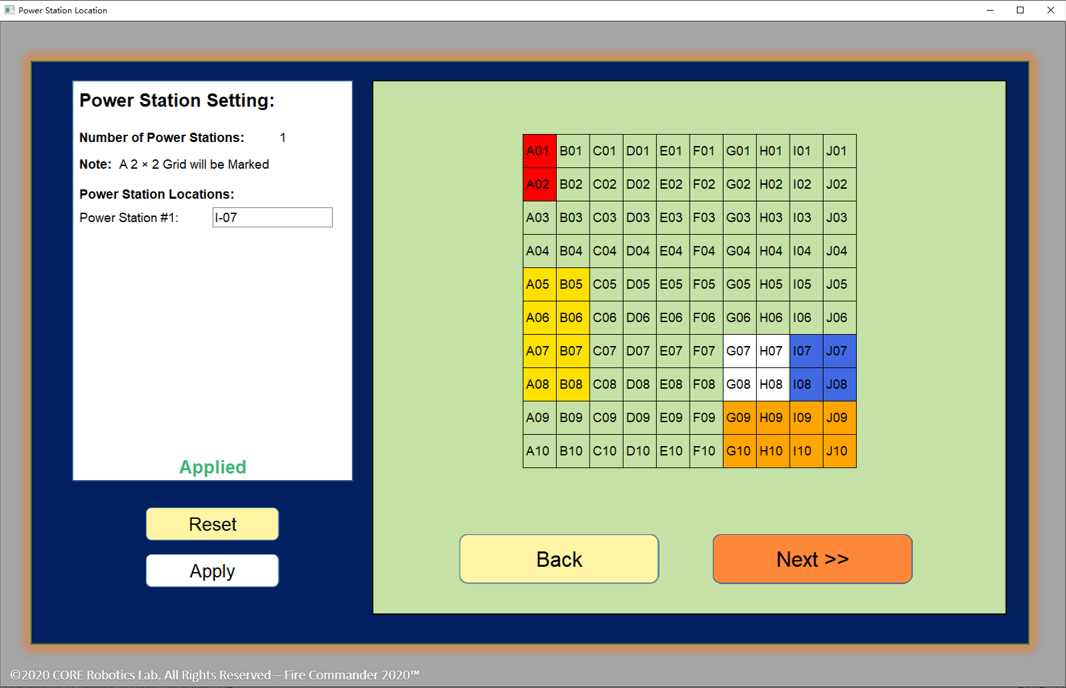}  
  \caption{Power Station}
  \label{fig:power}
\end{subfigure}
\begin{subfigure}{.45\textwidth}
  \centering
  % include second image
  \includegraphics[width=1\linewidth]{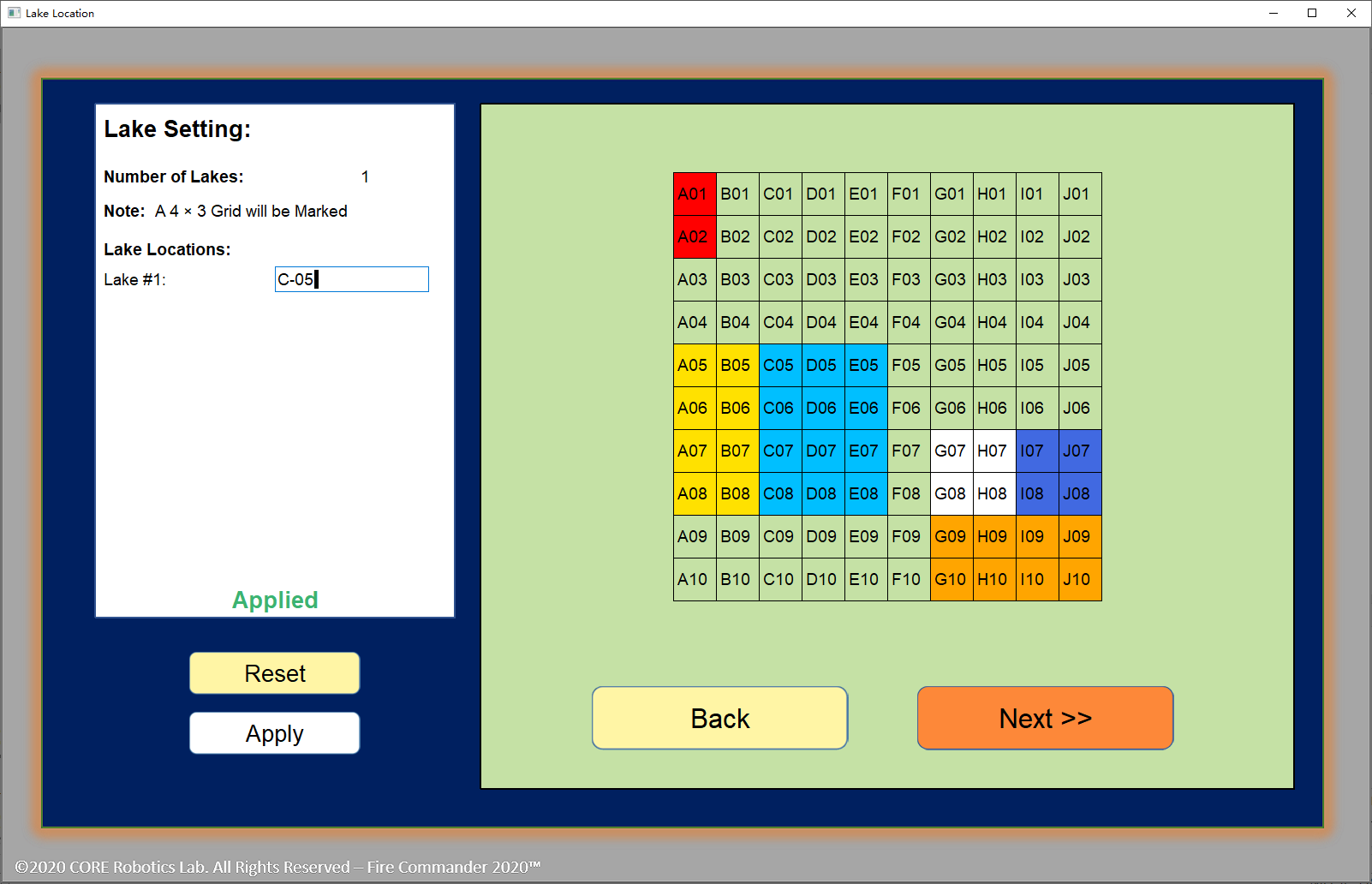}  
  \caption{Lake}
  \label{fig:lake}
\end{subfigure}
\caption{Target Setting Page}
\label{fig:target_setting}
\end{figure}
\noindent In the facility/target location design process, the first step is to specify the location and orientation of the agent base. Figure~\ref{fig:base_setting} presents the agent base setting page. Agent base is unique in all the pre-designed scenarios (in Scenario Mode), while its orientation and location could be changed in the Open-World Mode. Note that, the top-left grid of the agent base must be selected such that the entire base fits inside the map and it does not overlap with any of the existing facilities/targets. Two options are allowed for the base orientation:
\begin{itemize}
\item \textbf{Horizontal:} The agent base occupies 2$\times$4 on the map. In this case, the agent base must be located either on top or bottom edge of the map.

\item \textbf{Vertical:} The agent base occupies 4$\times$2 on the map. In this case, the agent base must be located either on left or right edge of the map. Vertical is the default option.
\end{itemize}

After the agent base setting page, the rest of the facility setting pages, such as house, hospital, power station and lake, will be shown to the user with relatively similar structures. The sizes of these facilities on the grid-map are 2$\times$2 for house, hospital and power station, and 4$\times$3 for lake. Similar constraints such as no overlapping facilities and fitting inside the map apply to all facilities.

\subsubsection{Agent Setting Page}
\label{subsubsec:AgentSettingPage}
\begin{figure}[t!]
\centering
    \begin{subfigure}{.45\textwidth}
      \centering
      % include first image
      \includegraphics[width=1\linewidth]{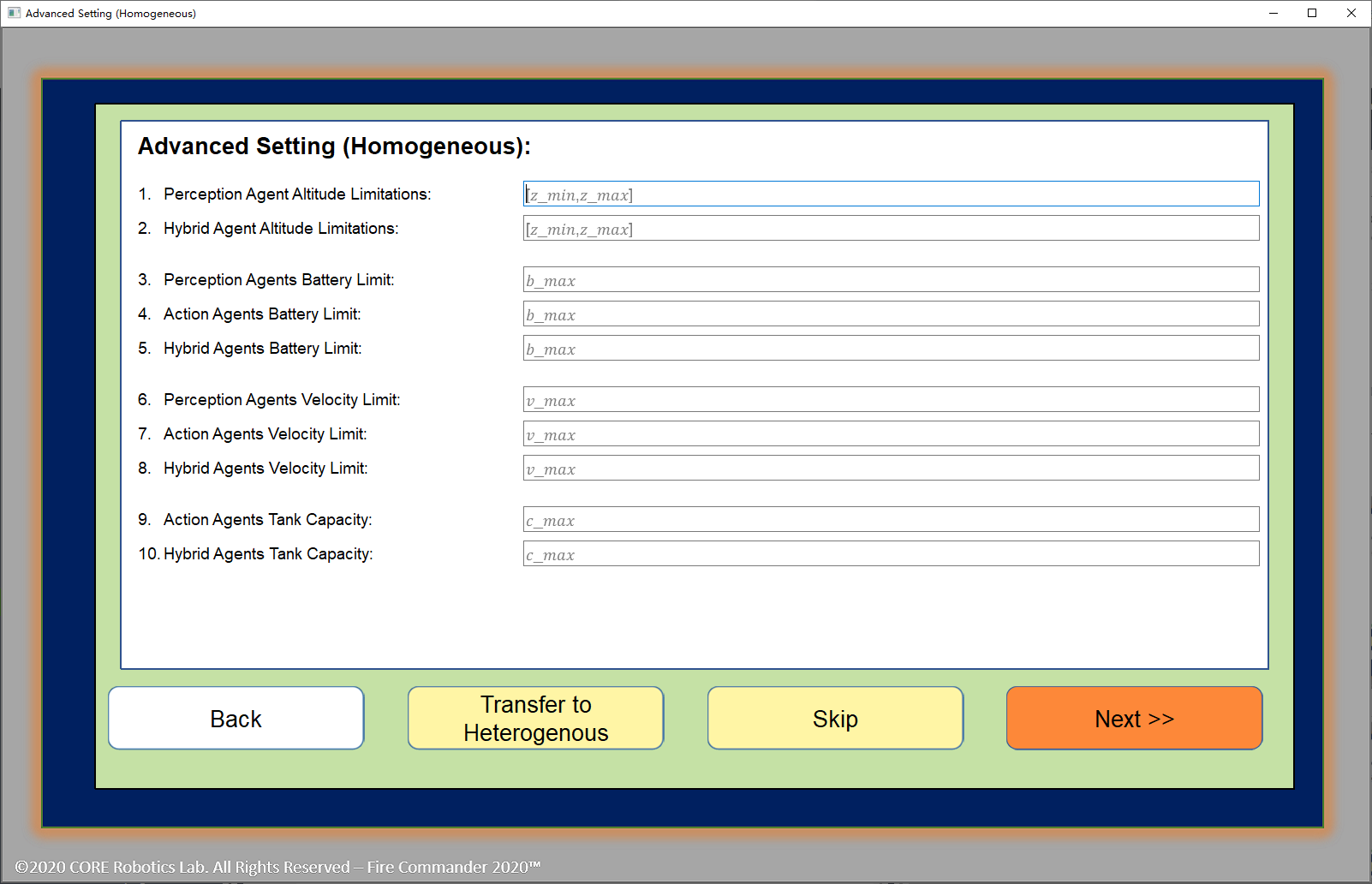}  
      \caption{Homogeneous Agent Setting Page}
      \label{fig:agent-homo}
    \end{subfigure}
    \begin{subfigure}{.45\textwidth}
      \centering
      % include second image
      \includegraphics[width=1\linewidth]{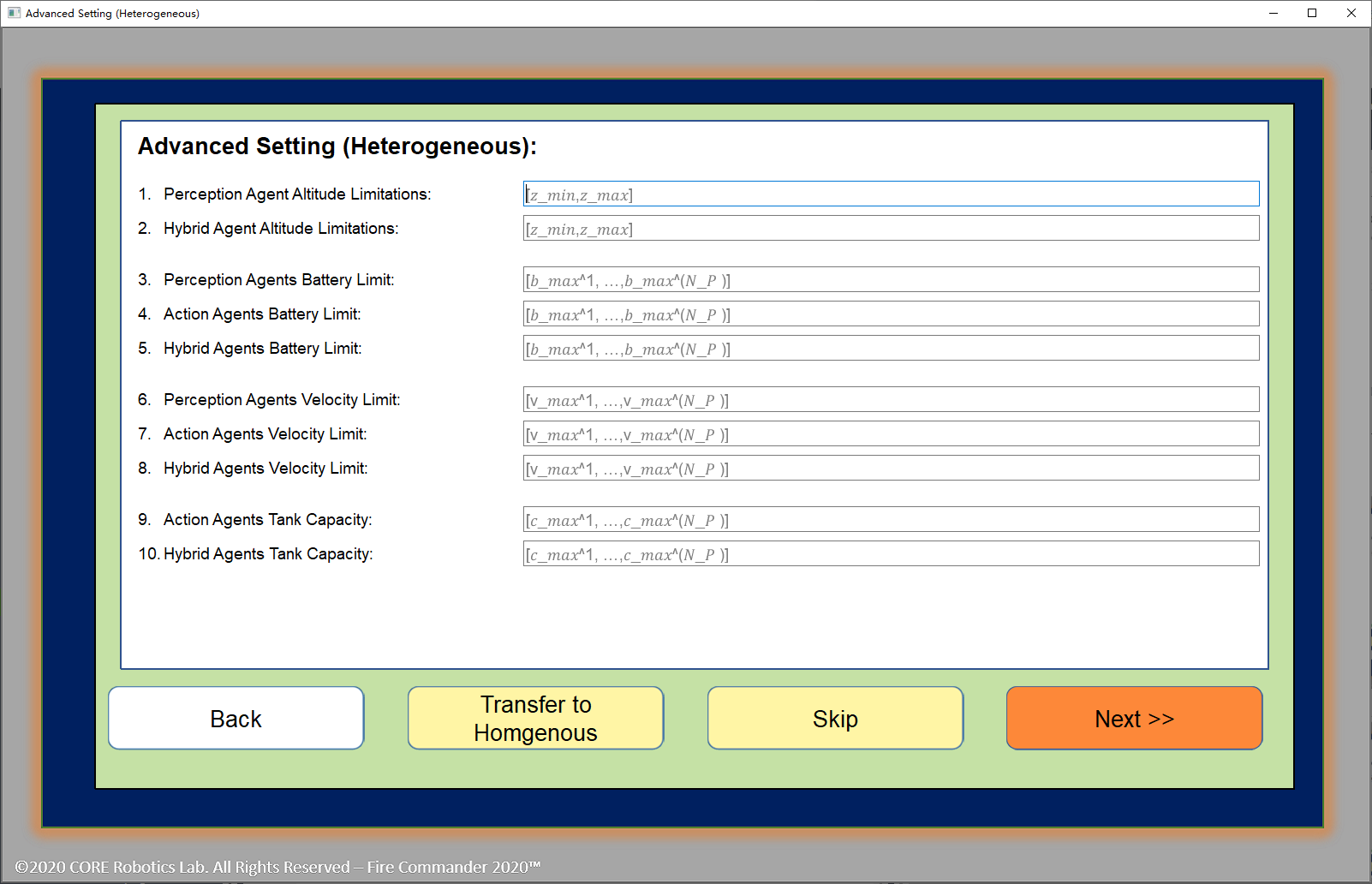}  
      \caption{Heterogeneous Agent Setting Page}
      \label{fig:agent-hetero}
    \end{subfigure}
    \caption{Agent Setting Page}
    \label{fig:agent_setting}
\end{figure}   

\noindent Figure~\ref{fig:agent_setting} presents the agent setting page that specifies the agents' characteristics and control parameters. The agents' parameters include:
\begin{itemize}
\item \textbf{Altitude Limit:} The upper and lower bound of the agents' flight altitude, which is an available option for Perception and Hybrid agents only. The flight height for the Action agents is fixed.

\item \textbf{Battery-Life Limit:} The battery capacity that determines the maximum flight duration. 

\item \textbf{Velocity Limit:} The maximum velocity for each agent, which is the default step size of the respective agent that builds its trajectory.

\item \textbf{Tanker Capacity Limit:} The number of pruning times that an Action or a Hybrid agent could could attempt during a single flight.
\end{itemize}

There are two parameter modes for the agents: (1) homogeneous parameters and (2) heterogeneous parameters. Homogeneous mode applies a similar set of parameters to all agents of the same type (i.e., all Perception agents), while the heterogeneous mode allows heterogeneity in aforementioned characteristics of agents of the same type. In homogeneous case, parameters can be entered as a single value or tuple (as shown with gray text inside the boxes) while in the heterogeneous case, parameters are inputted as lists, each element of which corresponds to one agent. Table.~\ref{tab:agent_constraint} presents the default values for the described parameters. Note that, the heterogeneous mode will automatically switch to homogeneous mode, if the parameter boxes are left empty.
\begin{table}[h!]
\centering
\begin{tabular}{ |l||c|c| }
\hline
\textbf{Description} & \textbf{Limitations (Range or Acceptable Value)} & \textbf{Default Value}\\ 
\hline
\hline
\textbf{Agent Altitude} & 10 -- 100 (m) & 10 -- 100 \\ 
\hline
\textbf{Battery Capacity} & 200 -- 800 (time steps) & 500 \\ 
\hline
\textbf{Velocity} & 10 -- 30 ($m/s$) & 20 \\
\hline
\textbf{Water Tank Capacity} & 1 -- 15 (\#) & 10 \\ 
\hline
\end{tabular}
\caption{The Constraint on Agent Setting}
\label{tab:agent_constraint}
\end{table}

Users can define all agents parameters and then proceed to preview their design by pressing \underline{Next}, or use default values for these parameters by pressing \underline{Skip}. Moreover, users are allowed to leave a few inputs blank (i.e., only specify altitude and battery limits and leave velocity and tanker capacity limits blank), in which case, the GUI will automatically fill up these inputs with default values.

%%%%%%%%%%%%%%%%%%%%%%%%%%%%%%%%%%%%%%%%%%%%%%%%%%%%%%%%%%%%%%%%%%%%%%%%%%%%%%%%%%%%%%%%%%%%%%%%%%%%%%%%%%%%%%%%%%%%%%%%%%%%%%%%%%%%%%%%%%%%%%%%%%%%%%%%%%%%%%%%%%%%%%%%%%%%%%%%%%%%%%%%%%%%%%%%%%%%%%%%%%%%%%%%%%%%%%%%%%%%%%%%%%%%%%%%%%%%%%%%%%%%%%%%%%%%%%%%%%%%%%%%%%%%%%%%%%%%%%%%%%%%%

\subsection{Preview, Tutorial, Information and Score Display Pages}
\label{subsec:OtherPages}
\noindent After all the parameters have been specified in the previous steps, users should be prepared to start a new game. In the Open-World Mode, users are first presented with a visualization of the scenario they just designed in the \underline{Preview Page}, and then, a \underline{Tutorial Page} appears to help users' with a better understandings of the game logistics and objectives. During a game, an online \underline{Information Display Screen} presents some information regarding the environment and the agents as well the game score at current time of the game to the player and eventually, after finishing a game, the player's performance in different aspects of the game will be calculated and reported in the \underline{Performance Evaluation Screen}. Details of each aforementioned page is presented in the following.

\paragraph{Preview Page:}
\begin{figure}[t!]
\centering
    \begin{subfigure}{.45\textwidth}
      \centering
      % include first image
      \includegraphics[width=1\linewidth]{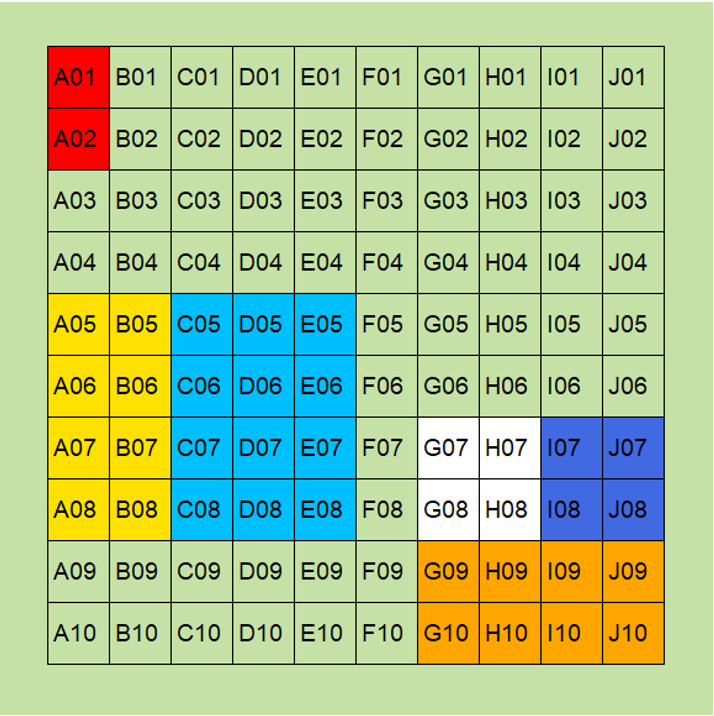}  
      \caption{Grid Map}
      \label{fig:prview-first}
    \end{subfigure}
    \begin{subfigure}{.45\textwidth}
      \centering
      % include second image
      \includegraphics[width=1\linewidth]{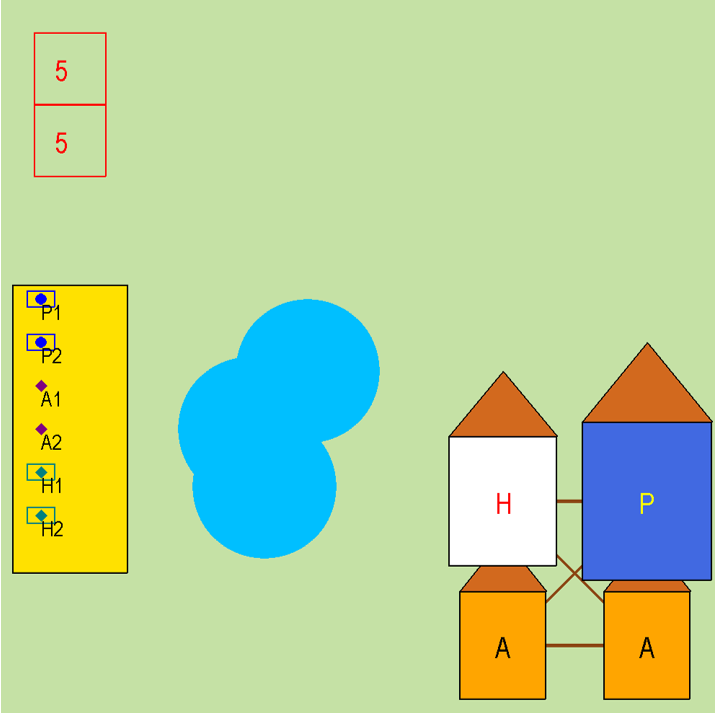}  
      \caption{Preview Page}
      \label{fig:prview-second}
    \end{subfigure}
    \caption{Preview Page}
    \label{fig:prview}
\end{figure}  

The preview page appears when users choose to proceed with their design after specifying all environment parameters. Figure~\ref{fig:prview} provides the mapping from the grid map in which users designed their desired scenario and the actual map in the game environment. Note that preview page only shows a static image of the game environment for user's confirmation of a satisfactory design. For agents, the preview page marks their ID and initial positions on the agent base. For wildfire regions, the preview page marks their position and number of initial firespots in each region. Once the user closes the preview page, a dialog box pops up with the question: ``\textsl{\underline{Do you like to proceed with this environment design}?}". Users could either choose to proceed or return to the previous design pages.

\paragraph{Tutorial Page:}
\begin{figure}[t!]
\centering
    \begin{subfigure}{.45\textwidth}
      \centering
      % include first image
      \includegraphics[width=1\linewidth]{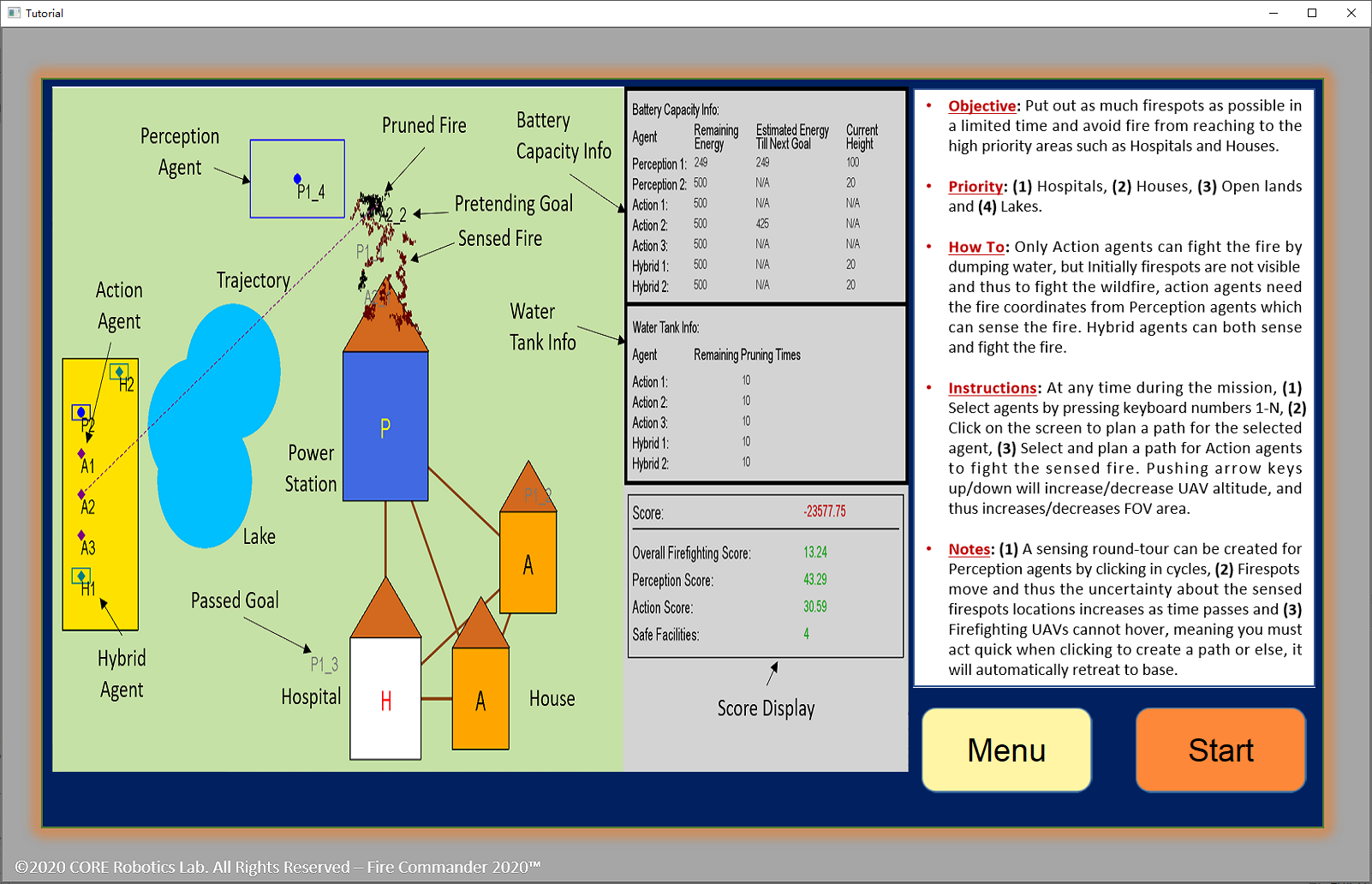}  
      \caption{Tutorial for Open-world Mode}
      \label{fig:tutorial-first}
    \end{subfigure}
    \begin{subfigure}{.45\textwidth}
      \centering
      % include second image
      \includegraphics[width=1\linewidth]{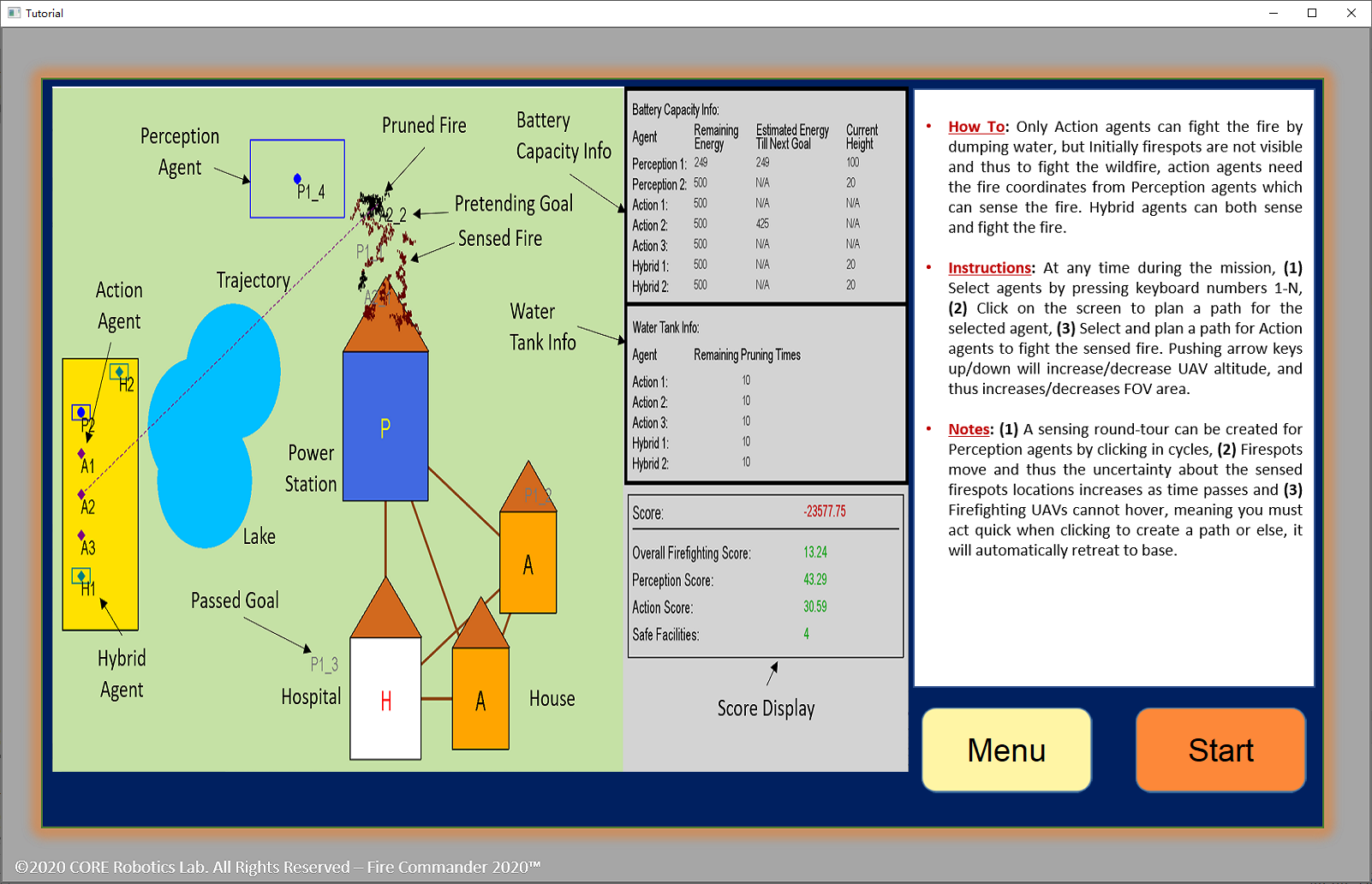}  
      \caption{Tutorial for Scenario Mode}
      \label{fig:tutorial-second}
    \end{subfigure}
    \caption{Tutorial Page}
    \label{fig:tutorial}
\end{figure}    

If users confirm the design scenario in the preview page, the GUI proceeds to a tutorial page. Figure~\ref{fig:tutorial-first} shows the tutorial page for the Open-World Mode, while Figure~\ref{fig:tutorial-second} shows the tutorial page for the Scenario Mode. The difference between the two tutorial pages is that in scenario mode, we avoid giving direct objective instructions to the users and we expect the LfD framework to determine these objectives automatically from user behaviors.

\paragraph{Information Display:}
\begin{figure}[t!]
    \centering
    \includegraphics[width=0.7\textwidth]{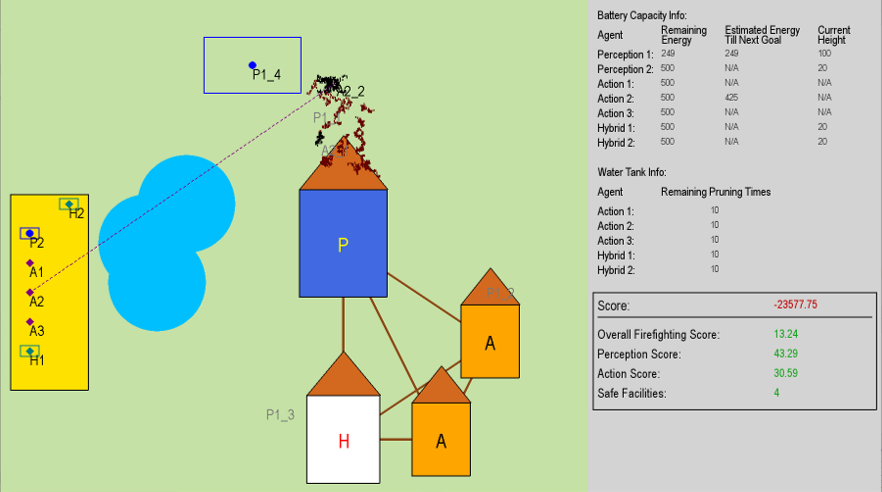}
    \caption{The Information Display.}
    \label{fig:info_display}
\end{figure}

When a game is running, the necessary information are presented to the player on an information display screen on the right-side of the environment. Figure~\ref{fig:info_display} presents the layout of the information display during a game. On the top of the display bar the battery capacity information for each agent at current time step is presented. The battery capacity info table includes the following information:
\begin{itemize}
\item \textbf{The current remaining energy in the battery:} Once the agent departs from the base, its battery begins to drain. The energy consumption rate when the agent is running is currently set to 0.1 per unit length, while the consumption rate is 0.05 for each iteration.

\item \textbf{The estimated remaining energy for agent when at goal:} This value shows the estimated remaining battery energy for the last node the player planned for the current agent. This value helps users plan and choose between agents strategically so that none of agents runs out of energy mid-flight during a mission (e.g., a planned trajectory).

\item \textbf{Current flight height:} Current flight altitude of an agent has a direct relation with the size of its FOV, and thus, is presented to players for planning. 
\end{itemize}

Under the battery capacity info table, the water tank capacity information table is presented which show the remaining tanker capacity (e.g., remaining number of times an agent can prune firespots) for the Action and Hybrid agents. Next, on the bottom of the information display screen, the online performance measures are presented to the player. The first score shown is the current total negative reward, which is always a negative value, determined by how much time has passed and how much fire has grown since the beginning of the game, as show in Equation~\ref{eq:totalnegreward}
\begin{align}
\label{eq:totalnegreward}
Total \ Negative \ Reward &= 0.1 \times Number \ of \ Active \ Firespots \\
&+ Penalty \ Coef + Number \ of \ Firespots \ in \ Targets \notag
\end{align}
Elements in Equation~\ref{eq:totalnegreward} are described below:
\begin{itemize}
\item \textit{Penalty Coefficients:} 0.1 per new firespot, 1 per House, 2 per Hospital, 5 per Power Station and 5 per Agent Base. These values can be changed in the open-source code, if required.

\item \textit{Active Firespots:} All firespots that have not been pruned yet (e.g., both sensed and not-sensed firespots are counted and the pruned firespots are excluded).

\item \textit{Firespots in Targets:} The active firespots (sensed or not-sensed) in each facility/target.
\end{itemize}

Under the total negative reward, there are four more performance metrics, which cover the following:
\begin{enumerate}
\item \textbf{Overall Firefighting Score:} The ratio between the \underline{number of the active firespots} and the \underline{total number of firespots} that have been generated so far in the game (sensed or not-sensed): 
\begin{align}
\label{eq:overallfirefightingscore}
    Overall \ Firefighting \ Score = \frac{Number \ of \ Pruned \ Firespots}{Total \ Number \ of \ Firespots} \times 100\%
\end{align}

\item \textbf{Perception Score:} The ratio between the \underline{number of sensed firespots} and the \underline{total number} \underline{of firespots} that have been generated so far in the game (sensed or not-sensed): 
\begin{align}
\label{eq:perceptionscore}
    Perception \ Score = \frac{Number \ of \ Sensed \ Firespots}{Total \ Number \ of \ Firespots} \times 100\%
\end{align}

\item \textbf{Action Score:} The ratio between the \underline{number of pruned firespots} and the \underline{total number of} \underline{sensed firespots} (e.g., numerator in Equation~\ref{eq:perceptionscore}).
\begin{align}
\label{eq:actionscore}
    Action \ Score = \frac{Number \ of \ Pruned \ Firespots}{Number \ of \ Sensed \ Firespots} \times 100\%
\end{align}

\item \textbf{Facility Protection Score:} Percentage of the number of facilities in the scenario that have been safe so far.
\begin{align}
\label{eq:facilityprrotectionscore}
    Facility \ Protection \ Score = \frac{Number \ of \ Facilities \ Never \ on \ Fire}{Total \ Number \ of \ Facilities} \times 100\%
\end{align}
\end{enumerate}

\paragraph{Performance Evaluation Page:}
\begin{figure}[t!]
    \centering
    \includegraphics[width=0.7\textwidth]{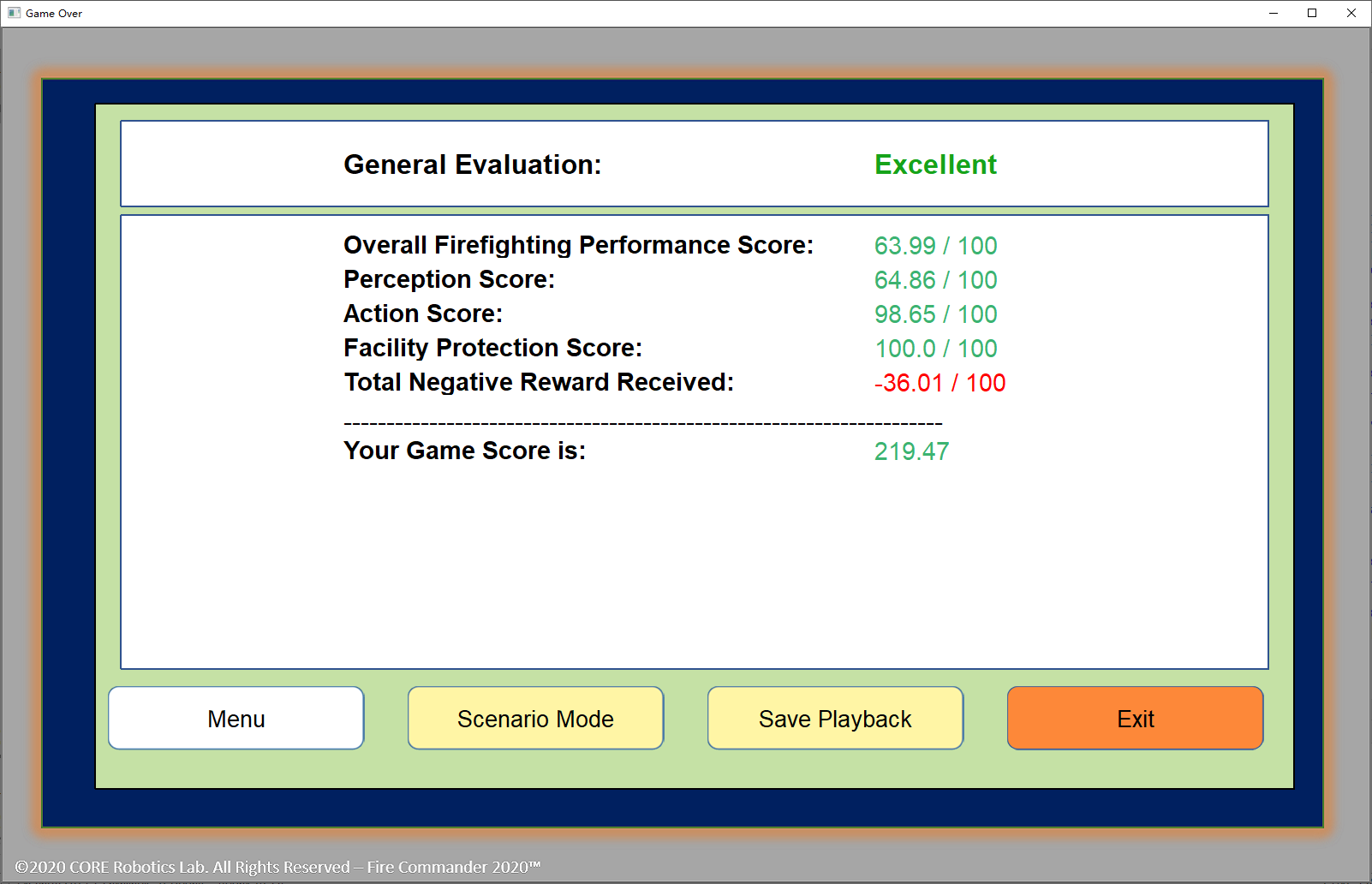}
    \caption{The Score Display.}
    \label{fig:score_display}
\end{figure}

The performance evaluation page appears when a game ends (both pre-designed and user-designed scenarios). A performance evaluation page is presented in Figure~\ref{fig:score_display}. On top, a verbal evaluation is reported to the player according to some pre-defined metrics, which are described below. Under the general verbal evaluation bar, the final values of the online scores introduced in Equations~\ref{eq:totalnegreward}--\ref{eq:facilityprrotectionscore} are again reported to the player. To generate the general verbal evaluation, we first calculate a final score as shown in Equation~\ref{eq:finalscore}:
\begin{align}
\label{eq:finalscore}
Final \ Score &= Perception \ Score + Action \ Score + Facility \ Protection \ Score \\
&-3 \times Negative \ Reward \ Ratio \notag
\end{align}
 The \textit{Negative Reward Ratio} in Equation~\ref{eq:finalscore} is computed as presented in Equation~\ref{eq:negativerewardratio} in which the numerator is the total negative reward calculated in Equation~\ref{eq:totalnegreward}. Moreover, since the penalty coefficient for fire propagation (e.g., 0.1), number of initial firespots and the game duration, $T$, are known to the environment, an expected negative reward can also be calculated as in Equation~\ref{eq:expectednegativereward}. We call the ratio between this expected negative reward and the actual total negative reward received by the user the \textit{Negative Reward Ratio}. We incorporate \textit{Negative Reward Ratio} to design the objective for protecting the facilities. Note that, according to Equation~\ref{eq:finalscore} and Equation~\ref{eq:negativerewardratio}, as long as none of the facilities/targets are on fire, the final score can be easily high (close to 90\%-100\%), given a reasonable perception and action performances. However, when firespots enter a facility, the value of negative reward ratio suddenly increases and the final score in Equation~\ref{eq:finalscore} drops. Note that the negative reward ratio also enforces agents to get rid of the firespots that have entered a facility as soon as possible, since the longer firespots are inside a facility, the larger will be the negative reward ratio (e.g., according to Equation~\ref{eq:totalnegreward}).
 \begin{align}
 \label{eq:negativerewardratio}
    Negative \ Reward \ Ratio = \frac{Total \ Negative \ Reward}{Expected \ Negative \ Reward} \times 100 \%
\end{align}
\begin{align}
\label{eq:expectednegativereward}
    Expected \ Negative \ Reward  = \sum_{t=1}^{T} 0.1 \times (Initial \ Number \ of \ Firespots \times t)
\end{align}

Now, leveraging the final score in Equation~\ref{eq:finalscore}, the verbal evaluation includes the following five levels: (1) \textbf{Failed} (\textit{Final Score} \textless 50), (2) \textbf{Fair} (50 \textless\textit{Final Score}\textless 60), (3) \textbf{Almost There!} (60 \textless\textit{Final Score}\textless 80), (4) \textbf{Well Done} (80 \textless\textit{Final Score}\textless 90) and, (5) \textbf{Excellent} (\textit{Final Score} \textgreater 90).

After the final performance evaluation page, users could either choose to return to the main menu, move to scenario mode, or simply exit the GUI. Additionally, we provided the option that users can save a playback of their game (e.g., their performance) as a video file. For this option, users can select the \underline{Save Playback} button to call the animation reconstruction function, which creates a video playback of the game from the saved data and stores the video on the computer.

\subsection{Animation Reconstruction}
\label{subsec:AnimationReconstruction}
\noindent We provided the option that users can save a playback of the game they just played (e.g., their performance) as a video file. For this option, users can select the \underline{Save Playback} button on the final performance evaluation page to call the animation reconstruction function, which creates a video playback of the game from the saved data and stores the video on the computer. The playback video can be used in image-based LfD and LfHD algorithms, or other observations for HRI/Psychology studies. During each trial of the game, the GUI writes the user data into a file with \textit{*.pkl} format with a frequency of 100Hz. The stored data is then used to reconstruct the video playback.

\subsection{Data Formats}
\label{subsec:DataFormats}
\begin{figure}[t!]
\centering
\includegraphics[width=\textwidth]{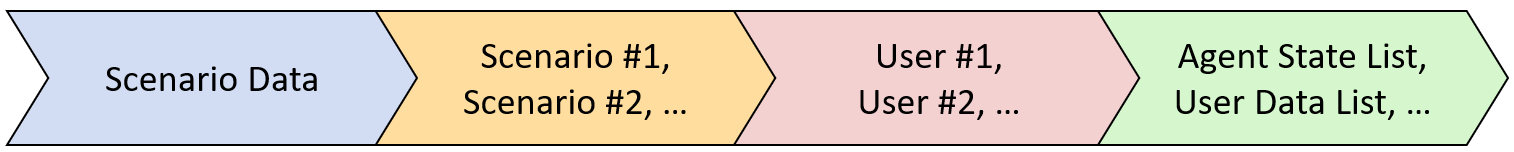}
\caption{The Data Storage Structure.}
\label{fig:data_store}
\end{figure}

\noindent All required information from a trial of the game are recorded and automatically stored in a designated folder. The recorded data, e.g., states of the environment, must cover 4 main aspects of the game: (1) agent states, (2) wildfire states, (3) facility/target states and (4) user-data information. We elaborate each category separately in the following sub-sections and show the structure of the stored data for each case. Figure~\ref{fig:data_store} shows the general data storage structure.

\begin{figure}[t!]
    \centering
    \includegraphics[width=\textwidth]{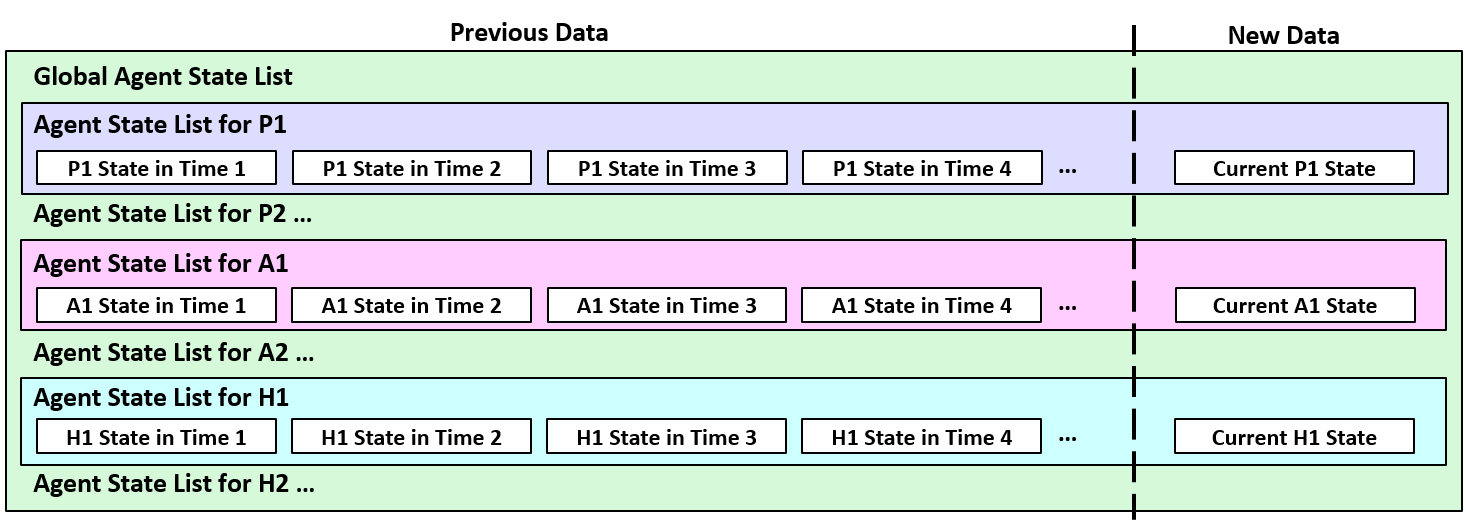}
    \caption{The Structure of the Global Agent State List.}
    \label{fig:global_agent_state}
\end{figure}

\begin{figure}[t!]
\centering
\includegraphics[width=\textwidth]{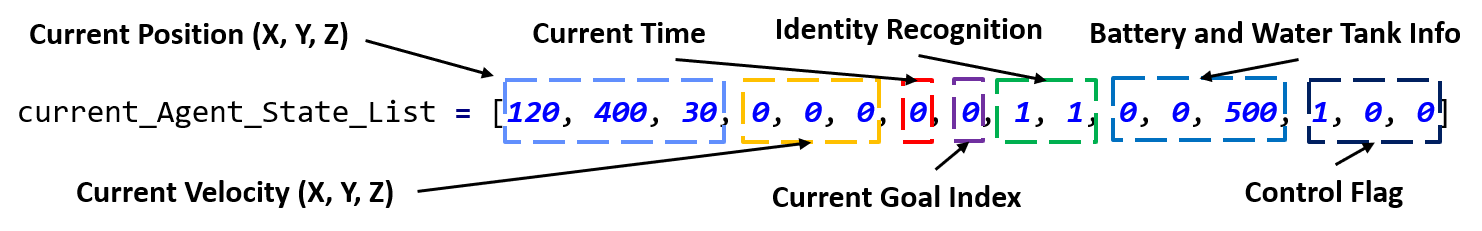}
\caption{The Agent State List for P1 at Time 0.}
\label{fig:agent_sample_list}
\end{figure}

\paragraph{Agent Information (States):} 
The agent information list includes all agents' states at all time-steps during the game. The agents states include their positions, velocities, identity and other necessary information for trajectory planning, sensing and pruning tasks. Figure~\ref{fig:global_agent_state} presents the global structure of the stored agents' state information. Note that, inside each individual agent's state list, the elements-order follows the structure depicted in Figure~\ref{fig:agent_sample_list}. Each Bit is described below:
\begin{itemize}
\item \textbf{Bit 0 -- 2:} Current Pose, (X, Y, Z) coordinates.

\item \textbf{Bit 3 -- 5:} Current Velocity along (X, Y, Z) axes, which determine the agent's next-step pose.

\item \textbf{Bit 6:} Current time-step.

\item \textbf{Bit 7:} Current goal index (e.g., in normal trajectory mode).

\item \textbf{Bit 8 -- 9:} Agent's Identity such that, bit 8 shows agent's type: (0) for Perception, (1) for Action and (2) for Hybrid agents. Bit 9 shows the agent's index in its specific category.

\item \textbf{Bit 10 -- 12:} Battery and water tank information, where the three bits represent: (1) current cumulative distance, (2) current cumulative waiting time and, (3) water tank capacity (Action and Hybrid agents), respectively.

\item \textbf{Bit 13 -- 15:} Control flag, where the three bits respectively represent: (1) \underline{movement flag}, which determines whether to finish the trajectory or to retreat to the base, (2) \underline{patrolling flag}, which determines the normal or the patrolling trajectory modes and, (3) patrolling goal index for the patrolling trajectory mode.
\end{itemize}

\begin{figure}[t!]
\centering
\includegraphics[width=\textwidth]{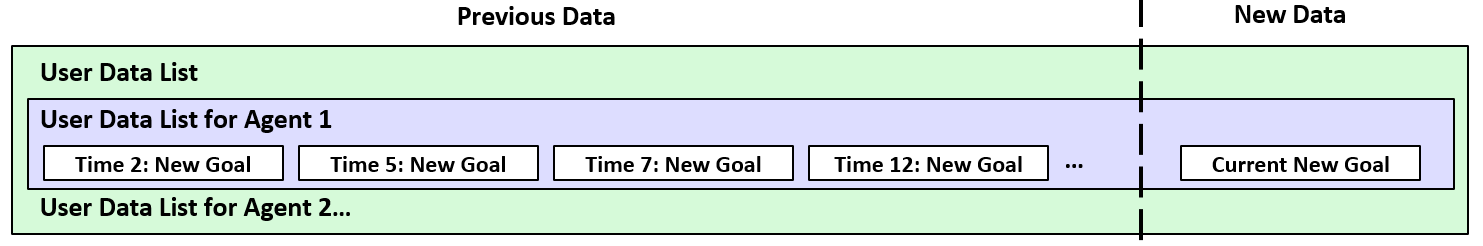}
\caption{The User Data List.}
\label{fig:user_data}
\end{figure}

\begin{figure}[t!]
\centering
\includegraphics[width=0.6\textwidth]{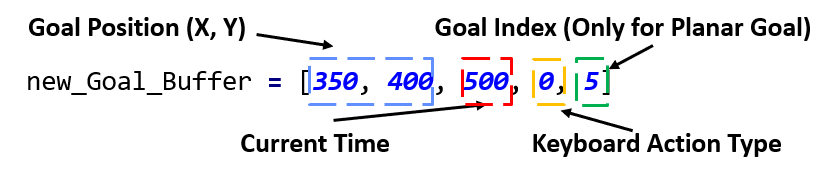}
\caption{The New Goal Buffer List.}
\label{fig:new_goal_buffer}
\end{figure}

\paragraph{User Data Information:}
The user-data information is a list including the player's actions on keyboard and mouse. Such user actions include, selecting an agent and planning trajectories for agents by adding new goal to the goal buffer, designing a normal or a patrolling trajectory and switching the current agent's flight altitude. Figure~\ref{fig:user_data} presents the general structure of the user-data list. The goal information for all the agents is stored in a single list. Since the agents' lists are stored sequentially, the list recording states for a specific agent can then be easily tracked. Once the new goal is created for the current agent, a unit list called new goal buffer is appended at the end of the user data list. The new goal buffer list stores the goal information for an agent generated at time $t$. The user-data list is arranged chronologically, in which the keyboard and mouse actions are stored together and are distinguished by the action type flag. Figure~\ref{fig:new_goal_buffer} presents the detail of the new goal buffer list, the bits of which are described below:
\begin{itemize}
\item \textbf{Bit 0 -- 1:} Goal position, (X, Y) coordinate of the user click on screen. Note that, in case of vertical motion, agent's previous position is repeated for current position. Diagonal motions are not allowed in this environment.

\item \textbf{Bit 2:} Current time of user's actions (e.g., mouse or keyboard use).

\item \textbf{Bit 3 -- 4:} Keyboard action type, such that, (0) is for a planar motion (mouse click) and, (1) is for vertical motion (arrow keys on keyboard).

\item \textbf{Bit 5:} Goal index for the planar goal.
\end{itemize} 

\begin{figure}[t!]
\centering
\includegraphics[width=\textwidth]{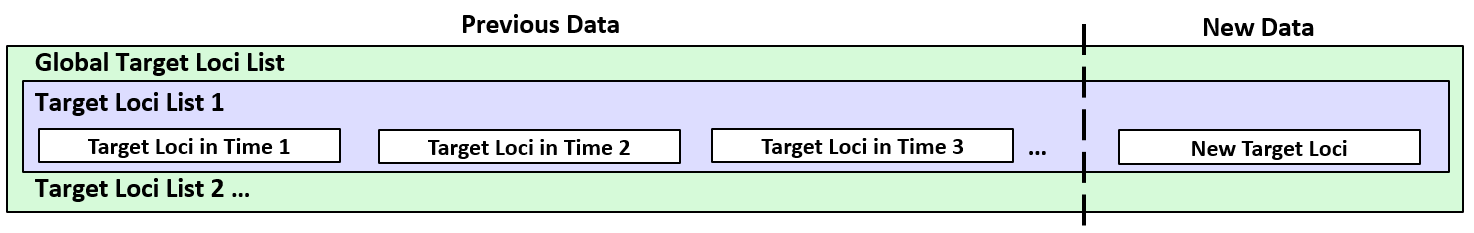}
\caption{The Target Loci List.}
\label{fig:target_loci_list}
\end{figure}

\begin{figure}[t!]
    \centering
    \includegraphics[width=\textwidth]{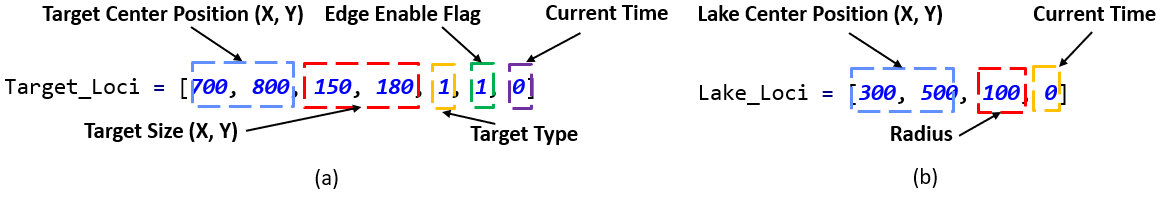}
    \caption{The Target Loci List Unit.}
    \label{fig:target_loci_list_unit}
\end{figure}

\begin{figure}[t!]
    \centering
    \includegraphics[width=\textwidth]{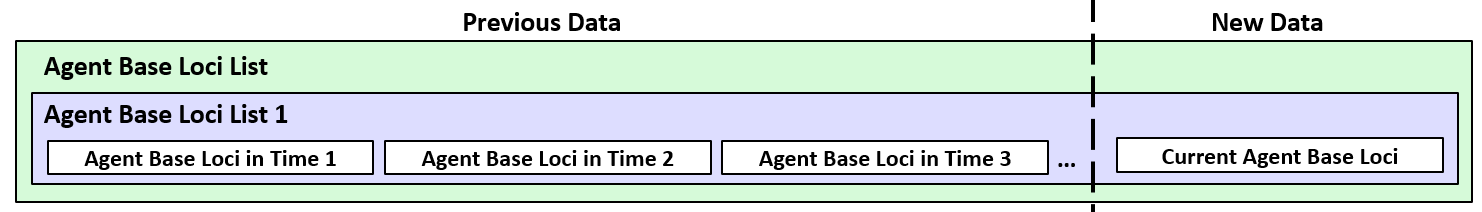}
    \caption{The Agent Base Loci List.}
    \label{fig:agent_loci_list}
\end{figure}

\begin{figure}[t!]
    \centering
    \includegraphics[width=\textwidth]{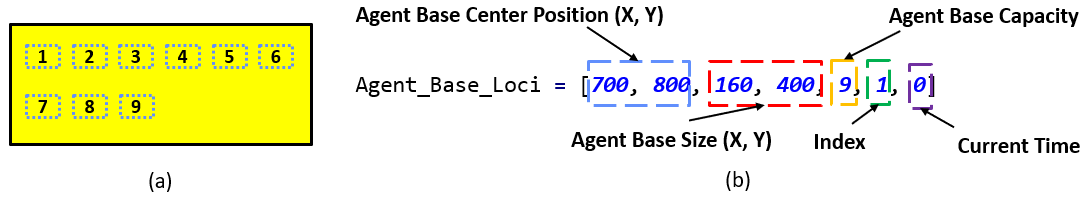}
    \caption{The Agent Base Layout and Its Composition Unit List.}
    \label{fig:agent_base_layout_detail}
\end{figure}

\paragraph{Target Loci List:}
The target loci list is used to store the stats of all facilities/targets in the environment at each moment. Although in our current version of the FireCommander the targets are always static, we provide the recording of targets' states at every time-step which allows further extensions of the environment to include the dynamic target. The target loci data structures are shown in Figures~\ref{fig:target_loci_list}--\ref{fig:agent_base_layout_detail}. The figures are self-descriptive and the notations are mostly similar to the previous notations used for agents' states data and user-data lists. Nevertheless, here are some useful information to help understand these figures better:
\begin{itemize}
    \item The general target loci list applies to the house, hospital, power station and lake. Although using a similar template, the information for the first three targets, house, hospital and power station, are stored in the same list, while the information for the lake is stored separately. The reason for rises from the fact that the designed lake has a large irregular shape.
    
    \item The agent base information is separated and is stored in a different file to make it easier for the users to have access to this data. Having a moving base also poses an interesting robotics problem in which a large, dynamic ground robot is used as a recharge station for small quadcopters. The quadcopters can be considered as Perception and Action agents which now need to coordinate and communicate with the moving base as well as their other teammates.
\end{itemize}

\begin{figure}[t!]
    \centering
    \includegraphics[width=\textwidth]{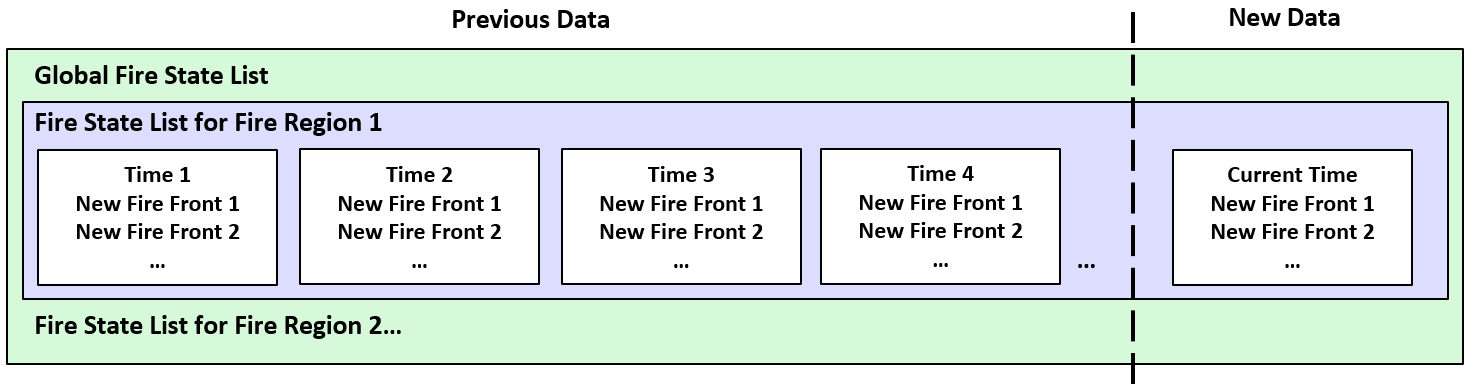}
    \caption{The Fire State List.}
    \label{fig:fire_state_list}
\end{figure}

\begin{figure}[t!]
    \centering
    \includegraphics[width=0.5\textwidth]{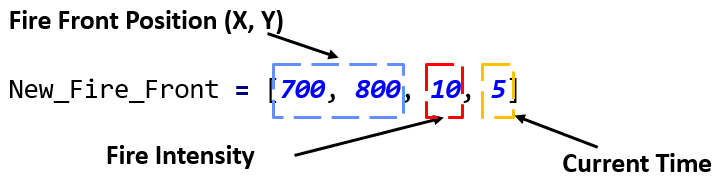}
    \caption{The New Fire Front List.}
    \label{fig:new_fire_list}
\end{figure}

\begin{figure}[t!]
    \centering
    \includegraphics[width=\textwidth]{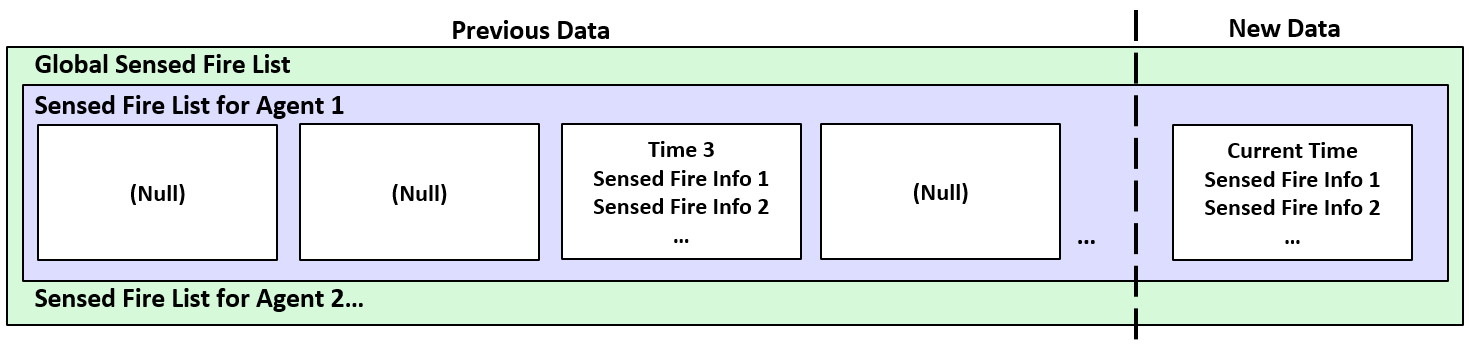}
    \caption{The Structure of the Sensed and Pruned Fire List.}
    \label{fig:sensed_list}
\end{figure}

\begin{figure}[t!]
    \centering
      \includegraphics[width=\linewidth]{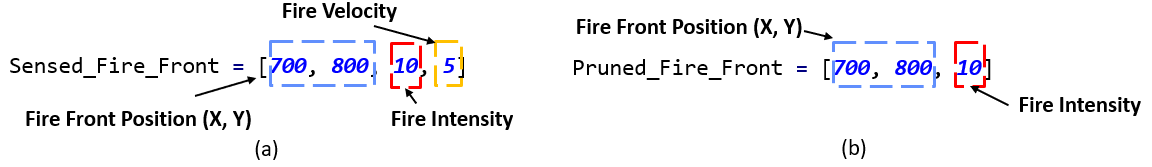}  
      \caption{The Sensed and Pruned Fire List Unit}
      \label{fig:sensed_pruned_list_unit}
\end{figure}

\paragraph{Wildfire States List:}
The wildfire list stores the states of firespots, that is the essential information required to track the fire propagation, including firespot coordinates and intensities at each time step. The fire states list consists of three separate types of fire information: (1) new firefronts generated, (2) sensed firespots and (3) pruned firespots. Figure~\ref{fig:fire_state_list} presents the structure of the fire state list that stores all of the generated firefronts, separately for each initial fire region and over time. Figure~\ref{fig:new_fire_list} presents the details for the new firefront list that records the position, intensity and generation time for new firefronts. Note that, in the discrete setting of PyGame, the fire coordinates (and all other coordinates as well) are always discrete values, and thus, often it may occur that two firespots propagate to the same location. In such cases, the fire intensity of the new firespot will be the linear summation of the two parent firespots. Figure~\ref{fig:sensed_list} presents the structure of stored data for the sensed and pruned firespots. The structure of the sensed and pruned fire front list are generally the same, and they both are similar to the general new firefront data list to some extend. The differences are: (1) rather than across fire regions and over time, firespot data here is stored across agents and over time and, (2) The sensed and pruned fires list entry at some time steps may be null. Figure~\ref{fig:sensed_pruned_list_unit}(a) presents the structure of the sensed fire front list that records the position, intensity and fire velocity for the sensed fire spots. The fire velocity is necessary to compute the center of mass for the continuous fire regions. Figure~\ref{fig:sensed_pruned_list_unit}(b) presents the structure of the pruned fire front list that records the position, intensity for the pruned fire spots.

\begin{figure}[t!]
\centering
\includegraphics[width=\textwidth]{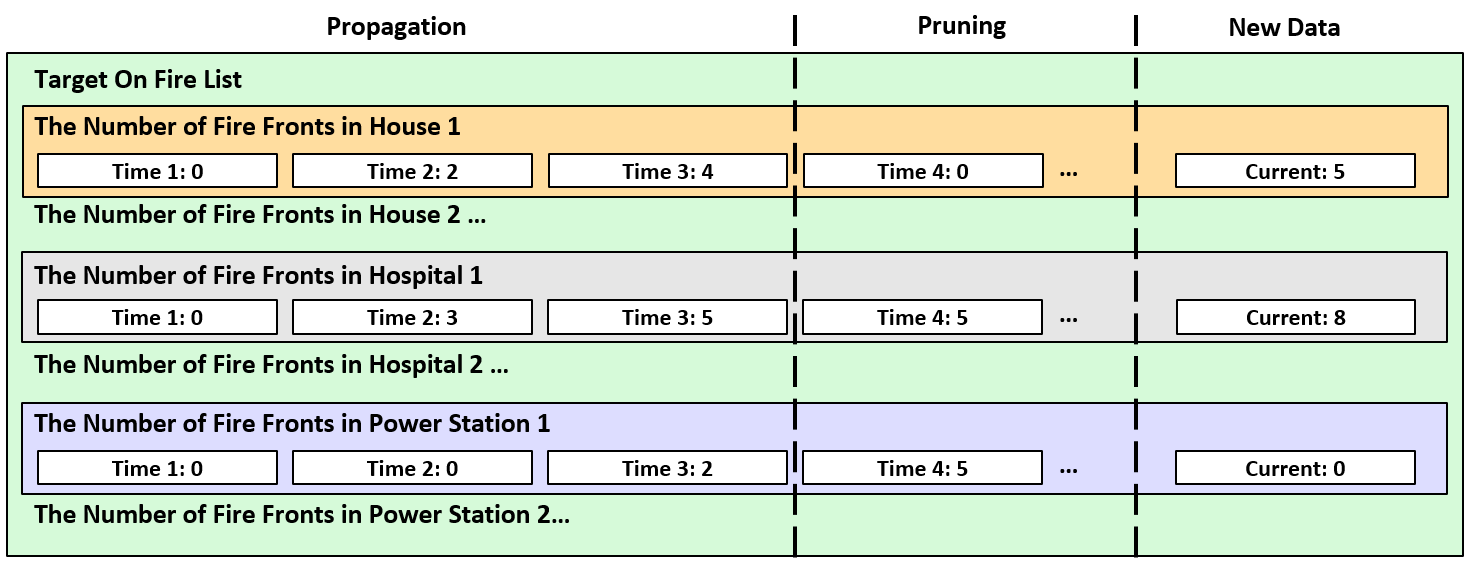}
\caption{The Target on Fire List.}
\label{fig:target_onfire}
\end{figure}

\paragraph{Storing the Information for the Firespots That Enter Facilities (Target-On-Fire List):}
We also need to keep track of the information of the firespots that enter facilities/targets (e.g., the target-on-fire list). Figure~\ref{fig:target_onfire} presents the target-on-fire list structure. This list stores the number of firespots that locate inside each facility/target at each time step. When an Action or Hybrid agent prunes some of the firespots inside a target, the respective firespots are removed from the list for that time-step onward. Note that, although the user receives the positive reward for removing the firespots inside a target and potentially avoid receiving exponentially growing negative rewards at future time steps, the current target is still counted as non-protected and thus, a negative reward will be counted in the facility protection score in Equation~\ref{eq:facilityprrotectionscore}.

\section{Acknowledgment}
The authors would like to thank Letian Chen and Rohan Paleja for their insightful comments and feedback.\blfootnote{Other Toolboxes by E. Seraj:
\begin{itemize}
    \item Essential Motor Cortex Signal Processing: an ERP and functional connectivity MATLAB toolbox~\cite{seraj2019essential}
    
    \item Cerebral Signal Instantaneous Parameters Estimation MATLAB Toolbox~\cite{seraj2016cerebral}
\end{itemize}}

\section{References}
\bibliographystyle{IEEEtran}
\bibliography{IEEEabrv,references}

\end{document}